\documentclass{article}

\usepackage{arxiv}

\usepackage[utf8]{inputenc} % allow utf-8 input
\usepackage[T1]{fontenc}    % use 8-bit T1 fonts
\usepackage{hyperref}       % hyperlinks
\usepackage{url}            % simple URL typesetting
\usepackage{booktabs}       % professional-quality tables
\usepackage{amsfonts}       % blackboard math symbols
\usepackage{nicefrac}       % compact symbols for 1/2, etc.
\usepackage{microtype}      % microtypography
\usepackage{lipsum}		% Can be removed after putting your text content
\usepackage{graphicx}
\usepackage{natbib}
\usepackage{doi}

\usepackage{amssymb}

\usepackage{subfig}

\usepackage{siunitx}
\usepackage{etoolbox}
\usepackage[ruled,norelsize,vlined]{algorithm2e}

\usepackage{scrextend}

\usepackage{lineno}

\usepackage{chngcntr}
\usepackage{apptools}

\usepackage{bbm}%indicator function
\usepackage{amsfonts} %for real numbers
\usepackage{amsmath} %for formulas
\usepackage{accents}
\usepackage{dsfont}
\usepackage{upgreek}
\usepackage{bold-extra}

\usepackage{enumerate}
\usepackage[mathscr]{euscript}

\usepackage{float}

\usepackage{geometry}%to change margin
\usepackage{caption}
\usepackage{subcaption}

\usepackage{bookmark}

\usepackage{marginnote}
\usepackage{mathtools}
\usepackage{mathrsfs}

\usepackage{nicefrac}

\usepackage{array}

\usepackage{tikz} %for graph
\usetikzlibrary{fit,positioning,calc,decorations}

\usepackage{wrapfig}
\usepackage{xcolor,colortbl} %to color cell in table

\usepackage{url}

\usepackage{comment}
\usepackage{makecell}
\usepackage{booktabs} 
\usepackage{multirow}
\usepackage{tabu}

\usepackage{float}

\usepackage{xcolor}

\usepackage{appendix}

\usepackage[utf8]{inputenc}

\allowdisplaybreaks
\usepackage{fix-cm} % for arbitrary font sizes
\newcommand\intermediate[1]{\begingroup\fontsize{9}{11}\selectfont#1\endgroup}

\usepackage{palatino}
\fontfamily{ppl}\selectfont

\definecolor{airforceblue}{rgb}{0.36, 0.54, 0.66}
\definecolor{blue(pigment)}{rgb}{0.2, 0.2, 0.6}
\definecolor{ashgrey}{rgb}{0.7, 0.75, 0.71}
\definecolor{darkgrey}{rgb}{0.66, 0.66, 0.66}
\definecolor{cadetgrey}{rgb}{0.57, 0.64, 0.69}
\definecolor{aoe}{rgb}{0.0, 0.5, 0.0}
\definecolor{burntorange}{rgb}{0.8, 0.33, 0.0}
\definecolor{bostonuniversityred}{rgb}{0.8, 0.0, 0.0}
\definecolor{burgundy}{rgb}{0.0, 0.0, 1}

\DeclareMathAlphabet\mathbfcal{OMS}{cmsy}{b}{n}

\newcommand{\E}{\mathbb{E}}

\newcommand{\bzero}{ {\bf 0} }

\newcommand{\bx}{ {\bf x} }

\newcommand{\by}{ {\bf y} }
\newcommand{\bX}{ {\bf X} }
\newcommand{\bR}{ {\bf R} }

\newcommand{\bu}{ {\bf u} }
\newcommand{\bD}{ {\bf D} }
\newcommand{\bI}{ {\bf I} }

\newcommand{\bV}{ {\bf V} }

\newcommand{\bA}{ {\bf A} }

\newcommand{\bC}{ {\bf C} }

\newcommand{\br}{ {\bf r} }

\newcommand{\bepsilon}{ {\boldsymbol \epsilon} }

\newcommand{\bLambda}{ {\mathbf \Lambda} }

\newcommand{\bGamma}{ {\mathbf \Gamma} }

\newcommand{\bbeta}{ {\boldsymbol \beta} }
\newcommand{\bmu}{ {\boldsymbol \mu} }
\newcommand{\bSigma}{ {\mathbf \Sigma} }
\newcommand{\bsigma}{ {\boldsymbol \sigma} }
\newcommand{\bOmega}{ {\mathbf \Omega} }
\newcommand{\bTheta}{ {\mathbf \Theta} }

\newcommand{\bphi}{ {\boldsymbol \phi} }

\newcommand{\bz}{ {\bf z} }
\newcommand{\btheta}{ {\boldsymbol \theta} }
\newcommand{\boeta}{ {\boldsymbol \eta} }

\newcommand{\argmax}{\mbox{argmax}}

\DeclareSymbolFont{matha}{OML}{txmi}{m}{it}% txfonts
\DeclareMathSymbol{\varv}{\mathord}{matha}{118}

\makeatletter
\newcommand*\bigcdot{\mathpalette\bigcdot@{.7}}
\newcommand*\bigcdot@[2]{\mathbin{\vcenter{\hbox{\scalebox{#2}{$\m@th#1\bullet$}}}}}
\makeatother

\newtheorem{Theorem}{\bf Theorem}

\newtheorem{Proposition}[Theorem]{\bf Proposition}

\def\simind{\stackrel{\mbox{\scriptsize{\rm ind}}}{\sim}}
\def\simiid{\stackrel{\mbox{\scriptsize{\rm iid}}}{\sim}}

\DeclareRobustCommand{\bbone}{\text{\usefont{U}{bbold}{m}{n}1}}

\makeatletter
\newcommand{\vast}{\bBigg@{3}}
\newcommand{\Vast}{\bBigg@{4}}
\makeatother

\usepackage{enumitem}
\setenumerate[1]{label=\arabic*.}

\usepackage{titlesec}

\usepackage{adjustbox} % in the preamble
\usepackage{tabularx}

\definecolor{violet}{rgb}{0.0, 0.0, 0.0} 
\definecolor{teal}{rgb}{0.0, 0.0, 0.0}

\title{Bayesian Joint Additive Factor Models \\ for Multiview Learning}

\author{Niccolo Anceschi\\
	Department of Statistical Science\\
	Duke University\\
	Durham, NC 27708, USA \\
	\texttt{niccolo.anceschi@duke.edu} \\
    \And
	Federico Ferrari \\
    Biostatistics and Research Decision Sciences  
    \\
	Merck \& Co., Inc.\\
	Rahway, NJ 07065, USA \\
	\texttt{federico.ferrari@merck.com}	
    \And
	David B.~Dunson$^*$ \\
	Department of Statistical Science \\
	Duke University\\
	Durham, NC 27708, USA \\
	\texttt{dunson@duke.edu}	
    \And
	Himel Mallick\thanks{Co-corresponding authors}\\
	Department of Population Health Sciences\\
	Division of Gastroenterology and Hepatology\\
	Weill Cornell Medicine\\
	New York, NY 10065, USA\\
	Department of Statistics and Data Science\\
	Cornell University\\
	Ithaca, NY 14850, USA\\
	\texttt{him4004@med.cornell.edu}
 }
  
\date{}

\renewcommand{\headeright}{}
\renewcommand{\undertitle}{}
\renewcommand{\shorttitle}{Bayesian Joint Additive Factor Models for Multiview Learning}

\hypersetup{
pdftitle={Bayesian Joint Additive Factor Models for Multiview Learning},
pdfauthor={Niccolo Anceschi, Federico Ferrari, David B. Dunson, Himel Mallickx},
}

\begin{document}
\maketitle

\begin{abstract}
It is increasingly common to collect data of multiple different types on the same set of samples. Our focus is on studying relationships between such multiview features and responses. A motivating application arises in the context of precision medicine where multi-omics data are collected to correlate with clinical outcomes. It is of interest to infer dependence within and across views while combining multimodal information to improve the prediction of outcomes. The signal-to-noise ratio can vary substantially across views, motivating more nuanced statistical tools beyond standard late and early fusion. This challenge comes with the need to preserve interpretability, select features, and obtain accurate uncertainty quantification.
\textcolor{teal}{
To address these challenges, we introduce two complementary factor regression models. A baseline Joint Factor Regression (\textsc{jfr}) captures combined variation across views via a single factor set, and a more nuanced Joint Additive FActor Regression (\textsc{jafar}) that decomposes variation into shared and view-specific components. For \textsc{jfr}, we use independent cumulative shrinkage process (\textsc{i-cusp}) priors, while for \textsc{jafar} we develop a dependent version (\textsc{d-cusp}) designed to ensure identifiability of the components. We develop Gibbs samplers that exploit the model structure and accommodate flexible feature and outcome distributions.
}
Prediction of time-to-labor onset from immunome, metabolome, and proteome data illustrates performance gains against state-of-the-art competitors. Our open-source software (\texttt{R} package) is available at \href{https://github.com/niccoloanceschi/jafar}{\texttt{https://github.com/niccoloanceschi/jafar}}.
\end{abstract}

% keywords can be removed
\keywords{
Bayesian inference \and Multiview data integration \and Factor analysis \and Identifiability \and Latent variables \and Precision medicine}

%%%@@@@@@@@@@@@@@@@@@@@@@@@@@@@@@@@@@@@@@@@@@@@@@@@
%%                  Introduction
%%%@@@@@@@@@@@@@@@@@@@@@@@@@@@@@@@@@@@@@@@@@@@@@@@@
\section{Introduction}
In personalized medicine, it is common to gather vastly different kinds of complementary biological data by simultaneously measuring multiple assays in the same subjects, ranging across the genome, epigenome, transcriptome, proteome, and metabolome 
% \citep{Stelzer_2021_labor_onset, Ding_2022_coopL}.
\citep{Ding_2022_coopL}.
Integrative analyses that combine information across such data views can deliver more comprehensive insights into patient heterogeneity and the underlying pathways dictating health outcomes \citep{mallick2024integrated}. Similar setups arise in diverse scientific contexts including wearable devices, electronic health records, and finance, among others \citep{lee2020multimodal, li2021integrating, mcnaboe2022design}, where there is enormous potential to integrate the concurrent information from distinct vantage points to better understand between-view associations and improve prediction of outcomes.

Multiview datasets have specific characteristics that complicate their analyses: (i) they are often high-dimensional, noisy, and heterogeneous, with confounding effects unique to each layer (e.g., platform-specific batch effects); (ii) sample sizes are often very limited, particularly in clinical applications; and (iii) signal-to-noise ratios can vary substantially across views, which must be accounted for in the analysis to avoid poor results. Many methods face difficulties in identifying the predictive signal since it is common for most of the variability in the multiview features to be unrelated to the response.
Our primary motivation in this article is thus to enable accurate and interpretable outcome prediction while allowing inferences on within- and across-view dependence structures. 

Carefully structured factor models that infer low-dimensional joint- and view-specific sources of variation are particularly promising. Early contributions in this space focused on the unsupervised paradigm \citep{Lock_2013_JIVE, Li_2017_SIFA, Argelaguet_2018_MOFA}. Two-step approaches, \textcolor{teal}{first fitting these unsupervised factorizations on the predictors and then using their scores to predict the response,} often fail to identify subtle response-relevant factors, leading to subpar predictive accuracy. More recent contributions considered integrative factorizations in a supervised setting \citep{Palzer_2022_sJIVE, Li_2022_IFR}. 

\textcolor{teal}{
Bayesian formulations of supervised integrative factorizations have also been proposed \citep{Chekouo2021,samorodnitsky2024bayesian}. However, current approaches have limitations, particularly in high-dimensional settings. The priors used to learn structured activity patterns impose overly rigid constraints or introduce bias, reducing the flexibility of the model and ultimately impairing the predictive accuracy. Computations are often inefficient, limiting scalability to larger or more complex datasets. Furthermore, no existing framework is fully Bayesian in simultaneously estimating the number of factors while reliably disentangling shared from view-specific components.}

Alternative approaches focusing \textcolor{teal}{only} on prediction accuracy include Cooperative Learning \citep{Ding_2022_coopL} and IntegratedLearner \citep{mallick2024integrated}.  These methods combine usual squared-error loss-based predictions with a suitable machine learning algorithm. However, by conditioning on the multiview features, neither approach allows inferences on or exploits information from inter- and intra-view correlations. One consequence is a tendency for unstable and unreliable feature selection, as from a predictive standpoint it is sufficient to select any one of a highly correlated set of features.

\textcolor{teal}{To address these gaps, we propose two complementary strategies. First, a baseline joint factor regression (\textsc{jfr}) model captures the combined variation across multiple views via a single set of factors.
Building on this, we introduce a more nuanced joint additive factor regression (\textsc{jafar}) model, which explicitly decomposes variation into shared and view-specific components while preserving model flexibility. The benefits of isolating variation unique to each modality, whether genuine signals or measurement artifacts, become increasingly evident in high-dimensional settings.
In both cases, we introduce novel priors on the shared loadings (\textsc{i-cusp} and \textsc{d-cusp}) that generalize the cumulative shrinkage process (\textsc{cusp}) \citep{Legramanti_2020_CUSP}. 
For \textsc{jafar}, we induce dependence across views and ensure correct separation of shared and view-specific factors.
We develop Gibbs samplers that exploit the model structures and propose a modification of the \texttt{Varimax} step in \texttt{MatchAlign} \citep{poworoznek2025_MatchAlign} to resolve rotational ambiguity in the shared component.
We further propose a tempered version of the \textsc{cusp} constructions to better control rank estimation in extreme large-p-small-n settings.
Both \textsc{jfr} and \textsc{jafar} are validated through simulation studies and real data analyses, demonstrating improved estimation and prediction compared to published methods.
}

The remainder of the paper is organized as follows. The proposed methodology is presented in detail in Section~\ref{sec_methodology} focusing on an initial Gaussian specification, while flexible semiparametric extensions are detailed in the Appendix. 
In Section~\ref{sec_simulations}, we focus on simulation studies to validate the performance of \textsc{jfr} and \textsc{jafar} against state-of-the-art competitors.
The empirical studies from Section~\ref{sec_application} further showcase the benefits of our contribution on real data.
An open-source implementation of our methods is available in the \texttt{R} package \href{https://github.com/niccoloanceschi/jafar}{\texttt{jafar}}.

%%%@@@@@@@@@@@@@@@@@@@@@@@@@@@@@@@@@@@@@@@@@@@@@@@@
%%             Material and methods
%%%@@@@@@@@@@@@@@@@@@@@@@@@@@@@@@@@@@@@@@@@@@@@@@@@
\section{Factor Regression Formulations }\label{sec_methodology}

To address the aforementioned challenges and deliver accurate response prediction from multiview data, \textcolor{teal}{we consider two different versions of Bayesian factor regression.
The first is a joint factor regression (\textsc{jfr}) structure with a single set of latent factors}
\begin{equation}\label{eq_jfr_linear}
\raisebox{-0.3\height}{\text{\textsc{jfr}}}
\quad \left| \quad
\begin{aligned}
    % ~ & \\[-32pt]
    \bx_{m i} &= \bmu_m + \bLambda_{m} \boeta_i + 
    \bepsilon_{m i}  \\
    y_i &= \mu_y + \btheta^\top \boeta_i + e_i \, .
\end{aligned}
\right.
\end{equation}
Here $ \bx_{m i} \in \Re^{p_m} $ and $y_i \in \Re $ represent the multiview data and the response, respectively, for each observation $i \in \{1, \dots, n\}$ and modality $m \in \{1, \dots, M\}$, where $n$ and $M$ are the total number of subjects and views, respectively.
$\bLambda_{m} \in \Re^{p_m \times K}$ are loadings matrices associated with $K$ latent factor $\boeta_{i} \in \Re^{K}$, while $\btheta \in \Re^{K}$ is a set of factor regression coefficients connecting the response to the shared latent factors, complemented by an offset term $\mu_y \in \Re$.
The residual components $e_i$ and $\bepsilon_{m i} $ are assumed to follow normal distributions $\mathcal{N}(0,\sigma_y^2)$ and $\mathcal{N}_{p_m}(\bzero_{p_m}, \operatorname{diag}(\bsigma_m^2))$, with $\bsigma_m^2=\{\sigma_{m j}^2\}_{j=1}^{p_m}$.
Here $\mathcal{N}$ and $\mathcal{N}_p$ represent the univariate and $p$-variate Gaussian distributions.
\textcolor{teal}{
While $\eta_i$ is potentially connected to all components of the data, the distinction between fully shared, partially shared, and view-specific factors depends on the sparsity pattern in the loading matrices. Similar joint factorizations have been considered in the literature, including \citet{FENG2018,Argelaguet_2018_MOFA,Gaynanova2019,Chekouo2021,Yi2023}.
}

\textcolor{teal}{Secondly, we propose modifying   \textsc{jfr} to a joint additive factor regression (\textsc{jafar}) structure}
\begin{equation}\label{eq_jafar_linear}
\raisebox{-0.3\height}{\text{\textsc{jafar}}}
\quad \left| \quad
\begin{aligned}
    % ~ & \\[-32pt]
    \bx_{m i} &= \bmu_m + \bLambda_{m} \boeta_i + \bGamma_{m} \bphi_{m i} + 
    \bepsilon_{m i}  \\
    y_i &= \mu_y + \btheta^\top \boeta_i + \textstyle{\sum_{m=1}^M} \btheta_m^\top \bphi_{m i} + e_i \, ,
\end{aligned}
\right.
\end{equation}
\textcolor{violet}{isolating $\{K_m\}_{m=1}^M$ view-specific latent factors $\bphi_{m i} \in \Re^{K_m}$, connected to the corresponding view via loading matrices $\bGamma_{m} \in \Re^{p_m \times K_m}$ and to the response via factor regression coefficients $\theta_m \in \Re^{K_m}$.
In \textsc{jafar}, the $\boeta_i$'s are designated to capture only fully and partially shared factors, impacting at least two data views.
The local factors $\{ \bphi_{m i} \}_{m=1}^M$ aim at capturing view-specific sources of variation, unrelated to other modalities.}
\textcolor{teal}{Related additive decompositions with local factors  have been proposed, including \citet{Lock_2013_JIVE,Park2019_BIDIFAC,Palzer_2022_sJIVE,Lock2022,samorodnitsky2024bayesian}.}

\textcolor{teal}{
The two models aim at capturing the same signal, so that $K$ in \textsc{jafar} is smaller than in \textsc{jfr}, whereas $K + \sum_m K_m$ in \textsc{jafar} matches the overall rank in \textsc{jfr}.
Imposing structured sparsity as in \textsc{jafar} can be greatly beneficial in high dimensions, increasing interpretability, and substantially improving scalability.
These scalability advantages can be leveraged in posterior sampling, predictions, and post-processing of the results.
However, \textsc{jafar}'s structure is sensible only under correct separation of shared and specific components.}
In fact, additive latent factor formulations suffer from a specific form of non-identifiability, beyond the typical issues of single-view factor models, such as rotational ambiguity and columns and sign switching \citep{poworoznek2025_MatchAlign}.

\subsection{Non-identifiability of Additive Factor Models}
\begin{figure}[t!]
   \centering
   \includegraphics[height=0.4\linewidth]{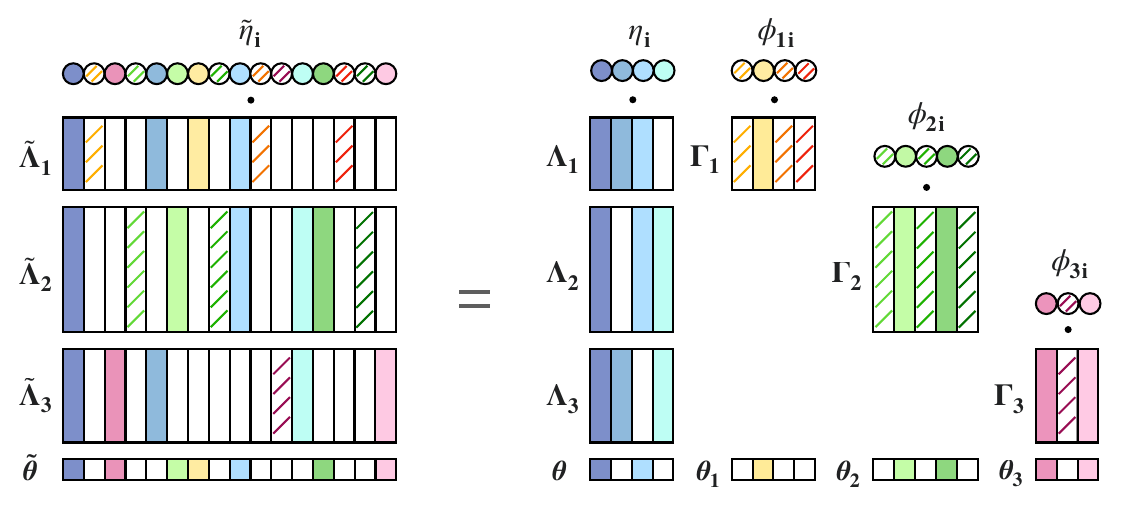}
   \vspace{-5pt}
   \caption{Schematic visualization of the representations given by \textsc{jfr} (left) and \textsc{jafar} (right), highlighting their potential equivalence. White boxes represent blocks of zero loadings, signaling the inactivity of the corresponding factor in a given data component.
   Isolating view-specific factors enhances interpretability and computations.
   However, additive factor models suffer from a specific form of non-identifiability, as formulations with global-local latent factors could be re-expressed as a global-only block-sparse version. 
   Hence, the usefulness of additive representations hinges on reliably distinguishing view-specific factors from fully or partially shared ones.}
   \label{fig_non_identif}
\end{figure}

\textcolor{violet}{Intuitively, the source of this additional non-identifiability comes from the possibility of obtaining the 
\textsc{jafar} structure by enforcing an appropriate sparsity pattern on the matrices $\bLambda_m$ in \textsc{jfr}.}
To illustrate this, note that the inter- and intra-view covariances induced by \textsc{jafar} in marginalizing out the latent factors are
\begin{equation}\label{eq_induced_cov}
\operatorname{cov}(\bx_{m i}) = \bLambda_m \bLambda_m^\top + \bGamma_m \bGamma_m^\top + \operatorname{diag}(\bsigma_m^2),\quad \operatorname{cov}(\bx_m,\bx_{m'}) = \bLambda_m \bLambda_{m'}^\top \; .
\end{equation}
Concatenating all views into $\bx_i = [\bx_{1i}^\top,\dots,\bx_{M i}^\top]^\top$, this entails that the view-specific components $\bGamma_m$ affect only the block-diagonal element of the induced covariance.
Crucially, the same joint covariance matrices could be obtained 
by collapsing the \textsc{jafar} formulation into a \textsc{jfr} format, by dropping all view-specific components in favor of a unique set of shared factors 
$\widetilde{\boeta}_i = [\boeta_i, \bphi_{1 i}, \dots, \bphi_{M i}]$ and
sparse loadings matrices 
% $\widetilde{\bLambda}_m$, $\widetilde{\boeta}_i$
\begin{equation*}
% \label{eq_shared_fact_ambiguity}
\begin{aligned}
    \widetilde{\bLambda}_m &= [\bLambda_m, \bzero_{p_m \times K_1}, \dots, \bzero_{p_m \times K_{m-1}}, \bGamma_m, \bzero_{p_m \times K_{m+1}}, \dots, \bzero_{p_m \times K_M }] \; .
\end{aligned}
\end{equation*}
The same holds for the conditional distribution of the response $y_i \mid \bx_i$ under extended regression coefficients $\widetilde{\btheta} = [\btheta^\top, \btheta_1^\top, \dots, \btheta_M^\top]^\top$.
This equivalence is \textcolor{violet}{ visually represented in Figure~\ref{fig_non_identif}. 
Hence, we cannot distinguish view-specific from shared patterns without constraints.} 

\vspace{-5pt}
% \vspace{-10pt}

\subsubsection{Approaches to solving non-identifiability of additive decompositions}
\textcolor{teal}{
Identifiability of the local versus global components of joint latent factorization models has been widely studied.
In deterministic and optimization-based factorizations, identifiability is sought through exact geometrical constraints such as orthogonality or linear independence of latent axes of variation.
Prominent contributions include
\textsc{jive} \citep{Lock_2013_JIVE},
s\textsc{jive} \citep{Palzer_2022_sJIVE},
\textsc{ajive}\citep{FENG2018},
\textsc{slide} \citep{Gaynanova2019}, 
\textsc{bidifac} \citep{Park2019_BIDIFAC}, 
\textsc{bidifac}+ \citep{Lock2022}, and
\textsc{hnn} \citep{Yi2023}.
Since our formulations aim at capturing all of fully-shared, partially-shared, and view-specific factors, the conditions discussed in \textsc{slide} or \textsc{hnn} are the most relevant.
}

\textcolor{teal}{
By contrast, our approach is fully Bayesian and grounded in probabilistic generative models.
Orthogonality and related constraints are seldom implemented in such a setting, as they substantially complicate \textsc{mcmc} design \citep{Duan2019relaxation}.
Standard practice in the Bayesian factor model focuses on identifiability of the induced covariance (and of its low-rank and diagonal components) in the initial posterior sampling stage of the analysis. Rotational ambiguity of the latent factors and loading vectors is then addressed in a post-processing stage; e.g, by rotationally aligning MCMC samplings of the factor loadings and factors via \texttt{Varimax} \citep{poworoznek2025_MatchAlign}
}

\textcolor{teal}{
We embrace such an approach, addressing correct separation of the additive components of the covariance matrices directly in the \textsc{mcmc}, similarly to parallel threads on multi-study settings \citep{Roy_2021_Perturbed_FA,chandra2025_sufa}.
Rotational ambiguity is solved via a modified post-processing, accounting for the composite structure of the problem.
In Bayesian contexts, it is sufficient for statistical inferences to focus on {\em generic} identifiability, which is weaker than {\em strict} identifiability in permitting non-identifiability for a measure-zero subset of parameter space.
For example, refer to 
\cite{Gu2023Pyramids}.
The approach we take is similar, in that we achieve generic identifiability by choosing priors on the loadings $\bLambda_m$ of the shared component in equation~\eqref{eq_jafar_linear} that assign measure zero 
to factor misallocation between shared and specific components.
}

%%%$$$$$$$$$$$$$$$$$$$$$$$$$$$$$$$$$$$$$$$$$$$$$$$$$$
\subsection{Prior formulation}\label{sec_dep_cusps}

To maintain computational tractability in high dimensions, we assume conditionally conjugate priors for most components of the model.
\begin{equation*}
\begin{aligned}
    \boeta_{i } &\simiid \mathcal{N}_K(\bzero_K, \bI_K) \qquad &
    \mu_y &\sim \mathcal{N}(0, \upsilon_y^2) \qquad & 
    \sigma_y^2 &\sim \mathcal{I}nv\mathcal{G}a (a_y^{\scriptscriptstyle{(}\scriptstyle{\sigma}\scriptscriptstyle{)}},b_y^{\scriptscriptstyle{(}\scriptstyle{\sigma}\scriptscriptstyle{)}}) \\
    \bphi_{m i} &\simiid \mathcal{N}_{K_m}(\bzero_{K_m}, \bI_{K_m}) \qquad\; &
    \mu_{m j} &\simiid \mathcal{N}(0, \upsilon_m^2) \qquad\; &
    \sigma_{m j}^2 &\simiid \mathcal{I}nv\mathcal{G}a (a_m^{\scriptscriptstyle{(}\scriptstyle{\sigma}\scriptscriptstyle{)}},b_m^{\scriptscriptstyle{(}\scriptstyle{\sigma}\scriptscriptstyle{)}}) 
\end{aligned}
\end{equation*}
Consistent with standard practice, we assume independent standard normal priors for all factors, including $\bphi_{m i}$ in \textsc{jafar}.
By $\mathcal{I}nv\mathcal{G}a$ and $\mathcal{B}e$ we denote the inverse gamma and the beta distributions, respectively.
\textcolor{teal}{To distinguish clinically relevant factors from irrelevant ones,} \textcolor{violet}{we postulate a spike and slab prior on the response loadings, with shared hyper-variance}
\begin{equation*}
\begin{aligned}
    \btheta_h &\simind \mathcal{N}(0, \psi^2_{h}) \qquad 
    & \psi^2_{h} &\simind r_h \; \delta_{\psi^2_{o}}   + (1-{r}_h) \; \delta_{\psi^2_{\infty}} \; 
    &r_{zh}, r_{m\, h} &\simiid \mathcal{B}e(a^{\scriptscriptstyle (r)},b^{\scriptscriptstyle (r)}) \\
    \btheta_{m \, h} &\simind \mathcal{N}(0, \psi^2_{m \, h})  \qquad
    & \psi^2_{m \, h} &\simind r_{m\, h} \; \delta_{\psi^2_{o}}   + (1-r_{m\, h}) \; \delta_{\psi^2_{\infty}} \; 
    &\psi^2_{o} &\sim \mathcal{I}nv\mathcal{G}a(a^{\scriptscriptstyle (\theta)},b^{\scriptscriptstyle (\theta)})
\end{aligned}
\end{equation*}

\textcolor{teal}{
As for loading matrices, recall that we want to simultaneously learn the number of factors and differentiate between fully shared, partially shared, and view-specific latent factors.}

\subsubsection{Factor loadings in \textsc{jfr}: independent cumulative shrinkage processes (\textsc{i-cusp})} 
\textcolor{violet}{
To achieve these goals, we propose an extension of the \textsc{cusp} prior of \citet{Legramanti_2020_CUSP}.
\textsc{cusp} adaptively removes unnecessary factors from an over-fitted factor model by progressively shrinking the loadings to zero by
leveraging stick-breaking representations of Dirichlet processes \citep{ishwaran2001_DP}.
In \textsc{jfr}, we assume independent \textsc{cusp} priors (\textsc{i-cusp}) for the view-specific loadings $\bLambda_m \sim \textsc{cusp}(a^{\scriptscriptstyle (L)}_m,b^{\scriptscriptstyle (L)}_m,\tau^2_{m \, \infty },\alpha^{\scriptscriptstyle (L)}_m) $, with}
\begin{equation}\label{eq_lambda_spike_and_slab}
  \bLambda_{m j h} \sim \mathcal{N}(0, \tau^2_{m h}) \qquad \qquad
\tau^2_{m h} \sim \pi_{mh}^{\scriptscriptstyle(\Lambda)} \; \mathcal{I}nv\mathcal{G}a(a^{\scriptscriptstyle (L)}_m,b^{\scriptscriptstyle (L)}_m) + \big(1-\pi_{mh}^{\scriptscriptstyle(\Lambda)} \big) \; \delta_{\tau^2_{m \infty}} \; .
\end{equation}
Accordingly, the increasing shrinkage behavior is induced by the weight of the spike and slab
\begin{equation}\label{eq_lambda_stick_breaking}
    \pi_{mh}^{\scriptscriptstyle(\Lambda)} = 1 - \textstyle{\sum_{l=1}^h} \omega_{m l} \qquad \;
    \omega_{m h} = \nu_{m h} \, \textstyle{\prod_{l=1}^{h-1}} (1-\nu_{m l}) \qquad \;
    \nu_{m h} \sim \mathcal{B}e (1, \alpha^{\scriptscriptstyle (L)}_m) \; ,
\end{equation}
such that $\mathbb{P}\big[|\bLambda_{m j h+1}| \leq \varepsilon\big] > \mathbb{P}\big[|\bLambda_{m j h}| \leq \varepsilon\big]$, $\forall \varepsilon>0$, provided that $b^{\scriptscriptstyle (L)}_m/a^{\scriptscriptstyle (L)}_m > \tau^2_{m \infty}$.

The stick-breaking process can be rewritten in terms of latent ordinal variables $\zeta_{m h} \in \mathbb{N}$, where a priori $\mathbb{P}[\zeta_{m h} = l] = \omega_{m l}$ for each $h,l \geq 1$, such that $\pi_{mh}^{\scriptscriptstyle(\Lambda)} = \mathbb{P}[\zeta_{m h} > h]$.
The $h^{th}$ column is defined as active when it is sampled from the slab, namely if $\zeta_{m h} > h$, and inactive otherwise.
\textcolor{teal}{
In fact, the spike components $\delta_{\tau^2_{m \infty}}$ provide a continuous relaxation of exact sparsity, setting $\tau^2_{m \infty}$ to small values.
Accordingly, the $h^{th}$ factor is effectively fully shared if $\sum_{m=1}^M \mathbbm{1}_{(\zeta_{m h} > h)} = M$, partially shared if $1<\sum_{m=1}^M \mathbbm{1}_{(\zeta_{m h} > h)} < M$ and modality-specific if $\sum_{m=1}^M \mathbbm{1}_{(\zeta_{m h} > h)} = 1$.
}
\textcolor{violet}{
Here $\mathbbm{1}_{( C )}$ denotes the indicator function, taking the value 1 if the condition $C$ is satisfied and 0 otherwise.
}

It is standard practice to truncate such infinite representations, starting from a conservative upper bound for the number of factors $K$. This retains sufficient flexibility while allowing for tractable posterior inference via a conditionally conjugate Gibbs sampler.
\textcolor{teal}{Following \citet{Legramanti_2020_CUSP}, we then learn $K$ throughout the inferential procedure via an adaptive Gibbs sampler. 
With decreasing probability, the sampler either drops columns that are inactive in all $\bLambda_{m}$'s while maintaining a buffer inactive factor in the rightmost column, or adds a buffer factor if all columns are active.
}
Ergodicity of the resulting \textsc{mcmc} chain is still preserved, provided that the probability of performing such an operation meets suitable diminishing adaptation conditions \citep{roberts2007coupling}.

\subsubsection{Factor loadings in \textsc{jafar}: Dependent cumulative shrinkage processes (\textsc{d-cusp})}
\textcolor{teal}{In the case of \textsc{jafar}, we tackle identification of shared and view-specific factors via a modified prior structure, still leveraging the \textsc{cusp} construction.}
For the view-specific components, we still consider independent \textsc{cusp} priors on the loadings $\bGamma_m \sim \textsc{cusp}(a^{\scriptscriptstyle (L)}_m,b^{\scriptscriptstyle (L)}_m,\tau^2_{m \, \infty },\alpha^{\scriptscriptstyle (\Gamma)}_m) $, with
\begin{equation*}
\begin{aligned}
    & \bGamma_{m j h} \sim \mathcal{N}(0, \chi^2_{m h}) \qquad \;
    \chi^2_{m h}  \sim \pi_{mh}^{\scriptscriptstyle(\Gamma)} \; \mathcal{I}nv\mathcal{G}a(a^{\scriptscriptstyle (L)}_m,b^{\scriptscriptstyle (L)}_m) + \big(1-\pi_{mh}^{\scriptscriptstyle(\Gamma)} \big) \; \delta_{\tau^2_{m \infty}}  \\[2pt]
    & \hspace{-10pt} \pi_{mh}^{\scriptscriptstyle(\Gamma)} = 1 - \textstyle{\sum_{l=1}^h} \xi_{m l} \qquad \;
    \xi_{m h} = \rho_{m h} \, \textstyle{\prod_{l=1}^{h-1}} (1-\rho_{m l}) \qquad \;
    \rho_{m h} \sim \mathcal{B}e (1, \alpha^{\scriptscriptstyle (\Gamma)}_m) \; .
\end{aligned}
\end{equation*}

\textcolor{teal}{For the shared loading matrices $\{\bLambda_m \}_{m=1}^M$ instead, we introduce dependence across views within associated \textsc{cusp} constructions as in equations \eqref{eq_lambda_spike_and_slab} and \eqref{eq_lambda_stick_breaking}.
In particular, such dependence is mediated by the spike and slab mixture weights $\big\{\{\omega_{m h}\}_{h \geq 1}\big\}_{m=1}^M$, leveraging their representation in terms of latent ordinal variables $\big\{\{\zeta_{m h}\}_{h \geq 1}\big\}_{m=1}^M$.}
\textcolor{violet}{As before,} each column of the loading matrices will be sampled from the spike or slab depending on these indicators.
Recall that our goal is to prevent configurations where any shared factor is active in less than two model components.
Accordingly, it is reasonable to request that the $h^{th}$ factor is included in the shared variation part of equation~\eqref{eq_jafar_linear} if and only if the corresponding loadings are active in at least two views, \textcolor{teal}{that is if and only if $\textstyle{\sum_{m=1}^M} \mathbbm{1}_{(\zeta_{m \, h} > h)} > 1$.}

\textcolor{teal}{
We enforce this constraint through the adaptation step, where we drop all columns 
% $H_{\text{drop}} \subset \{1,\dots, K\}$
\begin{equation*}
    H_{\text{drop}} = \Big\{ h \in \{1,\dots, K\} \;\Big\vert\; \textstyle{\sum_{m=1}^M} \mathbbm{1}_{(\zeta_{m \, h} > h)} < 2 \Big\} 
\end{equation*}
that are active in less than two views.
This puts zero mass on configurations with misallocated factors.
Satisfying the diminishing adaptation condition guarantees once more ergodicity of the resulting sampler.
We refer to the resulting prior as the dependent \textsc{cusp}s (\textsc{d-cusp}) prior, and we write $\{\bLambda_m\}_m \sim \textsc{d-cusp}\big( \{ a^{\scriptscriptstyle (L)}_m,b^{\scriptscriptstyle (L)}_m,\tau^2_{m \, \infty },\alpha^{\scriptscriptstyle (\Lambda)}_m \}_m \big) $.}

\subsubsection{Alternative prior formulations and competitors}\label{sec_other_priors}
\textcolor{teal}{
Alternative prior formulations and Bayesian competitors will be considered. Here, we describe a selection of these approaches, with associated results reported in Sections~\ref{sec_simulations} and~\ref{sec_application}.
First, a naive variant (\textsc{naive}) of our methodology would simply postulate independent \textsc{cusp}s on the $\bLambda_m$'s in \textsc{jafar}, and omit the separation mechanism in the adaptation step. The main drawback is inconsistent factor allocation and reduced interpretability, possibly reflected in reduced accuracy.
A second alternative (\textsc{full-d}) would be to introduce stronger dependence across $\bLambda_m$'s, other than in the adaptation step, by making the prior activation probability of a factor in one view explicitly dependent on its activation in others, setting
\begin{equation*}
\begin{aligned}
    \pi_{mh}^{\scriptscriptstyle(\Lambda)} &= \mathbb{P} \Big[\{\zeta_{m h} >h \} \cap \big\{ {\textstyle \bigcup_{m' \neq m}} \{\zeta_{m' h} >h \} \big\} \Big] 
    %
    % = \mathbb{P} [\zeta_{m h} >h ] \, \mathbb{P} \Big[ {\textstyle \bigcup_{m' \neq m}} \{\zeta_{m' h} >h \} \Big] \\
    % &= \mathbb{P} [\zeta_{m h} >h ] \, \Big( 1 - \mathbb{P}\Big[ {\textstyle \bigcap_{m' \neq m}} \{\zeta_{m' h} \leq h \} \Big]  \Big) 
    %
    = \mathbb{P} [\zeta_{m h} >h ] \, \Big( 1 - {\textstyle \prod_{m' \neq m}} \mathbb{P}[ \zeta_{m' h} \leq h ] \Big) \;.
\end{aligned}
\end{equation*}
However, the results in Section~\ref{sec_sim_unsupervised} show that this approach tends to inflate the number of shared factors by artificially increasing their prior activation probability.
}

\textcolor{teal}{
\textsc{bip} \citep{Chekouo2021} parallels  \textsc{jfr} by positing a single set of factors and learning sparsity in the loadings $\{\bLambda_m\}_m$ via spike and slabs priors.
\textsc{bip} lacks a mechanism to set the rank and enforces exact zeros (i.e. $\tau^2_{m \infty}=0$) within a double layer of sparsity, both view-wise and feature-wise.
The authors develop a partially collapsed Gibbs sampler \citep{Park_vanDyk_2009_Partially_Collapsed_Gibbs_Samplers}, marginalizing out the loadings matrices to cope with the difficulties of handling exact zeros, requiring a Metropolis-within-Gibbs step to sample the mixture memberships.
Formulations with exact zeros are known to reduce mixing and hinder identification of the inactive factors compared to continuous shrinkage priors, which partly motivated the development of the original \textsc{cusp} construction in \citet{Legramanti_2020_CUSP}.
}

\textcolor{teal}{
Finally, \textsc{bsfp} incorporates view-specific components by extending \textsc{unifac} to a Bayesian context, setting the prior to the Gaussian equivalent of the nuclear norm penalty. 
Recall that \textsc{unifac} does not account for partially shared factors in the common component.
The ranks are set via the \textsc{unifac} solution, which corresponds to the posterior mode and is thus uniquely identified. The authors note that the decomposition at each posterior sampling iteration may not be identifiable, without addressing consequences on the induced covariances. \textsc{bsfp} inherits \textsc{unifac}’s bias induced by the nuclear norm penalization, which \citet{Yi2023} observed to negatively affect signal estimation performance.
}

%%%$$$$$$$$$$$$$$$$$$$$$$$$$$$$$$$$$$$$$$$$$$$$$$$$$$
\subsection{Posterior inference via Gibbs samplers}

Under the proposed extension of the \textsc{cusp} construction to the multiview case, the \textcolor{violet}{linear model} versions of \textcolor{violet}{\textsc{jfr} and} \textsc{jafar} still allow for straightforward Gibbs sampling via conjugate full conditionals.
\textcolor{violet}{
\textcolor{violet}{In \textsc{jfr},} most of the associated full conditionals take the same forms as those of a regular factor regression model under the \textsc{cusp} prior.
In \textsc{jafar}, one naive option is to sample sequentially the elements of each additive component, as done in \textsc{bsfp}.
However, simple one-at-a-time updates can lead to worse mixing.
Hence, we propose two joint updates for latent factors and shared loadings, respectively.}
\textcolor{teal}{Leveraging the structured sparsity still allows great computational improvement over the non-partitioned representation in \textsc{jfr}.
In the interest of readability, the details of the Gibbs samplers are provided in Appendix~\ref{app_gibbs_sampler}, while here we discuss only the resulting costs.
}

\textcolor{teal}{
Let us assume again that the overall rank in \textsc{jfr}  is equal to $K+\sum_{m=1}^M K_{m}$ from \textsc{jafar}. 
The sampling of loadings inevitably has a $  \mathcal{O}\big( (\sum_{m=1}^M p_m) \cdot (K+\sum_{m'=1}^M K_{m})^3 \big)$ cost in \textsc{jfr}, due to the lack of sparsity.
By contrast, joint updates can be performed at a total cost $ \mathcal{O}\big( \sum_{m=1}^M  p_m \cdot(K+ K_m)^3 \big)$ in \textsc{jafar}. 
This rapidly becomes the bottleneck step of the sampler as the dimension of the view increases. Jointly updating the latent factors
comes at the same cost in \textsc{jfr} and \textsc{jafar}.
Notably, the precision matrix of the corresponding full conditional is the same across all statistical units. This allows to reduce the cost to $\mathcal{O}\big((K+\sum_{m=1}^M K_m)^3 + n \cdot (K+\sum_{m=1}^M K_m)^2\big)$, rather than $\mathcal{O}\big(n \cdot(K+\sum_{m=1}^M K_m)^3 \big)$, clearly advantageous for large sample sizes.
In the unsupervised version of \textsc{jafar}, this cost can be further brought down to $\mathcal{O}\big(K^3+\sum_{m=1}^M K_m^3 + n \cdot (K^2+\sum_{m=1}^M K_m^2)\big)$ within a partially collapsed Gibbs sampler.
}

\subsubsection{Tempering against a curse of dimensionality in extreme large-p-small-n scenarios}\label{sec_curse_dim_tempering}

\textcolor{teal}{
Motivating applications include multiomics data, often falling in extreme large-p-small-n settings.
To the best of our knowledge, \textsc{cusp} remained untested in such extreme scenarios.
% partly due to an unnecessary  $\mathcal{O}(p_m^3)$ scaling of its original implementation.
In Section~\ref{sec_application}, we consider a dataset in such a regime, where \textsc{jfr} and \textsc{jafar} achieve competitive accuracy but infer a relatively high number of factors under the proposed extensions of \textsc{cusp}.
Specifically, the inferred ranks are far smaller than the $p_m$'s but greater than $n$.
This is undesirable and runs counter to some of the central rationales behind low-rank factorization.
}

\textcolor{teal}{
The \textsc{cusp} prior was introduced in the context of classical Bayesian factor models,
where factorization aims to reduce dimensionality with respect to the number of features $p$ and not the sample size $n$.
Instead of arbitrarily modifying hyperparameters to increase shrinkage, we propose an approach motivated by the Bayesian robustness literature on fractional posteriors \citep{Bhattacharya2019fractional}.
In particular, in the Gibbs sampling update of the latent ordinal variables $\zeta_{m h}$, which drives rank selection through factor activity, we choose a fractional power to induce $n$ as the effective sample size for this update instead of $p_m$. We refer to the resulting tempered versions as \textsc{jfr}$_T$ and \textsc{jafar}$_T$, and provide a more detailed description in Appendix~\ref{app_model_extensions_nonlinear}.
We emphasize that these tempered formulations are not intended as a complete replacement of the original ones, but rather as optional adaptations to be used in extremely unbalanced large-p-small-n scenarios.
}

% \vspace{-5pt}

\subsubsection{Missing Data}
In many applications, a significant proportion of features may have missing measurements. Missingness can occur in blocks, where certain modalities are measured only for specific subgroups of subjects. 
\textcolor{violet}{
Bayesian factor models provide an effective way to handle such cases by focusing on likelihood contributions from the induced submodel defined on the observed features, with the missing ones marginalized out.
Imputation of the missing values can also be performed, sampling from their full conditional posterior distributions.
We advise against using the resulting complete data to inform other Gibbs sampler updates, as this leads to worse mixing compared to conditioning only on the observed data.
}

%%%$$$$$$$$$$$$$$$$$$$$$$$$$$$$$$$$$$$$$$$$$$$$$$$$$
\subsection{Postprocessing and Multiview \texttt{MatchAlign}}

Loading matrices still suffer from notorious
rotational ambiguity, label switching, and sign switching. 
Indeed, 
% it is easy to verify that the induced joint covariance decomposition is not unique.
focusing on the shared components, consider semi-orthogonal matrices $\bR$ of dimensions $K \times K$.
Then, the transformed set of loadings $\ddot{\bLambda}_m = \bLambda_m \bR$ clearly satisfy $\ddot{\bLambda}_m \ddot{\bLambda}_{m'}^\top=\bLambda_m \bLambda_{m'}^\top$, which leaves the covariances $\operatorname{cov}(\bx_m)$ and $\operatorname{cov}(\bx_m,\bx_{m'})$ unaffected, while adequately transforming $\btheta$ and $\boeta_i$ to $\ddot{\btheta} = \btheta \bR$ and $\ddot{\boeta}_i = \bR^\top \boeta_i$  leave unchanged the distribution of $y_i$. 
This poses challenges when inferring latent variables and factor loadings. 

In Bayesian contexts, such issues can be effectively addressed via postprocessing. Here we follow \citep{poworoznek2025_MatchAlign}, first applying \texttt{Varimax} rotations \citep{Kaiser1958_Varimax} to each loadings sample and then resolving column label and sign ambiguities by matching each sample to a reference via a greedy optimization procedure.
\textcolor{teal}{To better respect the multiview structure, we introduce a modified \texttt{Varimax} criterion for joint components, mitigating the sensitivity of naive stacking to view dimensionality and preventing dominant views from distorting cross-view sparsity. Full details are provided in Web Appendix~C.}

%%%@@@@@@@@@@@@@@@@@@@@@@@@@@@@@@@@@@@@@@@@@@@@@@@@
%%             Simulations
%%%@@@@@@@@@@@@@@@@@@@@@@@@@@@@@@@@@@@@@@@@@@@@@@@@
\section{Simulation Studies}\label{sec_simulations}
\textcolor{violet}{To assess the performance of the proposed methodologies, we first conducted simulation experiments, running \textsc{jfr} and \textsc{jafar} 
under the \textsc{i-cusp} and \textsc{d-cusp} priors, respectively.}
\textcolor{teal}{
In Section~\ref{sec_sim_unsupervised}, we assess the effectiveness in separating shared and view-specific components in an unsupervised setting.
We include in the analysis the alternative priors for \textsc{jafar} mentioned in Section~\ref{sec_other_priors}, namely \textsc{naive} and \textsc{full-d}.
In Section~\ref{sec_sim_supervised}, we evaluate performance in jointly achieving accurate view reconstruction and strong predictive accuracy in supervised setups.
We consider two recent Bayesian competitors, \textsc{bip} and \textsc{bsfp}, the latter also serving as an indirect benchmark having recently demonstrated good performance compared to alternative latent factorization approaches \citep{samorodnitsky2024bayesian}.}
We also examine two non-factor models alternatives: Cooperative Learning (\texttt{CoopLearn}) and IntegratedLearner (\texttt{IntegLearn}).
\texttt{CoopLearn} complements usual squared-error loss-based predictions with an agreement penalty, which encourages predictions coming from separate data views to match with one another. 
\texttt{IntegLearn} combines the predictions of Bayesian additive regression trees (\textsc{bart}) fit separately to each view.

\textcolor{violet}{
We focus on large-p-small-n settings to reflect the structure of the motivating applications and to create challenging test cases. 
Data are generated from the factor model with the additive structure in equation~\eqref{eq_jafar_linear}.
To mimic realistic multiview data, we do not sample the loadings from the prior.
Doing so typically produces near-diagonal covariance structures—a shortcoming often overlooked in the literature.
Instead, we introduce a novel scheme for generating loading matrices that induces sensible block-structured correlations. 
Details of the data-generating mechanism are provided in Appendix~\ref{app_realistic_simulations}.
}
We standardized both the multiview features and the response prior to analysis, following the recommended procedure for applying \textsc{cusp} \citep{Legramanti_2020_CUSP}. \textcolor{teal}{\textsc{bsfp} internally rescales the predictors to have unit error variance -- rather than overall unit variance -- using the median absolute deviation estimator \citep{Gavish_2017_MAD}. These two rescaling strategies are not in conflict, and \textsc{bsfp} automatically transforms its output back to the scale of the original inputs.}

\textcolor{teal}{In all our analyses, the hyperparameters of the priors for \textsc{jfr} and \textsc{jafar} were set to}
\vspace{-5pt}
{
\begin{equation}\label{eq_hyperparams}
% \mycustomsize %,\small
\setlength{\arraycolsep}{10pt}
\renewcommand{\arraystretch}{1.2}
% \begin{array}{cccc}
\begin{array}{clll}
\text{( Data distributions )} &  a_m^{\scriptscriptstyle{(}\scriptstyle{\sigma}\scriptscriptstyle{)}} =
a_y^{\scriptscriptstyle{(}\scriptstyle{\sigma}\scriptscriptstyle{)}} = 3 & b_m^{\scriptscriptstyle{(}\scriptstyle{\sigma}\scriptscriptstyle{)}} =
b_y^{\scriptscriptstyle{(}\scriptstyle{\sigma}\scriptscriptstyle{)}} = 1 & 
\upsilon_m^2 = \upsilon_y^2 = 0.25 \\ 
\text{( Spike \& slab variances )} & 
a^{\scriptscriptstyle (L)}_m = a^{\scriptscriptstyle (\theta)} = 0.5 & 
b^{\scriptscriptstyle (L)}_m = b^{\scriptscriptstyle (\theta)} = 0.1 & 
\tau^2_{m \, \infty} = \chi^2_{\infty } = 0.005 \\ 
\text{( Spike \& slab weights )} & 
a^{(\xi)} = 3 & 
b^{(\xi)} = 2 & 
\textcolor{gray}{\alpha^{\scriptscriptstyle (L)}_m,
\alpha^{\scriptscriptstyle (\Lambda)}_m, \alpha^{\scriptscriptstyle (\Gamma)}_m} \; , \\ 
\end{array}
\end{equation}
}
\textcolor{teal}{and detailed guidance for such choices is provided in Web Appendix~A.
The default values in equation~\eqref{eq_hyperparams} have been tuned for robustness and perform well across a wide range of simulations and applications.
In practice, we recommend adjusting only $\alpha^{\scriptscriptstyle (L)}_m$, $\alpha^{\scriptscriptstyle (\Lambda)}_m$, and $\alpha^{\scriptscriptstyle (\Gamma)}_m$ to guide the prior expectation on the number of factors.}
Conversely, we run \textsc{bip} \textcolor{violet}{and \textsc{bsfp}} under the default setup and hyperparameters from the official implementation.
For \texttt{IntegLearn}, we use the default late fusion scheme to integrate the individual models.
We set the agreement parameter in \texttt{CoopLearn} to $\rho_\texttt{CL}=0.5$.

\subsection{Separation of shared and view-specific components in unsupervised settings}\label{sec_sim_unsupervised}

\textcolor{teal}{We considered 100 independent replicate datasets with
$M=3$ views of dimensions $p_m=\{100,200,300\}$ and sample sizes $n=50$.
The assumed number of shared factors was set to $K^{(true)}=4$, while the view-specific ones were $\big\{ K_m^{(true)} \big\}_{m=1}^M = \{9,10,11\}$. 
The upper bound on the number of factors was set to $K^\textsc{max}_{tot}=60$ in \textsc{jfr}, and to $K^\textsc{max}=40$ and $ \{ K_m^\textsc{max} \}
_{m=1}^M = \{30,30,30\}$ for shared and specifics components of \textsc{jafar}.
For all $m=1,\dots,M$, we set $\alpha^{\scriptscriptstyle (\Lambda)}_m = \alpha^{\scriptscriptstyle (\Gamma)}_m =10$ in \textsc{jafar} and an equivalent $\alpha^{\scriptscriptstyle (L)}_m=40$ in \textsc{jfr}.
We run all Gibbs samplers for a total of $T_{\textsc{mcmc}}=10000$ iterations, with a burn-in of $T_{\textsc{burn-in}}=5000$ steps and thinning every $T_\textsc{thin}=10$ samples for memory efficiency.
}

% \vspace{-5pt}
\begin{table}[hb!]
\centering
{Inferred Number of Active Factors}\\[5pt]
\begin{adjustbox}{width=0.87\textwidth}
\begin{tabular}{c|c|rrr|r|rrr|r
}
\toprule
\multicolumn{2}{c|}{} & \multicolumn{3}{c|}{\small{Specific Components}} & \multicolumn{5}{c}{\small{Shared Component}} \\
\small{Model} & \small{Prior} & \multicolumn{1}{l}{$\Gamma_{1}$} & \multicolumn{1}{l}{$\Gamma_{2}$} & \multicolumn{1}{l|}{$\Gamma_{3}$} & \multicolumn{1}{l|}{$\Lambda_{tot}$} & \multicolumn{1}{l}{$\Lambda_{1}$} & \multicolumn{1}{l}{$\Lambda_{2}$} & \multicolumn{1}{l|}{$\Lambda_{3}$} &  \multicolumn{1}{l}{$\Lambda_{\text{one-only}}$} \\ 
 \midrule
% Truth & 9 \hspace{15pt} & 10 \hspace{15pt} & 11 \hspace{15pt} & 4 \hspace{15pt}  & 3 \hspace{15pt}  & 2\hspace{19pt}  & 3\hspace{19pt}   & 0 \hspace{15pt}~ \\[2pt] 
\multirow{3}{*}{\vspace{-6pt}~\textsc{jafar}} & \textsc{d-cusp} & 5.6$_{1.4}$ & 8.5$_{1.1}$ & 8.5$_{1.3}$ & 4.2$_{0.9}$ & 2.7$_{0.6}$ & 1.9$_{0.8}$ & 2.9$_{0.5}$ & 0.1$_{0.2}$ \\[2pt]
& \textsc{naive} & 4.7$_{1.3}$ & 7.7$_{1.2}$ & 7.0$_{1.4}$ & 8.4$_{1.7}$ & 3.9$_{1.0}$ & 2.8$_{1.1}$ & 4.6$_{1.1}$ & 4.2$_{1.7}$ \\[2pt]
& \textsc{full-d} & 3.7$_{1.4}$ & 6.8$_{1.1}$ & 5.9$_{1.4}$ & 16.2$_{2.8}$ & 11.1$_{2.3}$ & 9.6$_{2.6}$ & 11.3$_{2.5}$ & 0.2$_{0.2}$ \\[2pt]
\cmidrule{1-10}
\textsc{jfr} & \textsc{i-cusp} & -\hspace{15pt} & -\hspace{15pt}  & -\hspace{15pt}  & 27.7$_{2.4}$ & 8.4$_{1.2}$ & 10.5$_{1.4}$ & 11.7$_{1.2}$ & 23.2$_{2.6}$ \\
\bottomrule
\end{tabular}
\end{adjustbox}
%%%%%%%%%%%%%%%%
~\\[15pt]
{Correlation Matrices - Reconstruction Error}\\[5pt]
\begin{adjustbox}{width=\textwidth}
\begin{tabular}{c|c|ccc|ccc
}
\toprule
\small{Model} & \small{Prior} & $cor(X_1)$ & $cor(X_2)$ & $cor(X_3)$ & $cor(X_1,X_2)$ & $cor(X_1,X_3)$ & $cor(X_2,X_3)$ \\ \midrule
\multirow{3}{*}{\vspace{-6pt}~\textsc{jafar}} & \textsc{d-cusp} & 0.0132$_{0.0025}$ & 0.0124$_{0.0015}$ & 0.0130$_{0.0021}$ & 0.0040$_{0.0033}$ & 0.0058$_{0.0025}$ & 0.0019$_{0.0006}$ \\[3pt]
 & \textsc{naive} & 0.0132$_{0.0024}$ & 0.0125$_{0.0016}$ & 0.0131$_{0.0021}$ & 0.0042$_{0.0034}$ & 0.0062$_{0.0032}$ & 0.0019$_{0.0006}$ \\[3pt]
 & \textsc{full-d} & 0.0140$_{0.0030}$ & 0.0129$_{0.0017}$ & 0.0136$_{0.0024}$ & 0.0032$_{0.0011}$ & 0.0064$_{0.0030}$ & 0.0021$_{0.0006}$ \\[3pt]
\cmidrule{1-8}
\textsc{jfr} & \textsc{i-cusp} & 0.0150$_{0.0028}$ & 0.0139$_{0.0017}$ & 0.0143$_{0.0025}$ & 0.0033$_{0.0010}$ & 0.0066$_{0.0032}$ & 0.0024$_{0.0006}$ \\
\bottomrule
\end{tabular}
\end{adjustbox}
%%%%%%%%%%%%%%%%
~\\[15pt]
{Shared and View-Specific Correlations - Reconstruction Error}\\[5pt]
\begin{adjustbox}{width=\textwidth}
\begin{tabular}{c|c|ccc|ccc
}
\toprule
\small{Model} & \small{Prior} & $\Lambda_1 \Lambda_1^\top $ & $\Lambda_2 \Lambda_2^\top $ & $\Lambda_3 \Lambda_3^\top $ & $\Gamma_1 \Gamma_1^\top $ & $\Gamma_2 \Gamma_2^\top $ & $\Gamma_3 \Gamma_3^\top $ \\ \midrule
\multirow{3}{*}{\vspace{-6pt}~\textsc{jafar}} & \textsc{d-cusp} & 0.0099$_{0.0051}$ & 0.0037$_{0.0034}$ & 0.0087$_{0.0071}$ & 0.0083$_{0.0044}$ & 0.0114$_{0.0030}$ & 0.0079$_{0.0050}$ \\[3pt]
& \textsc{naive} & 0.0109$_{0.0050}$ & 0.0047$_{0.0043}$ & 0.0099$_{0.0089}$ & 0.0089$_{0.0047}$ & 0.0124$_{0.0037}$ & 0.0089$_{0.0063}$ \\[3pt]
& \textsc{full-d} & 0.0128$_{0.0051}$ & 0.0056$_{0.0039}$ & 0.0098$_{0.0038}$ & 0.0096$_{0.0040}$ & 0.0133$_{0.0032}$ & 0.0087$_{0.0027}$ \\
\bottomrule
\end{tabular}
\end{adjustbox}
%%%%%%%%%%%%%%%
\vspace{15pt}
\caption{Number of active factors (top), overall correlation matrix reconstruction errors (middle), and shared and specific component reconstruction errors (bottom) for different model–prior combinations in the simulations from Section~\ref{sec_sim_unsupervised}.
We assess matrix reconstruction error using the squared Frobenius norm of the difference between the inferred and true correlation structures, rescaled by the corresponding dimensions.
In the top panel, column $\Lambda_{tot}$ reports the total number of active factors in the shared component; columns $\Lambda_{m}$ show how many of those are active in each view; and column $\Lambda_{one-only}$ reports how many nominally shared factors are active in only one view.
Each entry is the mean over 100 replicate simulations, with subscripts denoting the corresponding standard deviations.
Both the \textsc{naive} and \textsc{full-d} priors fail to correctly separate view-specific factors from shared ones. Notably, the associated inflation of shared rank does not lead to more accurate reconstruction, not even of intra-view correlations.
By contrast, \textsc{jafar} under the \textsc{d-cusp} prior infers a pattern consistent with that of \textsc{fr} under the \textsc{i-cusp} prior, achieving the desired results.}
\label{tab_sim2_Nfact_Xcor}
\end{table}

% \vspace{-5pt}

\textcolor{teal}{
Table~\ref{tab_sim2_Nfact_Xcor} reports the inferred number of factors and covariance reconstruction errors, including means and standard deviations across replicates. Under the \textsc{naive} prior, \textsc{jafar} incorrectly assigns roughly $50\%$ of study-specific factors to the shared component. Under the \textsc{full-d} prior, \textsc{jafar} overestimates the shared rank even more severely, although technically avoiding misallocation of specific factors. Both approaches result in subpar accuracy of the learned correlations, including intra-view ones.
We assess this via the rescaled Frobenius norm of the difference between the true and inferred correlation matrices, as in equation~\eqref{eq_induced_cov}. 
Similar patterns are observed in the reconstructed shared and view-specific covariance components. By contrast, \textsc{jfr} under \textsc{i-cusp} and \textsc{jafar} under \textsc{d-cusp} produce self-consistent estimates of factor activity, both in overall rank and in the number of effectively shared factors.
}

\textcolor{teal}{
Interestingly, we observed a very similar pattern in real data. In Appendix~\ref{app_extra_sim_unsupervised}, we report the analysis of the inferred number of factors for the same methods on a random subset of features from the dataset analyzed later in Section~\ref{sec_application}.
These examples underscore the effectiveness of the \textsc{d-cusp} prior in achieving the intended separation of view-specific factors.
As such, we discard the \textsc{naive} and \textsc{full-d} priors in the remainder of the analysis, and implicitly assume the use of the \textsc{i-cusp} and \textsc{d-cusp} priors when referencing \textsc{jfr} and \textsc{jafar}, respectively.
}

\subsection{Predictive accuracy and dependence reconstruction in supervised settings}\label{sec_sim_supervised}

We considered 100 independent replicate datasets, each with
$M=3$ views of dimensions $p_m=\{100,200,300\}$, for increasing sample sizes $n \in \{50,100,150,200\}$ and fixed test set size $n_{test}=200$. 
The assumed number of shared factors and view-specific ones was $K^{(true)}=15$ and $\big\{ K_m^{(true)} \big\}_{m=1}^M = \{8,9,10\}$, with the responses loading on 9 of them in total. 
\textcolor{violet}{The upper bound on the number of factors was set to $K^\textsc{max}_{tot}=80$ in \textsc{jfr} and to $K^\textsc{max}=20$ and $ \{ K_m^\textsc{max} \}
_{m=1}^M = \{20,20,20\}$ for shared and specifics components of \textsc{jafar}.
For all $m=1,\dots,M$, we set $\alpha^{\scriptscriptstyle (\Lambda)}_m = \alpha^{\scriptscriptstyle (\Gamma)}_m =5$ in \textsc{jafar} and an equivalent $\alpha^{\scriptscriptstyle (L)}_m=20$ in \textsc{jfr}.}
We run the Gibbs samplers of \textsc{jafar} for a total of $T_{\textsc{mcmc}}=20000$ iterations, with a burn-in of $T_{\textsc{burn-in}}=15000$ steps and thinning every $T_\textsc{thin}=10$ samples for memory efficiency.
\textcolor{violet}{Given the significantly higher computing times, for \textsc{bip} and \textsc{bsfp} we set $T_{\textsc{mcmc}}=6000$ and $T_{\textsc{burn-in}}=3000$, keeping $T_\textsc{thin}=10$, and we run \textsc{bip} with $K^\textsc{max}=20$.}

\vspace{-5pt}
\begin{figure}[b!]
    \centering
{Response Predictions}\\
\includegraphics[width=0.45\linewidth]{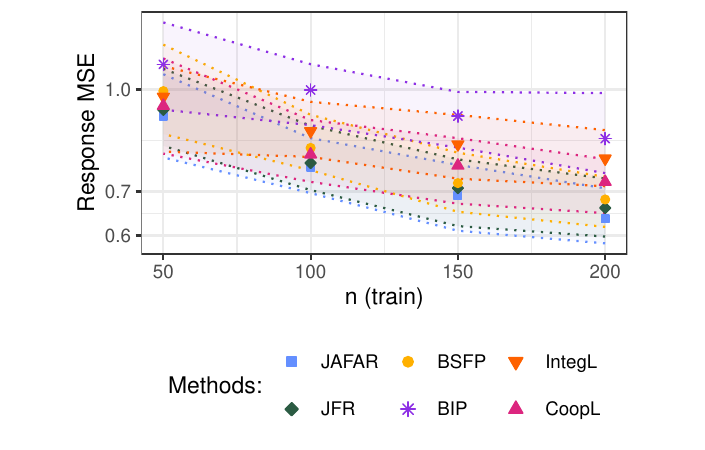}\includegraphics[width=0.45\linewidth]{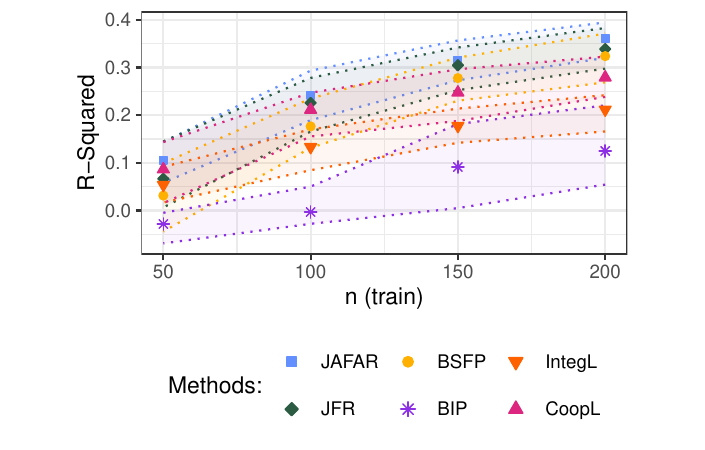} \\
{Correlation Matrices - Reconstruction Errors}\\
\includegraphics[width=\linewidth]{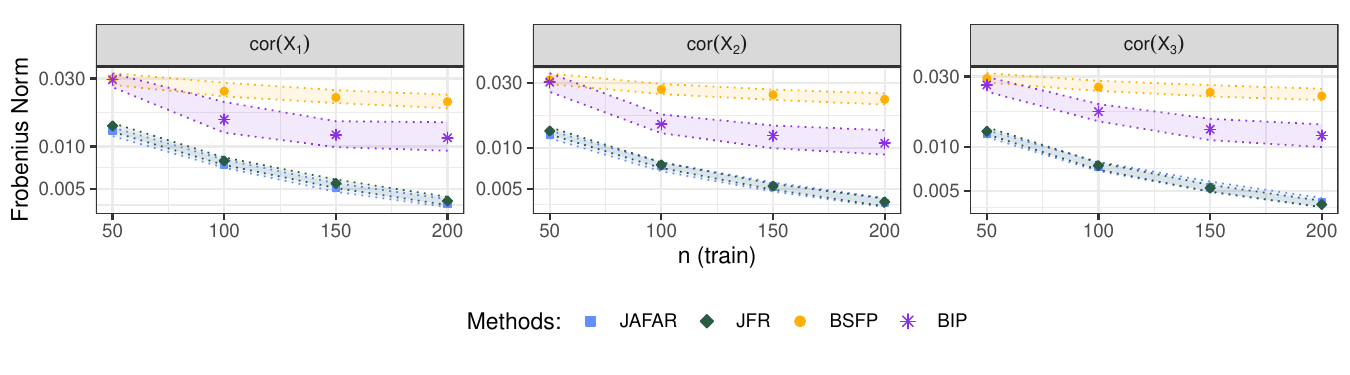}
    \vspace{-20pt}
    \caption{Out-of-sample predictive accuracy of the responses (above) and squared Frobenius norms of the differences between the true and inferred within-view correlations (below) for the simulations from Section~\ref{sec_sim_supervised}.
    All Frobenius norms are rescaled by the dimensions of the corresponding matrices.
    The x-axis values correspond to increasing training set sizes.
    Interior points and band edges correspond to quartiles across 100 independent replicates.
    }
    \label{fig_sim3_Ypred_Xcor}
\end{figure}
% \vspace{-5pt}

\textcolor{violet}{Figure~\ref{fig_sim3_Ypred_Xcor} shows that, in this setting, \textsc{jfr} and \textsc{jafar} achieve competitive or superior predictive accuracy compared to the other methods, crucially associated with an accurate reconstruction of the dependence structure underlying the data. \textsc{bsfp} achieves good predictive performance but suffers from severe overshrinking of the learned covariance. 
\textsc{bip} comparatively improves covariance reconstruction -- still worse than \textsc{jfr} and \textsc{jafar} -- but attains sub-par predictive performance.
\textsc{jfr} and \textsc{jafar} also achieve better coverage than competitors, as depicted in Figure~\ref{fig_sim_response_CI} in Appendix~\ref{app_sim_extra_ris}.
Therein, we also provide further insight into correlation reconstruction in an exemplar replication.}

\textcolor{teal}{
As expected, \textsc{jfr} and \textsc{jafar} provide essentially interchangeable representations. 
However, \textsc{jfr} consistently requires roughly 55\% more runtime than \textsc{jafar}, reflecting the computational benefits of exploiting structured sparsity.
In Appendix~\ref{app_extra_sim_supervised}, Table~\ref{tab_sec3_times} reports detailed runtimes, while Table~\ref{tab_sec3_ess} summarizes the effective sample size (\textsc{ess}) for \textsc{jfr}, \textsc{jafar}, and \textsc{bsfp} on selected summary statistics. We could not perform a similar analysis for \textsc{bip}, as available code only produces point-wise summaries.
For view reconstruction, \textsc{jafar} and \textsc{bsfp} achieve good \textsc{ess}, whereas \textsc{jfr} shows slightly worse performance as $n$ increases. For the response, \textsc{jfr} performs moderately better than \textsc{jafar}, although the performance gap narrows with larger 
$n$. \textsc{bsfp} consistently underperforms, with the gap widening as 
$n$ grows.
}

%%%@@@@@@@@@@@@@@@@@@@@@@@@@@@@@@@@@@@@@@@@@@@@@@@@
%%              Empirical studies
%%%@@@@@@@@@@@@@@@@@@@@@@@@@@@@@@@@@@@@@@@@@@@@@@@@

\section{Labor onset prediction from immunome, metabolome \& proteome}\label{sec_application}

To further showcase the performance of the proposed methodologies, we focus on predicting time-to-labor onset from immunome, metabolome, and proteome data for a cohort of women who went into labor spontaneously.
The dataset, available at \citet{mallick2024integrated}, considers repeated measurements during the last 100 days of pregnancy. Similar to \citet{Ding_2022_coopL}, we obtained a cross-sectional sub-dataset by considering only the first measurement for each woman included in the study, and we treat time-to-labor onset as a continuous outcome. 
The resulting dataset falls into a rather extreme large-$p$-small-$n$ scenario, as the $M=3$ layers of blood measurements provide information on $p_1=1141$ single-cell immune features, $p_2=3529$ metabolites, and $p_3=1317$ proteins \textcolor{violet}{for $n_{tot}=53$ subjects}.
\textcolor{teal}{Data preprocessing is described in Web Appendix~H.}
\textcolor{teal}{To better assess the performance and generalizability of the considered methods, we perform five random splits of the data (two of size 10 and three of size 11), sequentially using each split as the test set while training on the remaining subjects.}

% \vspace{-5pt}
\begin{table}[hb!]
\centering
{Inferred Number of Active Factors}\\[5pt]
\begin{adjustbox}{width=1\textwidth}
\begin{tabular}{c|rrr|r|rrr|r
}
\toprule
\multicolumn{1}{c|}{} & \multicolumn{3}{c|}{\small{Specific Components}} & \multicolumn{5}{c}{\small{Shared Component}} \\
    %%%%%%%%%%%%%%%%%
 & \multicolumn{1}{c}{\small{Immunome}} & \multicolumn{1}{c}{\small{Metabolome}} & \multicolumn{1}{c|}{\small{Proteome}} & \multicolumn{1}{c|}{\small{Total}} & \multicolumn{1}{c}{\small{Immunome}} & \multicolumn{1}{c}{\small{Metabolome}} & \multicolumn{1}{c|}{\small{Proteome}} &  \multicolumn{1}{c}{\small{Only one}} \\ 
 \cmidrule{1-9}
\textsc{bsfp} & 12.6$_{0.9}$ & 14.4$_{0.5}$ & 11.0$_{1.2}$ & 10.2$_{0.4}$ & -\hspace{15pt} & -\hspace{15pt} & -\hspace{15pt} & -\hspace{15pt} \\[2pt]
\textsc{jafar} & 17.2$_{0.8}$ & 24.8$_{1.6}$ & 17.5$_{1.1}$ & 15.2$_{2.4}$ & 5.6$_{2.0}$ & 11.5$_{2.8}$ & 11.5$_{2.0}$ & 1.7$_{0.9}$ \\[2pt]
\textsc{jafar}$_T$& 10.0$_{0.9}$ & 8.6$_{1.5}$ & 8.3$_{1.3}$ & 9.0$_{3.0}$ & 5.0$_{1.9}$ & 7.6$_{2.5}$ & 5.9$_{2.3}$ & 0.2$_{0.2}$ \\[2pt]
\textsc{jfr} & -\hspace{15pt} & -\hspace{15pt} & -\hspace{15pt} & 78.8$_{1.5}$ & 21.8$_{1.1}$ & 34.8$_{1.9}$ & 35.7$_{1.8}$ & 62.3$_{2.9}$ \\[2pt]
\textsc{jfr}$_T$ & -\hspace{15pt} & -\hspace{15pt} & -\hspace{15pt} & 39.0$_{6.7}$ & 18.7$_{2.1}$ & 29.3$_{3.3}$ & 25.4$_{5.3}$ & 13.8$_{3.4}$ \\
\bottomrule
\end{tabular}
\end{adjustbox}
\vspace{5pt}
\caption{Inferred numbers of active factors in the time-to-labor application from Section~\ref{sec_application}, aggregated over five random data splits (mean and standard deviation).
% The $M=3$ omics layers correspond to immunome, metabolome, and proteome data.
In \textsc{bsfp}, the ranks are fixed via the \textsc{unifac} initialization, whereas for all other methods the table reports posterior means of the number of active columns, computed from the latent indicators in the \textsc{cusp} constructions.
The base versions of \textsc{jfr} and \textsc{jafar} infer a relatively large number of factors -- overall exceeding the sample size ($n=42$ or $43$, depending on the data split) -- though still small compared to the views’ dimensions ($1141$, $3529$, and $1317$).
Nevertheless, \textsc{jfr} and \textsc{jafar} produce mutually consistent representations, with roughly 60 factors active in only one component and about 15 effectively shared.
The tempered versions, \textsc{jfr}$_T$ and \textsc{jafar}$_T$, successfully reduce the number of factors while retaining a fully adaptive Bayesian framework.}
\label{tab_sec4_N_active_factors}
\end{table}
% \vspace{-5pt}

As before, we compare \textcolor{violet}{\textsc{jfr} and} \textsc{jafar} to \textsc{bsfp}, \texttt{CoopLearn}, and \texttt{IntegLearn}, with the same hyperparameters from the previous section.
\textcolor{teal}{We also include the tempered versions \textsc{jfr}$_T$ and \textsc{jafar}$_T$ introduced in Section~\ref{sec_curse_dim_tempering}.}
\textcolor{violet}{We do not consider \textsc{bip} for its poor predictive performance shown in Section~\ref{sec_simulations} and its computational inefficiency.}
% Prior to analysis, we standardized the data and log-transformed the metabolomics and proteomics features. Despite these preprocessing steps, all omics layers exhibited considerable deviation from Gaussianity, with over $30\%$ of features in each view yielding univariate Shapiro test statistics below $0.95$. To address this challenge, we target copula factor model variants \citep{Murray_2013_Copula_FA} for \textsc{jfr}, \textsc{jafar}, and \textsc{bsfp}, as detailed in Appendix~\ref{app_tempering}. Given the continuous nature of the omics data and the absence of missing entries, the incorporation of the copula layer boils down to a deterministic preprocessing procedure, involving feature-wise transformations that leverage estimates of the associated empirical cumulative distribution functions.
\textcolor{violet}{The upper bound on the number of factors was set to $K^\textsc{max}_{tot}=85$ in \textsc{jfr} and to $K^\textsc{max}=25$ and $ \{ K_m^\textsc{max} \}
_{m=1}^M = \{25,35,25\}$ for shared and specifics components of \textsc{jafar}.
For all $m=1,\dots,M$, we set $\alpha^{\scriptscriptstyle (\Lambda)}_m = \alpha^{\scriptscriptstyle (\Gamma)}_m =5$ in \textsc{jafar} and an equivalent $\alpha^{\scriptscriptstyle (L)}_m=20$ in \textsc{jfr}.
We run the Gibbs sampler of both \textsc{jfr} and \textsc{jafar} for a total of $T_{\textsc{mcmc}}=20000$ iterations, with a burn-in of $T_{\textsc{burn-in}}=15000$ steps and $T_\textsc{thin}=10$.}
Given the higher computing times, for \textsc{bsfp} we set $T_{\textsc{mcmc}}=10000$ iterations, with a burn-in of $T_{\textsc{burn-in}}=5000$, keeping $T_\textsc{thin}=10$.

% \vspace{-7pt}
\begin{figure}[hb!]
    \centering
    {Prediction of Time-to-labor Onset}\\[15pt]
    \includegraphics[width=0.49\linewidth]{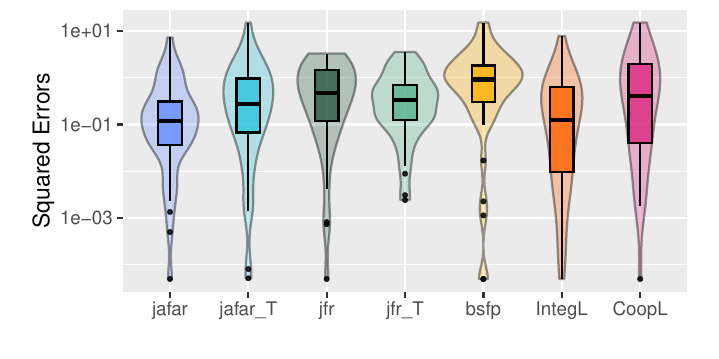}
    \put(-100,122){\makebox(0,0){{Train Set}}}
    \includegraphics[width=0.49\linewidth]{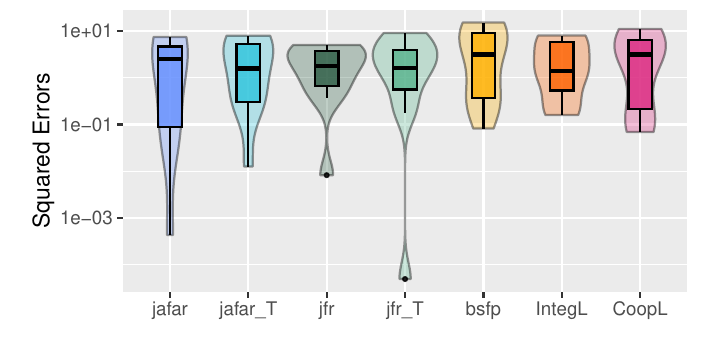} 
    \put(-100,122){\makebox(0,0){{Test Set}}}
    \\[5pt]
{Test Set - Aggregated Metrics Over Random Data Splits}\\[5pt]
\begin{tabular}{c|rrrrrrr}
\toprule
 &  jafar & jafar$_T$ & jfr & jfr$_T$ & bsfp & IntegL & CoopL \\
\cmidrule{1-8}
MSE & 3.21 & 3.24 & 3.23 & 2.96 & 4.46 & 4.24 & 3.8 \\
$R^2$ & 0.53 & 0.52 & 0.48 & 0.53 & 0.33 & 0.34 & 0.43 \\
90\% C.I. & 92.73 & 92.73 & 96.36 & 92.73 & 96.36 & 98.18 & -
\hspace{7pt} \\
\bottomrule
\end{tabular}
\vspace{5pt}
\caption{Prediction accuracy for time-to-labor (days) in the data application from Section~\ref{sec_application}.
The top panel shows box plots of the prediction squared errors for an exemplar random data split, both in-sample (left) and out-of-sample (right).
The bottom panel reports the mean of different accuracy metrics over five random splits, including mean squared error (MSE), coefficient of determination ($R^2$), and empirical coverage of the $90\%$ confidence intervals (90\% C.I.).
Both \texttt{CoopLearn} and \texttt{IntegLearn} achieve good predictive performance on the training set but generalize less well to out-of-sample observations, while \textsc{bsfp} performs worse in both cases.
Tempering does not reduce the predictive accuracy of \textsc{jfr} and \textsc{jafar}, while it partially improves coverage by producing narrower credible intervals.}
\label{fig_response_prediction}
\end{figure}
% \vspace{-7pt}

\textcolor{teal}{
Both \textsc{jfr} and \textsc{jafar} encounter the curse of dimensionality described in Section~\ref{sec_curse_dim_tempering}.
The inferred overall ranks reported in Table~\ref{tab_sec4_N_active_factors} are small compared to the view dimensions, but greater than the sample size.
\textsc{jfr}$_T$ and \textsc{jafar}$_T$ achieved the desired objective, flexibly learning substantially fewer factors than the untempered versions, comparable to \textsc{bsfp}.
Interestingly, the rank reduction induced by the tempering is associated with better accuracy in covariance reconstruction and no systematic loss of prediction power.
Figure~\ref{fig_response_prediction} summarizes the relative accuracy in predicting time-to-labor, where the proposed methodologies achieve superior performance across the five random data splits.
All procedures exhibit some degree of overcoverage.
This issue is less severe for \textsc{jafar}, both with and without tempering, and for the tempered version of \textsc{jfr}. 
Tempering appears to reduce the estimation noise caused by the high number of latent factors (relative to $n$), leading to narrower and better-calibrated 90\% credible intervals.
}

% \newpage

% \vspace{-5pt}
\begin{figure}[htbp!]
   \centering
   \begin{minipage}{0.29\textwidth}
       \centering
       \includegraphics[width=\linewidth]{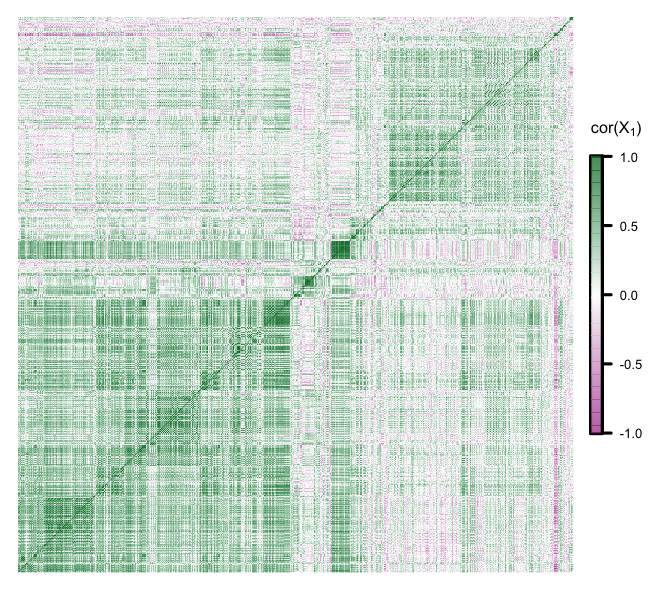}
   \put(-75,130){\makebox(0,0){Immunome}}
   \put(-145,60){\rotatebox{90}{\makebox(0,0){Empirical}}}
   \end{minipage}
   \begin{minipage}{0.29\textwidth}
       \includegraphics[width=\linewidth]{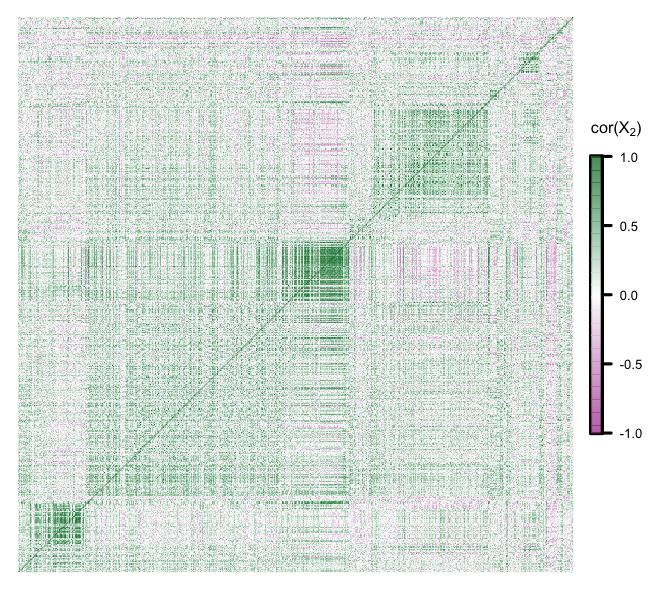}
   \put(-75,130){\makebox(0,0){Metabolome}}
   \end{minipage}
   \begin{minipage}{0.29\textwidth}
       \centering
       \includegraphics[width=\linewidth]{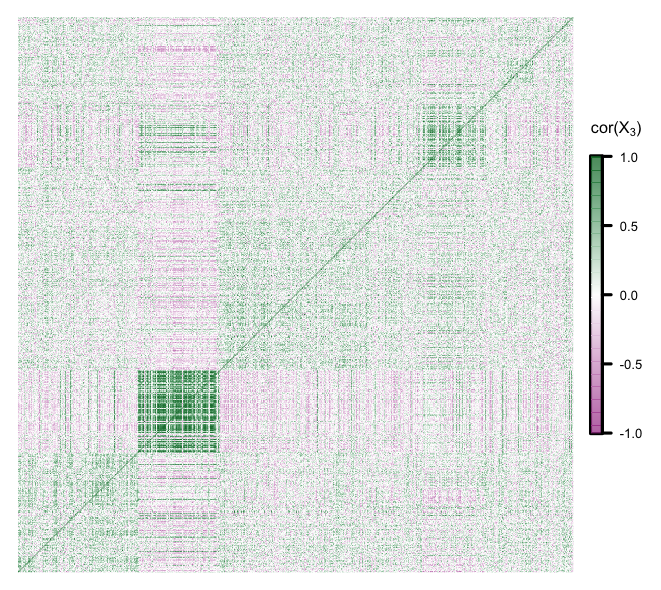}
   \put(-75,130){\makebox(0,0){Proteome}}
   \end{minipage}
   \\
   % % % % % % % % % % % % % % % 
   \begin{minipage}{0.29\textwidth}
       \centering
       \includegraphics[width=\linewidth]{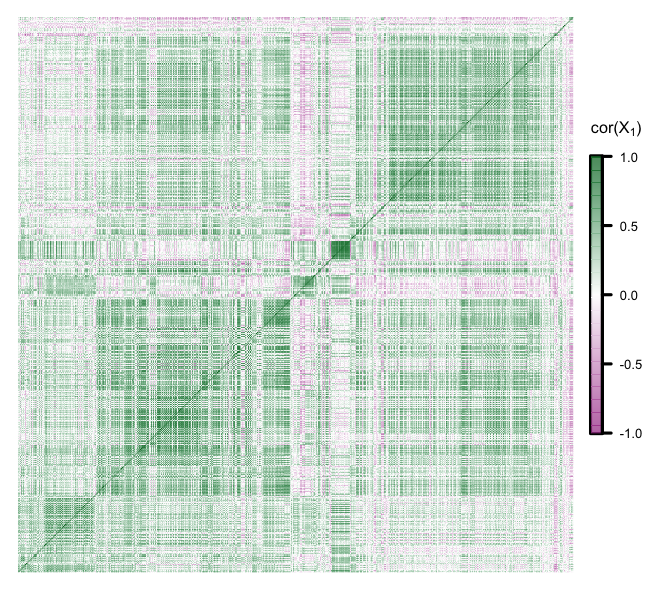}
   \put(-145,60){\rotatebox{90}{\makebox(0,0){\textsc{jfr}$_T$}}}
   \end{minipage}
   \begin{minipage}{0.29\textwidth}
       \centering
       \includegraphics[width=\linewidth]{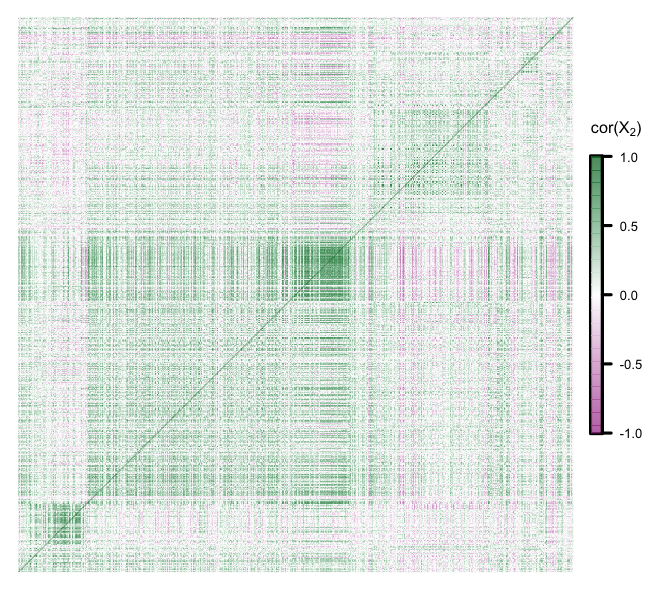}
   \end{minipage}
   \begin{minipage}{0.29\textwidth}
       \centering
       \includegraphics[width=\linewidth]{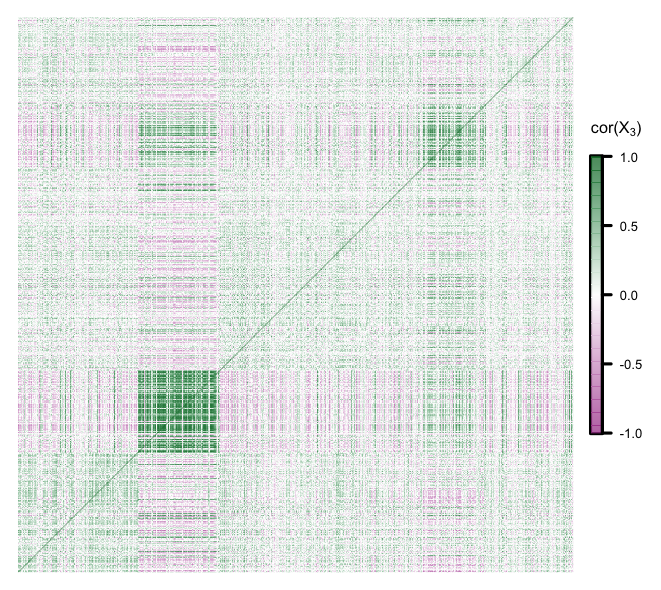}
   \end{minipage}
   \\
   % % % % % % % % % % % % % % % 
   \begin{minipage}{0.29\textwidth}
       \centering
       \includegraphics[width=\linewidth]{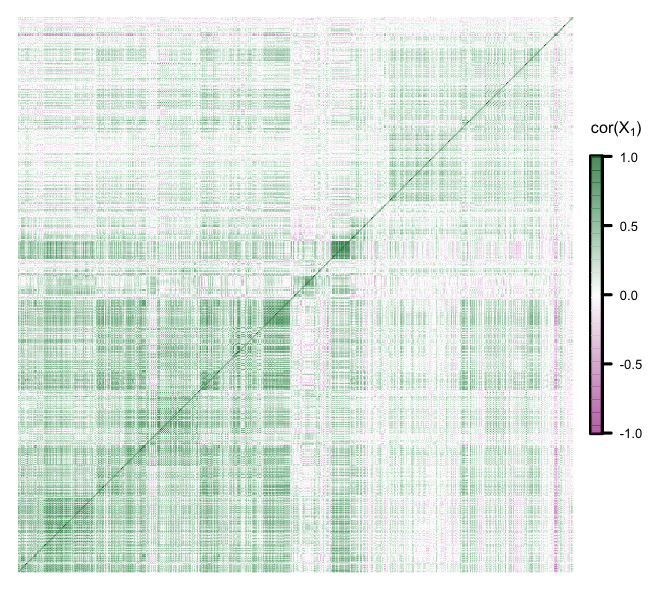}
   \put(-145,60){\rotatebox{90}{\makebox(0,0){\textsc{jafar}$_T$}}}
   \end{minipage}
   \begin{minipage}{0.29\textwidth}
       \centering
       \includegraphics[width=\linewidth]{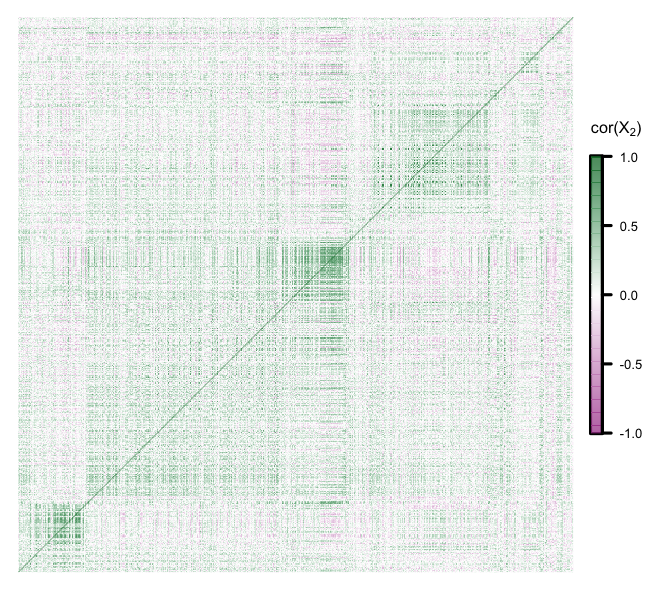}
   \end{minipage}
   \begin{minipage}{0.29\textwidth}
       \centering
       \includegraphics[width=\linewidth]{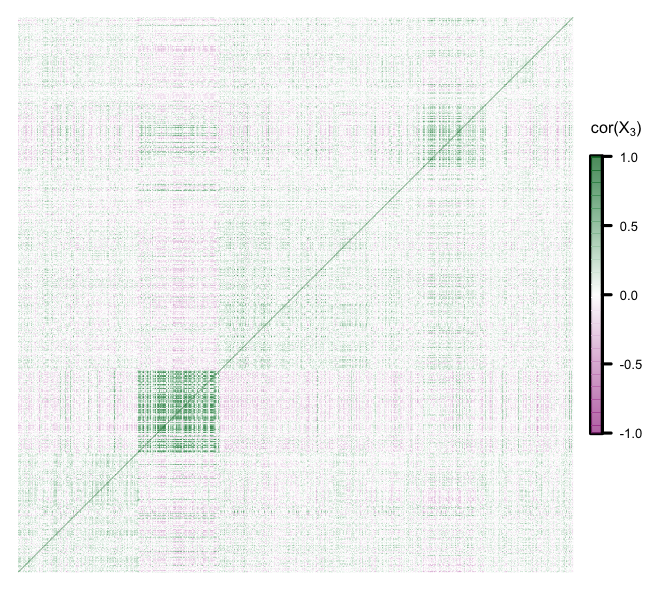}
   \end{minipage}
   \\
   % % % % % % % % % % % % % % % 
   \begin{minipage}{0.29\textwidth}
       \centering
       \includegraphics[width=\linewidth]{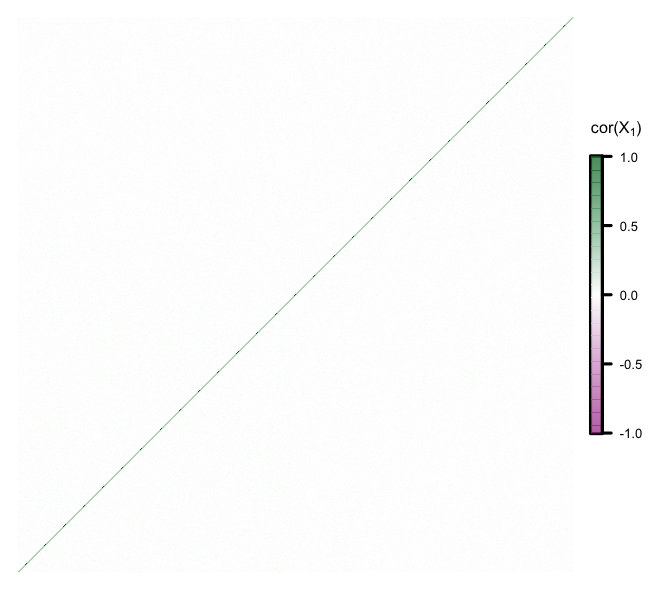}
   \put(-145,60){\rotatebox{90}{\makebox(0,0){\textsc{bsfp}}}}
   \end{minipage}
   \begin{minipage}{0.29\textwidth}
       \centering
       \includegraphics[width=\linewidth]{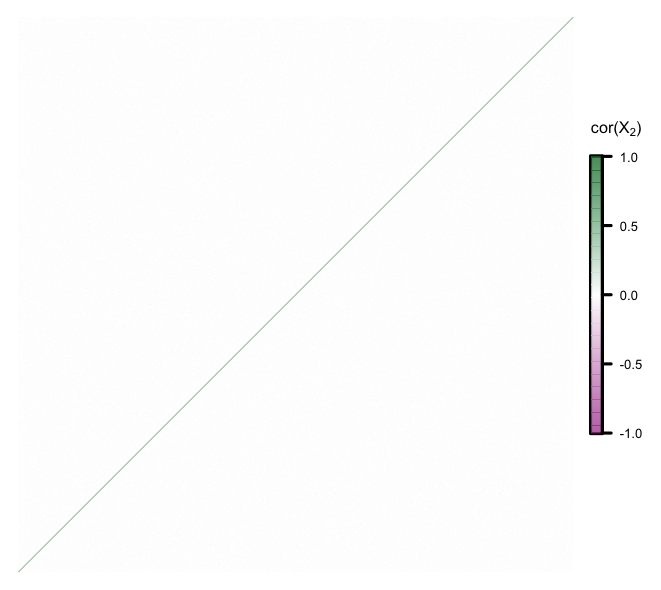}
   \end{minipage}
   \begin{minipage}{0.29\textwidth}
       \centering
       \includegraphics[width=\linewidth]{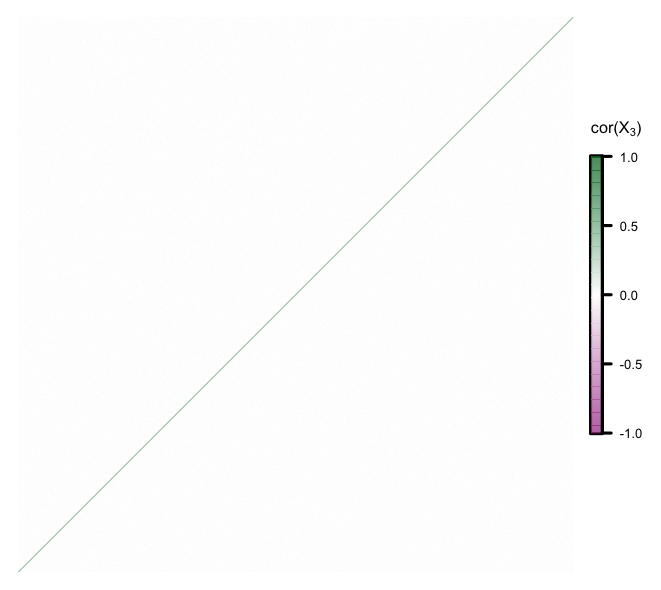}
   \end{minipage}
   \\[5pt]
   %%%%%%%%%%%%%%%%%%%%%%%%%%%%%%%
   \begin{tabular}{c|ccc|ccc
}
\toprule
 & $cor(X_1)$ & $cor(X_2)$ & $cor(X_3)$ & $cor(X_1,X_2)$ & $cor(X_1,X_3)$ & $cor(X_2,X_3)$ \\ \midrule
\textsc{jfr} & 0.0132$_{0.0026}$ & 0.0110$_{0.0007}$ & 0.0124$_{0.0018}$ & 0.0225$_{0.0005}$ & 0.0205$_{0.0008}$ & 0.0217$_{0.0003}$ \\
\textsc{jfr}$_T$ & 0.0190$_{0.0040}$ & 0.0116$_{0.0035}$ & 0.0172$_{0.0048}$ & 0.0205$_{0.0009}$ & 0.0189$_{0.0005}$ & 0.0201$_{0.0003}$ \\
\textsc{jafar} & 0.0139$_{0.0005}$ & 0.0163$_{0.0016}$ & 0.0146$_{0.0011}$ & 0.0244$_{0.0008}$ & 0.0226$_{0.0007}$ & 0.0234$_{0.0004}$ \\
\textsc{jafar}$_T$ & 0.0117$_{0.0011}$ & 0.0135$_{0.0008}$ & 0.0133$_{0.0007}$ & 0.0226$_{0.0010}$ & 0.0229$_{0.0007}$ & 0.0219$_{0.0009}$ \\
\textsc{bsfp} & 0.0695$_{0.0045}$ & 0.0520$_{0.0018}$ & 0.0418$_{0.0011}$ & -\hspace{15pt} & -\hspace{15pt} & -\hspace{15pt} \\
\bottomrule
\end{tabular}
   \caption{Correlation reconstruction for the three omics layers from Section~\ref{sec_application}, shown for one exemplar data split (top) and summarized across all splits with mean and standard deviation of the rescaled Frobenius norms of the differences from the empirical correlations (bottom).}
   \label{fig_inferred_cor}
\end{figure}
% \vspace{-5pt}  

\textcolor{teal}{
\textsc{bsfp} achieves substandard performance, both in response prediction and covariance reconstruction.
In Figure~\ref{fig_inferred_cor}, we report the empirical and inferred within-view correlation matrices for a subset of the methods, with Figure~\ref{sec4_inferred_cor_extra} in the Appendix giving the full picture.
\textsc{jfr} and \textsc{jfr}$_T$ seem to partially overestimate some correlations, as opposed to the slight underestimation in \textsc{jafar} and \textsc{jafar}$_T$.
Overall, such methods achieve comparable performances, relative to the extreme over-shrinkage suffered by \textsc{bsfp}.
}

\textcolor{teal}{
The main difference between \textsc{jfr} and \textsc{jafar}, or \textsc{jfr}$_T$ and \textsc{jafar}$_T$ seems to lie in their relative efficiency and ease of interpretability.
On the same machine used for the simulation studies, the median runtimes across splits were $426$, $146$, $134$, and $55$ minutes for \textsc{jfr}, \textsc{jfr}$_T$, \textsc{jafar}, and \textsc{jafar}$_T$, respectively. For comparison, \textsc{bsfp} required a median of $434$ minutes -- comparable to \textsc{jfr} but for half as many total \textsc{mcmc} samples, roughly half as many factors, and despite the sparser structure. 
Regarding interpretability, in Appendix~\ref{app_stelzer_extra_ris} we report the posterior means of the shared loading matrices after postprocessing with the extended \texttt{MatchAlign} algorithm using Multiview \texttt{Varimax}.
}

%%%@@@@@@@@@@@@@@@@@@@@@@@@@@@@@@@@@@@@@@@@@@@@@@@@
%%                  Discussion
%%%@@@@@@@@@@@@@@@@@@@@@@@@@@@@@@@@@@@@@@@@@@@@@@@@
\section{Discussion}

\textcolor{violet}{
We developed a fully Bayesian joint factor regression (\textsc{jfr}) model capturing combined variation across multiple views via global latent factors, and a more interpretable, computationally efficient additive version (\textsc{jafar}) that decomposes variation into shared and view-specific components. By isolating these factors, \textsc{jafar} facilitates inference, prediction, and feature selection. To ensure identifiability of shared variation in \textsc{jafar}, we introduce a novel extension of the \textsc{cusp} prior \citep{Legramanti_2020_CUSP}. We further adapt the \texttt{Varimax} procedure to multiview settings, preserving the composite structure while resolving rotational ambiguity.
}

\textcolor{violet}{\textsc{jfr} and} \textsc{jafar}'s performances are compared to state-of-the-art competitors using multiview simulated data and in an application focusing on predicting time-to-labor onset from multiview features derived from immunomes, metabolomes, and proteomes. 
The carefully designed structures enable accurate learning and inference of response-related latent factors, as well as the inter- and intra-view correlation structures. 
\textcolor{teal}{In practice, the latent representations inferred by \textsc{jfr} and \textsc{jafar} are essentially equivalent in terms of accuracy. The computational advantages and greater interpretability of \textsc{jafar} make it the preferred choice, particularly in high-dimensional settings.
}

The benefit of the proposed \textsc{d-cusp} prior extends to unsupervised scenarios, where the focus is solely on disentangling the sources of variability within integrated multimodal data. 
\textcolor{teal}{Our code provides implementations of both supervised and unsupervised versions of the proposed methodologies.}
In Appendix~\ref{app_tempering}, we discuss more flexible response modeling through interactions among latent factors \citep{Ferrari_2021_Interactions} and splines, while considering extensions akin to generalized linear models. 
Lastly, analogous constructions can be readily developed using the structured increasing shrinkage prior proposed by \citet{Schiavon_2022_GIF_SIS}, allowing for the inclusion of prior annotation data on features.

% \vspace{-5pt}

%%%@@@@@@@@@@@@@@@@@@@@@@@@@@@@@@@@@@@@@@@@@@@@@@@@
%%                  Acknowledgements
%%%@@@@@@@@@@@@@@@@@@@@@@@@@@@@@@@@@@@@@@@@@@@@@@@@
\section*{Acknowledgements}

This project has received funding from the European Research Council (ERC) under the European Union’s Horizon 2020 research and innovation program (grant agreement No 856506), 
and United States National Institutes of Health (R01ES035625, 5R01ES027498-05, 5R01AI167850-03), and was supported in part by Merck \& Co., Inc., through its support for the Merck Biostatistics and Research Decision Sciences (BARDS) Academic Collaboration.

%%%@@@@@@@@@@@@@@@@@@@@@@@@@@@@@@@@@@@@@@@@@@@@@@@@
%%                  Bibliography
%%%@@@@@@@@@@@@@@@@@@@@@@@@@@@@@@@@@@@@@@@@@@@@@@@@
\bibliographystyle{agsm}
\bibliography{ref} 

%%%@@@@@@@@@@@@@@@@@@@@@@@@@@@@@@@@@@@@@@@@@@@@@@@@
%%                  Appendix
%%%@@@@@@@@@@@@@@@@@@@@@@@@@@@@@@@@@@@@@@@@@@@@@@@@

\newpage

\appendix

\titleformat{\section}[block]
  {\normalfont\large\bfseries}
  {Appendix \thesection}{1em}{}

\renewcommand{\thesection}{\Alph{section}}
\renewcommand{\thetable}{\thesection\arabic{table}}
\renewcommand{\thefigure}{\thesection\arabic{figure}}

%%%%%%%%%%%%%%%%%%%%%%%%%%%%%%%%%%%
\section{Hyperparameters in \textsc{jfr} and \textsc{jafar}}\label{app_cusp_hyper_param}

\setcounter{table}{0}
\setcounter{figure}{0}
\setcounter{equation}{0}
\setcounter{algocf}{0}

In this section, we give a detailed discussion of the choice of the hyperparameters of the \textsc{i-cusp} and \textsc{d-cusp} priors.
As by equation~(6), the suggested default values are 
{
\begin{equation*}\label{app_eq_hyperparams}
% \mycustomsize %,\small
\setlength{\arraycolsep}{10pt}
\renewcommand{\arraystretch}{1.2}
% \begin{array}{cccc}
\begin{array}{clll}
\text{( Data distributions )} &  a_m^{\scriptscriptstyle{(}\scriptstyle{\sigma}\scriptscriptstyle{)}} =
a_y^{\scriptscriptstyle{(}\scriptstyle{\sigma}\scriptscriptstyle{)}} = 3 & b_m^{\scriptscriptstyle{(}\scriptstyle{\sigma}\scriptscriptstyle{)}} =
b_y^{\scriptscriptstyle{(}\scriptstyle{\sigma}\scriptscriptstyle{)}} = 1 & 
\upsilon_m^2 = \upsilon_y^2 = 0.25 \\ 
\text{( Spike \& slab variances )} & 
a^{\scriptscriptstyle (L)}_m = a^{\scriptscriptstyle (\theta)} = 0.5 & 
b^{\scriptscriptstyle (L)}_m = b^{\scriptscriptstyle (\theta)} = 0.1 & 
\tau^2_{m \, \infty} = \chi^2_{\infty } = 0.005 \\ 
\text{( Spike \& slab weights )} & 
a^{(\xi)} = 3 & 
b^{(\xi)} = 2 & 
\textcolor{gray}{\alpha^{\scriptscriptstyle (L)}_m,
\alpha^{\scriptscriptstyle (\Lambda)}_m, \alpha^{\scriptscriptstyle (\Gamma)}_m} \; . \\ 
\end{array}
\end{equation*}
}
We refer to \citet{Legramanti_2020_CUSP} for an introduction to \textsc{CUSP}.
As we state in the main paper, the values above have proven effective in a wide range of settings, including both simulated and real data. In practice, we recommend adjusting only $\alpha^{\scriptscriptstyle (L)}_m$,
$\alpha^{\scriptscriptstyle (\Lambda)}_m$, $\alpha^{\scriptscriptstyle (\Gamma)}_m$ to guide the prior expectation of the number of factors. 

\begin{figure}[h!]
\centering
\includegraphics[width=0.5\linewidth]{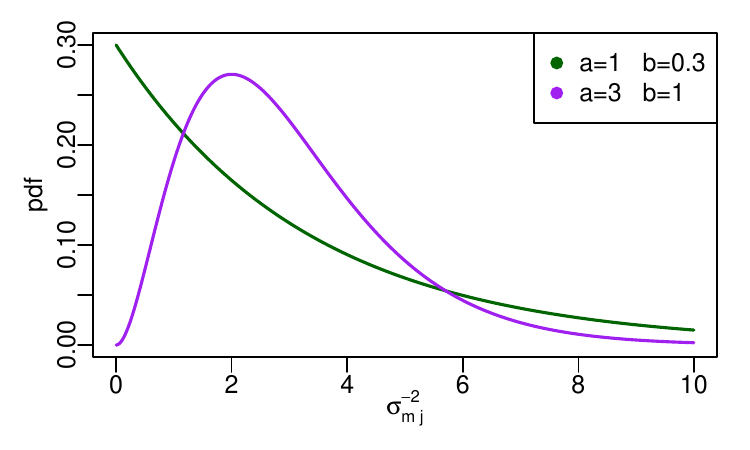}
\vspace{-5pt}
   \caption{PDFs of the noise precision induced different choices of the hyperparameters $a_m^{\scriptscriptstyle{(}\scriptstyle{\sigma}\scriptscriptstyle{)}}$ and $b_m^{\scriptscriptstyle{(}\scriptstyle{\sigma}\scriptscriptstyle{)}}$.
   The green line corresponds to the choice in \citet{Legramanti_2020_CUSP}, while the violet is obtained under the value we suggest using.}
\label{fig_HP_prec_noise}
\end{figure}

The choice of the prior parameters $\bsigma_m^2$ is meant to avoid very large values for the variance of idiosyncratic components -- or equivalently, avoid very small precisions -- as opposed to the original choice $a_m^{\scriptscriptstyle{(}\scriptstyle{\sigma}\scriptscriptstyle{)}} = 1$ and $ b_m^{\scriptscriptstyle{(}\scriptstyle{\sigma}\scriptscriptstyle{)}} = 0.3$.
In Figure~\ref{fig_HP_prec_noise}, we plot the PDFs of the precision induced by these two choices on the Gaussian distribution of the noise elements in \textsc{jfr} and \textsc{jafar}.
The original setup induces a high probability of very small precisions.
Conversely, our choice encourages the model to explain the variability observed in the data through the signal components rather than via noise, while still allowing large variances when needed. The same rationale applies to the response component for $\sigma_y^2$.

% \vspace{-15pt}

% \vspace{-5pt}

The precision $\upsilon_m^2$ and $\upsilon_y^2$ on the intercepts are chosen to be weakly informative, although a much smaller value would also be legitimate, given that the data have been standardized.
Keeping a relatively flexible intercept could be useful in the copula versions of the factor models.
The choice of $a^{(\xi)}$ and $b^{(\xi)}$ reflects a slight prior preference for the active entries $\btheta_h$ in the response loadings, rather than inactive -- without increasing shrinkage on $h$.

\begin{figure}[ht!]
\centering
\includegraphics[width=0.45\linewidth]{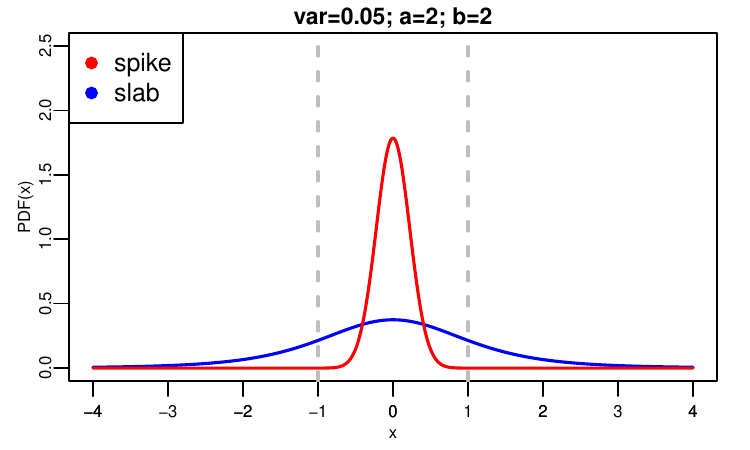}
\includegraphics[width=0.45\linewidth]{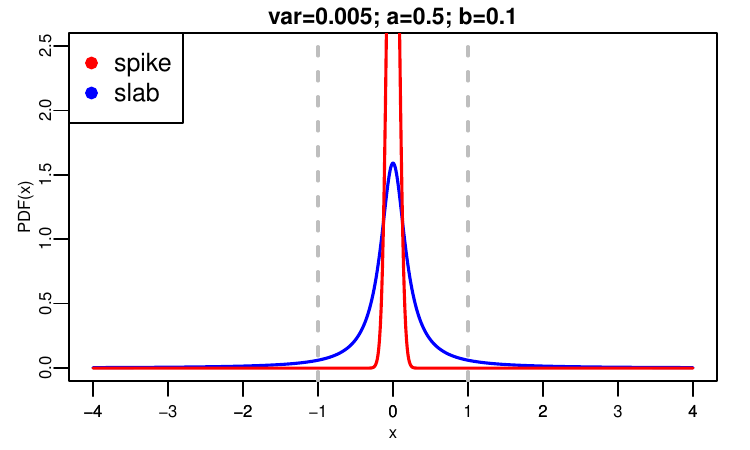}
\vspace{-5pt}
   \caption{PDFs of the spike and slab priors on the loadings component in the \textsc{cusp} construction, induced different choices of the hyperparameters $a^{\scriptscriptstyle (L)}_m$, $b^{\scriptscriptstyle (L)}_m$, and $\tau^2_{m \, \infty} $.
   The left panel corresponds to the parameters suggested in \citet{Legramanti_2020_CUSP}, the right one corresponds to the choice we suggest in this manuscript.}
\label{fig_HP_spike_slab}
\end{figure}

The hyperparameters $a^{\scriptscriptstyle (L)}_m$, $b^{\scriptscriptstyle (L)}_m$, and $\tau^2_{m \, \infty} $ of the spike-and-slab distributions
target a scaled version of the original one in \citet{Legramanti_2020_CUSP}, where 
the author set $\tau^2_{m \, \infty} = 0.05$ and $a^{\scriptscriptstyle (L)}_m = b^{\scriptscriptstyle (L)}_m = 2$.
Notice that, marginalizing out the slab hypervariance, the induced slab component is a Student-t distribution with PDF  
$t_{2 \, a^{\scriptscriptstyle(L)}_m} (\,\cdot\; ;\, 0, b^{\scriptscriptstyle (L)}_m / a^{\scriptscriptstyle (L)}_m) $, plotted in Figure~\ref{fig_HP_spike_slab} alongside the spike's PDF for both parameter setups.
The original setup does work well when the focus is solely on adaptively learning the number of factors, leveraging relative probabilities of coming from the spike or from the slab in the \textsc{cusp} construction.
However, the specific structure we are targeting calls for two corrections.

% \vspace{-15pt}

First, each feature was scaled to have unit variance.
Although we avoid imposing hard constraints on the model to keep the \textsc{mcmc} design simple, it is desirable that the diagonal elements of the induced covariance concentrate around 1.
The latter reads $\sum_{h=1}^K (\bLambda_{m j h})^2 + \sum_{m=1}^{M} \sum_{h=1}^{K_m} (\bGamma_{m j h})^2 + \sigma_{mj}^2$ in \textsc{jafar}.
A minimal requirement to achieve this is that the slab component assigns a low probability
on each $\bLambda_{m j h}$ and $\bGamma_{m j h}$ lying outside of $[-1,1]$.
The prior probability of lying outside such an interval is $\approx 43\%$ under the choice in \citet{Legramanti_2020_CUSP}, while only $\approx 12\%$ under the values proposed here.
Modifying the slab requires modifying the spike accordingly to preserve the right balance.
Secondly, setting $\tau^2_{m \, \infty} = 0.05$ as in \citet{Legramanti_2020_CUSP} still allows the spike to capture a decent amount of signal.
This is suboptimal in our setting, as we rely on the inactivity of any given factor to effectively capture the spuriousness of given latent directions in some data component.
This is crucial for \textsc{jafar}, but important as well in \textsc{jfr}.
The smaller spike variance $\tau^2_{m \, \infty} = 0.005$ is better suited for this.

Our choice above roughly corresponds to reducing the variance of both spike and slab by a factor of 10 compared to the choice in \citet{Legramanti_2020_CUSP}. 
Strictly speaking, this would be obtained via $a^{\scriptscriptstyle (L)}_m = 2$ and $b^{\scriptscriptstyle (L)}_m = 0.2$, but we found that setting $a^{\scriptscriptstyle (L)}_m = 0.5$ and $b^{\scriptscriptstyle (L)}_m = 0.1$ works better in practice, particularly in providing a clear separation of active and inactive elements, while inducing a similar scale.
As before, the same rationale applies to the response component.

We use the same hyperparameters $a^{\scriptscriptstyle (L)}_m$, $b^{\scriptscriptstyle (L)}_m$, and $\tau^2_{m \, \infty }$ of the spike and slab structures for shared and view-specific components in \textsc{jafar}.
This is also to preserve the equivalence with \textsc{jfr}, modulo imposing a structured sparsity pattern on the loading matrices.
Instead, we vary the concentration parameters $\alpha^{\scriptscriptstyle (\Lambda)}_m$ and $\alpha^{\scriptscriptstyle (\Gamma)}_m$ of the latent stick breaking processes based on the scenario considered and the dimensionality of the problem. 
In the original \textsc{cusp} construction, the latter gives the prior expectation for the number of active factors, which is useful in eliciting reasonable values for the hyperparameters.

\begin{figure}[ht!]
\centering
\includegraphics[width=0.45\linewidth]{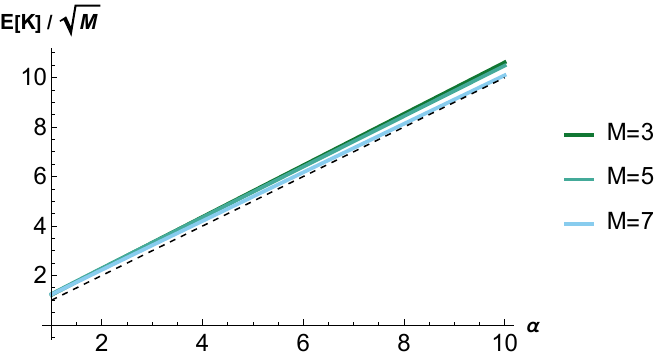}
\includegraphics[width=0.45\linewidth]{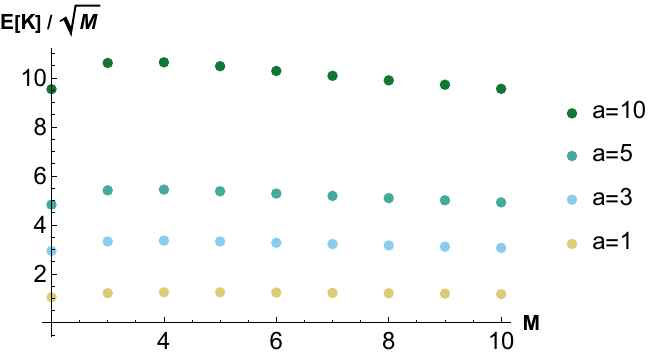}
% \vspace{-5pt}
   \caption{Expected number of factors $\mathbb{E} [K]$ in the shared component under the proposed \textsc{d-cusp} prior, varying number of views and common \textsc{dp} concentration parameter $\alpha$. Empirically, $\mathbb{E} [K]$ is roughly equal to $\alpha \cdot \sqrt{M}$, whereas in the original \textsc{cusp} it is just $\alpha$. }
\label{fig_HP_E_K}
\end{figure}

Coherently with the rationale underlying the \textsc{d-cusp} prior, the probability of any shared factor $\boeta_{\bigcdot h}$ being inactive becomes
\begin{equation*}
\begin{aligned}
    \mathbb{P} [ \, & \boeta_{\bigcdot h} \, \operatorname{inactive} \,] = \mathbb{P}[\textstyle{\sum_{m}} \mathbbm{1}_{( \delta _{mh} > h)} \leq 1] =
    \mathbb{P} [\, \textstyle{\sum_{m}} \mathbbm{1}_{( \delta _{mh} > h)} = 0] +
    \mathbb{P} [\, \textstyle{\sum_{m}} \mathbbm{1}_{( \delta _{mh} > h)} = 1]\\
    &= \textstyle{\prod_{m}} \mathbb{P}[ \delta_{mh} \leq h] + \textstyle{\sum_m} \mathbb{P}[ \delta_{mh} > h] \, 
    \textstyle{\prod_{m' \neq m}} \mathbb{P}[ \delta_{m'h} \leq h]
    \\
    &= \textstyle{\prod_{m}}\mathbb{P}[ \delta_{mh} \leq h] + \mathbb{P}[ \delta _{h}= 0] \, \textstyle{\sum_{m}}\mathbb{P}[ \delta_{mh} > h] \, \textstyle{\prod_{m' \neq m }}\mathbb{P}[ \delta_{m' h} \leq h] \; .
\end{aligned}
\end{equation*}
Consequently, the prior expectation of the shared ranks in \textsc{jafar} is
\begin{equation*}
\begin{aligned}
    \mathbb{E} [K] &= \mathbb{E} \big[ \, \textstyle{\sum_{h=1}^\infty} \big(1 -\mathbb{P} [ \, \boeta_{\bigcdot h} \, \operatorname{inactive} \, ] \, \big) \big] \\
    &= \intermediate{ \sum_{h=1}^\infty \Bigg(
    1 - \prod_{m=1}^M \bigg( 1 - \Big(\frac{\alpha^{\scriptscriptstyle (\Lambda)}_m}{1+\alpha^{\scriptscriptstyle (\Lambda)}_m}\Big)^h \bigg) -  \sum_{m=1}^M \Big(\frac{\alpha^{\scriptscriptstyle (\Lambda)}_m}{1+\alpha^{\scriptscriptstyle (\Lambda)}_m}\Big)^h \prod_{m' \neq m } \bigg( 1 - \Big(\frac{\alpha^{\scriptscriptstyle (\Lambda)}_{m'}}{1+\alpha^{\scriptscriptstyle (\Lambda)}_{m'}}\Big)^h \bigg) \Bigg) } . 
\end{aligned}
\end{equation*}
Contrary to the original \textsc{cusp} construction, $\mathbb{E} [K]$ lacks a closed form but can be trivially computed numerically. Setting $\alpha^{\scriptscriptstyle (\Lambda)}_m = \alpha$ for every $m$ for simplicity, we observe that $\mathbb{E} [K]$ scales roughly linearly in $\sqrt{M}$, as depicted in Figure~\ref{fig_HP_E_K}.
A similar argument applies to \textsc{jfr} under the \textsc{i-cusp} prior.
Accordingly, the common $\alpha$ should be set to the desired expected number of factors divided by $\sqrt{M}$. 
To make prior elicitation more user-friendly, our accompanying code in the \texttt{R} package available at \href{https://github.com/niccoloanceschi/jafar}{\texttt{https://github.com/niccoloanceschi/jafar}} performs this rescaling internally and lets the user specify directly the expected number of factors.

% ||||||||||||||||||||||||||||||||||||||||||||||||||||||||||||||
\section{Gibbs Sampler for \textsc{jafar} under \textsc{d-cusp}}\label{app_gibbs_sampler}

\setcounter{table}{0}
\setcounter{figure}{0}
\setcounter{equation}{0}
\setcounter{algocf}{0}

In this section, we provide the details on the implementation of the Gibbs sampling algorithm for \textsc{jfr} and \textsc{jafar}.
We first discuss joint sampling of shared and specific components -- both for loading matrices and latent factors -- as opposed to a sequential one.
We then report the pseudo code only for \textsc{jafar}, where \textsc{jfr} is analogous after dropping all view-specific elements.
Similarly to the main text, we will assume that the overall rank in \textsc{jfr} is equal to $K+\sum_{m=1}^M K_m$ from \textsc{jafar}.

\vspace{-5pt}

\subsection{Joint sampling of shared and view-specific loadings}
In \textsc{jafar}, for each $m=1,\dots, M$ and $j=1,\dots,p_m$ we sample jointly $[\bLambda_{m j \bigcdot},\bGamma_{m j \bigcdot}]$ from a $(K+K_m)$-dimensional normal distribution, for a total cost of $\sum_{m=1}^M  \mathcal{O}\big(p_m \cdot(K+ K_m)^3 \big)$. 
Here, the notation $`\bigcdot'$ in matrix subscripts indicates looping over possible index values along a given dimension.
Sampling from separate $K$ and $K_m$-dimensional normals is detrimental to good mixing because of the interdependence between $\bGamma_{m j \bigcdot}$ and $\bLambda_{m j \bigcdot}$.
The analogous step in \textsc{jfr} inevitably comes with a cost of $\sum_{m=1}^M  \mathcal{O}\big(p_m \cdot (K+\sum_{m'=1}^M K_{m'})^3 \big)$, due to the lack of exact structured sparsity. This rapidly becomes the bottleneck step of the sampler as the dimension of the view increases, so the \textsc{jafar} structure becomes very advantageous.
A similar rationale applies to the regression coefficients $\theta$ and $\{\theta_m\}_m$, although less impactful due to the univariate nature of the response.

\vspace{-5pt}

\subsection{Joint sampling of shared and view-specific latent factors}
Similarly, \textsc{jfr}'s latent factors should not be sampled jointly. Importantly, the precision matrix of the full conditionals is the same for all statistical units, which allows the computation to be cached.
This reduces the cost to $\mathcal{O}\big((K+\sum_{m=1}^M K_m)^3 + n \cdot (K+\sum_{m=1}^M K_m)^2\big)$ instead of $\mathcal{O}\big(n \cdot(K+\sum_{m=1}^M K_m)^3 \big)$.
For \textsc{jafar}, two different scenarios are possible depending on the presence or absence of the response.

In the unsupervised case, joint sampling of all latent factors can be achieved more efficiently via blocking and marginalization, exploiting the factorization
\begin{equation*}
\begin{aligned}
    \mathbb{P} \big[ \boeta_i, \{\bphi_{m i} \}_{m=1}^M \mid \,  \relbar \big] &= \mathbb{P} \big[\{\bphi_{m i} \}_{m=1}^M \mid  \boeta_i, \relbar \big] \mathbb{P} \big[ \boeta_i \mid \, \relbar \big] = \left( \prod_{m=1}^M \mathbb{P} \big[\bphi_{m i}  \mid  \boeta_i, \relbar \big] \right) \mathbb{P} \big[ \boeta_i \mid \, \relbar \big]. 
\end{aligned}
\end{equation*}
The hyphen $``\relbar''$ is a shorthand to specify the conditioning on all other variables, and $\mathbb{P} \big[ \boeta_i \mid \; \relbar \big]$ denotes the full conditional of the shared factors in a collapsed version of the model, where all view-specific factors have been marginalized out.
This results in a partially collapsed Gibbs sampler \citep{Park_vanDyk_2009_Partially_Collapsed_Gibbs_Samplers}.
Crucially, sampling from $\mathbb{P} \big[ \boeta_i \mid \; \relbar \big]$ and each $\mathbb{P} \big[\{\bphi_{m i} \}_{m=1}^M \mid  \boeta_i, \relbar \big]$ entails only the ${\mathcal{O}\big(\sum_{m=1}^M K_m^3 \big)}$ and ${\mathcal{O}\big( K_m^3 \big)}$ cost, respectively, to compute the cached precision matrices.

In the supervised case, the interdependence introduced by the response component makes the above partially collapsed Gibbs sampler as expensive as sampling directly the original $\mathbb{P} \big[ \boeta_i, \{\bphi_{m i} \}_{m=1}^M \mid \,  \relbar \big]$.
In fact, term $\mathbb{P} \big[\{\bphi_{m i} \}_{m=1}^M \mid  \boeta_i, \relbar \big]$ does not factorize over $m$ anymore, and the marginalized step $\mathbb{P} \big[ \boeta_i \mid \, \relbar \big]$ now costs ${\mathcal{O}\big((\sum_{m=1}^M K_m^3 )\big)}$.
Accordingly, sampling jointly all \textsc{jafar}'s latent factors comes at the same cost as in \textsc{jfr}.
If $n$ is very large, \textsc{jafar} still allows to explore possible trade-offs between reduced mixing and faster steps by sequentially targeting $\mathbb{P} \big[ \boeta_i \mid \{\bphi_{m i} \}_{m=1}^M, \relbar \big]$ and $\mathbb{P} \big[  \bphi_{m i} \mid \boeta_i, \{\bphi_{m' i} \}_{m' \neq m}, \relbar \big]$, for each $m=1,\dots,M$.

% \vspace{-5pt}

\subsection{Gibbs sampler pseudo-codes}

\setcounter{figure}{0}
\renewcommand{\thefigure}{\thesection\arabic{figure}}
\setcounter{equation}{0}
\renewcommand{\theequation}{\thesection\arabic{equation}}
\setcounter{algocf}{0}
\renewcommand{\thealgocf}{\thesection\arabic{algocf}}

Algorithm~\hyperref[alg:B1]{B1} outlines the steps of each Gibbs sampler iteration for \textsc{jafar} under the proposed \textsc{d-cusp} prior on $\{\bLambda_m\}_m$.
Similarly to \citet{Legramanti_2020_CUSP}, the loadings' likelihood contributions in Step 4. arise from the collapsed full conditionals after the marginalization of the variances $\chi_{m h}^2$, $\tau_{m h}^2$, $\psi_h^2$, and $\psi_{m h}^2$
\begin{equation*}
\begin{aligned}
    f \big(\bLambda_{m \bigcdot h} \mid \zeta_{m h} = \ell_{m h } \big) &= \begin{cases}
        t_{p_m, 2 \, a^{\scriptscriptstyle(L)}_m} \big( \bLambda_{m \bigcdot h} ; \bzero_{p_m}, (b^{\scriptscriptstyle (L)}_m / a^{\scriptscriptstyle (L)}_m) \bI_{p_m} \big) \qquad\quad & \operatorname{if} \ell_{m h} > h \\
        \phi_{p_m} \big(\bGamma_{m \bigcdot h} ; \bzero_{p_m}, \tau_{m\infty}^2 \bI_{p_m} \big)  & \operatorname{otherwise} \\
    \end{cases} \\[2pt]
    f \big(\bGamma_{m \bigcdot h} \mid \delta_{m h} = \ell_{m h } \big) &= \begin{cases}
        t_{p_m, 2 \, a^{\scriptscriptstyle(L)}_m} \big( \bGamma_{m \bigcdot h} ; \bzero_{p_m}, (b^{\scriptscriptstyle (L)}_m / a^{\scriptscriptstyle (L)}_m) \bI_{p_m} \big) \qquad\quad & \operatorname{if} \ell_{m h} > h \\
        \phi_{p_m} \big(\bGamma_{m \bigcdot h} ; \bzero_{p_m}, \tau_{m\infty}^2 \bI_{p_m} \big)  & \operatorname{otherwise} \\
    \end{cases} \\[2pt]
    f \big(\btheta_{h} \mid r_{h} = s_{h } \big) &= \begin{cases}
        t_{2 \, a^{\scriptscriptstyle (\theta)}} \big( \btheta_{h} ; 0, (b^{\scriptscriptstyle (\theta)} / a^{\scriptscriptstyle (\theta)}) \big) \qquad\quad & \operatorname{if} s_{ h} =1 \\
        \phi \big(\btheta_{h} ; 0, \psi_{\infty}^2 \big)  & \operatorname{otherwise} \\
    \end{cases} \\[2pt]
    f \big(\btheta_{m h} \mid r_{m h} = s_{m h } \big) &= \begin{cases}
        t_{2 \, a^{\scriptscriptstyle (\theta)}} \big( \btheta_{m h} ; 0, (b^{\scriptscriptstyle (\theta)} / a^{\scriptscriptstyle (\theta)}) \big) \qquad\quad & \operatorname{if} s_{m h} =1 \\
        \phi \big(\btheta_{m h} ; 0, \psi_{\infty}^2 \big)  & \operatorname{otherwise} \\
    \end{cases}
\end{aligned}
\end{equation*}
where $t_{p, \kappa} \big( \, \cdot \; ; \br, \bC \big)$ and $\phi_{p} \big( \, \cdot \; ; \br, \bC \big)$ denote the pdf of $p$-variate Student-$t$ and normal distributions, respectively, where $\kappa > 1$ is the degrees of freedom, $\br$ is a location vector and $\bC$ is a scale matrix.

% \newpage
\vspace{10pt}

\noindent\rule{\linewidth}{1pt} \\[-1pt]
\noindent \textbf{Algorithm B1:} Gibbs sampler for \textsc{jafar} under the \textsc{d-cusp} prior $\textcolor{white}{\bigg)}$
\vspace{-3pt}
\\ \noindent\rule{\linewidth}{1pt}
\\[-10pt]
\begin{enumerate}[left=5pt] \label{alg:B1}
\setlength\itemsep{5pt} 
% features loadings ----------------
\item \textbf{for}   $m=1,\dots,M$ \textbf{do} \, \textcolor{gray}{\vtop{\vskip-3pt\hsize=260pt \hrulefill} \, \textbf{Loadings Vectors}} \\
$\textcolor{white}{.} \quad$ \textbf{for}   $j=1,\dots,p_m$ \textbf{do} \\
$\textcolor{white}{.} \quad \quad$ sample $[\bmu_{m j}, \bLambda_{m j \bigcdot},\bGamma_{m j \bigcdot}]$ from $\mathcal{N}_{1+K+K_m}(\bV_{m j} \bu_{m j}, \bV_{m j})$, where \\[2pt]
$\textcolor{white}{.} \quad \quad \quad$ $\begin{cases}
    \bV_{m j} = \left(\operatorname{diag}\big([\upsilon_m^{-2}, \{ \tau_{m h}^{-2} \}_{h=1}^{K},\{ \chi_{m h}^{-2}\}_{h=1}^{K_m}] \big) + \bsigma_{m j}^{-2} [\mathbf{1}_n, \boeta, \bphi_{m}]^\top[\mathbf{1}_n, \boeta, \bphi_{m}] \right)^{-1} \\
    \hspace{3pt} \bu_{m j} = \bsigma_{m j}^{-2} [\mathbf{1}_n, \boeta, \bphi_{m}]^\top \bX_{m \bigcdot j}
    \end{cases}$ \\[10pt]
% response loadings
sample $[\mu_{y},\btheta^\top,\btheta_1^\top,\dots,\btheta_M^\top]^\top$ from $\mathcal{N}_{1+\widetilde{K}}(\bV_y \bu_y, \bV_y)$, where $\widetilde{K}=K+{\sum_m}K_m$ and \\[2pt]
$\textcolor{white}{.} \quad$
$\begin{cases}
\bV_y = \left(\operatorname{diag}( [\upsilon_y^{-2}, \{ \psi_{h}^{-2}\}_{h=1}^{K}, \{\{ \psi_{m h}^{-2}\}_{h=1}^{K}\}_{m=1}^M]) + \bsigma_{y}^{-2} [\mathbf{1}_n, \tilde{\boeta}]^\top [\mathbf{1}_n, \tilde{\boeta}] \right)^{-1}  \\
\hspace{3pt} \bu_y = \sigma_{y}^{-2} [\mathbf{1}_n,\tilde{\boeta}]^\top \by \qquad \text{with} \qquad \tilde{\boeta} = [\boeta, \bphi_1, \dots, \bphi_M]
\end{cases}$     
\item \textbf{for}   $m=1,\dots,M$ \textbf{do} \, \textcolor{gray}{\vtop{\vskip-3pt\hsize=220pt \hrulefill} \, \textbf{
Idiosyncratic Variances}}\\
$\textcolor{white}{.} \quad $ \textbf{for}   $j=1,\dots,p_m$ \textbf{do} \\
$\textcolor{white}{.} \quad \quad \quad$ sample $\sigma_{m j}^2$ from $\mathcal{I}nv\mathcal{G}a ( a^{(m)} + 0.5 \, n, b^{(m)} + 0.5 d_{m j} )$, where \\
$\textcolor{white}{.} \quad \quad \quad \quad$ $d_{m j} = \operatorname{sum}\big( (\bX_{m \bigcdot j} - \mathbf{1}_n \bmu_{m j} - \boeta \bLambda_{m j \bigcdot} - \bphi_m \bGamma_{m j \bigcdot})^2 \big) $ \\
sample $\sigma_y^2$ from $\mathcal{I}nv\mathcal{G}a \left( a^{(y)} + 0.5 \, n, b^{(y)} + 0.5 \operatorname{sum}\big( (\by - \mathbf{1}_n \mu_y - \boeta\,\btheta -  \textstyle{\sum_{m=1}^M } \bphi_m \btheta_m )^2 \big) \right)$ 
\item 
set $\bV = \bA^{-1} \in \Re^{\widetilde{K} \times \widetilde{K}}$ where $ \bA = \{\{ \bA_{[m][m']} \}_{m=0}^M \}_{m'=0}^M$ and \, \textcolor{gray}{\vtop{\vskip-3pt\hsize=80pt \hrulefill} \, \textbf{Latent Factors}}  \\[2pt]
$\textcolor{white}{.} \quad $ $\bA_{[m][m']} = \bA_{[m'][m]}^\top = \left\{\,
\begin{aligned}
&\bI_K + \sigma_y^{-2} \btheta \btheta^\top + \textstyle{\sum_{m=1}^M} \bLambda_{m}^\top \operatorname{diag}(\bsigma_m^{-2}) \bLambda_{m}
\qquad && \operatorname{if} \, m=m'=0 \\
&\bI_{K_m} + \sigma_y^{-2} \btheta_m \btheta_m^\top + \bGamma_{m}^\top \operatorname{diag}(\bsigma_m^{-2}) \bGamma_{m} \qquad && \operatorname{if} \, m=m'>0 \\
&\sigma_y^{-2} \btheta \btheta_{m'}^\top + \bLambda_{m'}^\top \operatorname{diag}(\bsigma_{m'}^{-2}) \bGamma_{m'} \qquad && \operatorname{if} \, m=0, \, m'>0 \\
&\sigma_y^{-2} \btheta_m \btheta_{m'}^\top \qquad && \operatorname{otherwise} 
\end{aligned} \right.$\\
\textbf{for}   $i=1,\dots,n$ \textbf{do} \\
$\textcolor{white}{.} \quad $ sample $\widetilde{\boeta}_i$ from $\mathcal{N}_{\widetilde{K}}(\bV \bu_i, \bV)$, where $\bu_i = \{ \bu_{i[m]} \}_{m=0}^M$ with \\[2pt]
$\textcolor{white}{.} \quad\quad $ $\bu_{i[m]} = \left\{\,
\begin{aligned}
&\sigma_y^{-2} \btheta (\by_i - \mu_y) + \textstyle{\sum_{m=1}^M} \bLambda_{m}^\top \operatorname{diag}(\bsigma_m^{-2}) (\bx_{m i} - \bmu_m)
\qquad && \operatorname{if} \, m=0 \\
&\sigma_y^{-2} \btheta_m (\by_i - \mu_y) + \bGamma_{m}^\top \operatorname{diag}(\bsigma_m^{-2}) (\bx_{m i} - \bmu_m) \qquad && \operatorname{otherwise} 
\end{aligned} \right.$\\
\item
\textbf{for}   $m=1,\dots,M$ \textbf{do} \, \textcolor{gray}{\vtop{\vskip-3pt\hsize=190pt \hrulefill} \, \textbf{Spike and Slab Memberships}} \\
$\textcolor{white}{.} \quad $ \textbf{for}   $h=1,\dots,K$ \textbf{do} \\
$\textcolor{white}{.} \quad \quad$ sample $\zeta_{m h} \in \{1,\dots,K\}$ from $\mathbb{P} \big[ \zeta_{m h} = \ell_{m h} \mid \,\relbar \big] \propto 
\omega_{m \ell_{m h}} \cdot f \big( \bLambda_{m \bigcdot h} \mid \zeta_{m h} = \ell_{m h} \big)$ \\
$\textcolor{white}{.} \quad \quad \quad$ where $\omega_{m h} = \nu_{m h} \, \textstyle{\prod_{l=1}^{h-1}} (1-\nu_{m l})$\\
$\textcolor{white}{.} \quad $ \textbf{for}   $h=1,\dots,K_m$ \textbf{do} \\
$\textcolor{white}{.} \quad \quad$ sample $\delta_{m h} \in \{1,\dots,K_m\}$ from $\mathbb{P} \big[ \delta_{m h} = \ell_{m h} \mid \,\relbar \big] \propto
\varpi_{m \ell_{m h}} \cdot f \big( \bGamma_{m \bigcdot h} \mid \delta_{m h} = \ell_{m h} \big)$ \\
$\textcolor{white}{.} \quad \quad \quad$ where $\varpi_{m h} = \rho_{m h} \, \textstyle{\prod_{l=1}^{h-1}} (1-\rho_{m l})$\\
\textbf{for}   $h=1,\dots,K$ \textbf{do} \\
$\textcolor{white}{.} \quad $ sample $r_h \in \{0,1\}$ from $\mathbb{P} \big[r_h = s_h \mid \,\relbar \big] \propto (1-\xi)^{s_h} \cdot \xi^{s_h} \cdot f \big(\btheta_h \mid r_h = s_h\big)$ \\
\textbf{for}   $m=1,\dots,M$ \textbf{do} \\
$\textcolor{white}{.} \quad $ \textbf{for}   $h=1,\dots,K_m$ \textbf{do} \\
$\textcolor{white}{.} \quad \quad$ sample $r_{m h} \in \{0,1\}$ from $\mathbb{P} \big[r_{m h} = s_{m h} \mid \,\relbar \big] \propto (1-\xi)^{s_{m h}} \cdot \xi^{s_{m h}} \cdot f \big(\btheta_{m h} \mid r_{m h} = s_{m h}\big)$ 
\item 
\textbf{for}   $m=1,\dots,M$ \textbf{do} \, \textcolor{gray}{\vtop{\vskip-3pt\hsize=210pt \hrulefill} \, \textbf{
Stick Breaking Elements}} \\
$\textcolor{white}{.} \quad $ \textbf{for}   $h=1,\dots,K-1$ \textbf{do} \\
$\textcolor{white}{.} \quad \quad$ sample $\nu_{m h}$ from $\mathcal{B}e \big(1 + \sum_{l=1}^K \bbone_{(\zeta_{ml} = h)}, \alpha^{\scriptscriptstyle (\Lambda)}_m + \sum_{l=1}^K \bbone_{(\zeta_{ml} > h)} \big)$ \\
$\textcolor{white}{.} \quad $ \textbf{for}   $h=1,\dots,K_m-1$ \textbf{do} \\
$\textcolor{white}{.} \quad \quad$ sample $\rho_{m h}$ from $\mathcal{B}e \big(1 + \sum_{l=1}^{K_m} \bbone_{(\delta_{ml} = h)}, \alpha^{\scriptscriptstyle (\Gamma)}_m + \sum_{l=1}^{K_m} \bbone_{(\delta_{ml} > h)} \big)$ \\
sample $\xi$ from $\mathcal{B}e \big(a^{(\xi)}+\sum_{h=1}^K (1-r_h) + \sum_{m=1}^M \sum_{h=1}^{K_m} (1-r_{m h}), b^{(\xi)} + \sum_{h=1}^K r_h + \sum_{m=1}^M \sum_{h=1}^{K_m} r_{m h} \big)$
\item 
\textbf{for}   $m=1,\dots,M$ \textbf{do} \, \textcolor{gray}{\vtop{\vskip-3pt\hsize=210pt \hrulefill} \, \textbf{
Loadings Hyper-Variances}} \\
$\textcolor{white}{.} \quad $ \textbf{for}   $h=1,\dots,K$ \textbf{do} \\
$\textcolor{white}{.} \quad \quad$ set $\tau_{m h}^2 = \tau_{m \infty}^2 $; \, if $\zeta_{m h} > h$ sample $\tau_{m h}^2$ from $\mathcal{I}nv\mathcal{G}a \big(a^{\scriptscriptstyle(L)}_m + 0.5 \, p_m  , b^{\scriptscriptstyle (L)}_m + 0.5 \sum_{j=1}^{p_m} \bLambda_{m j h}^2 \big)$\\
$\textcolor{white}{.} \quad $ \textbf{for}   $h=1,\dots,K_m$ \textbf{do} \\
$\textcolor{white}{.} \quad \quad$ set $\chi_{m h}^2 = \tau_{m \infty}^2 $; \, if $\delta_{m h} > h$ sample $\chi_{m h}^2$ from $\mathcal{I}nv\mathcal{G}a \big(a^{\scriptscriptstyle(L)}_m + 0.5 \, p_m  , b^{\scriptscriptstyle (L)}_m + 0.5 \sum_{j=1}^{p_m} \bGamma_{m j h}^2 \big)$\\
sample $\psi_o^2$ from $\mathcal{I}nv\mathcal{G}a \big(a^{\scriptscriptstyle(\theta)} + 0.5 \, \tilde{\br}^\top \mathbf{1}_{\widetilde{K}}, b^{\scriptscriptstyle (\theta)} + 0.5 \, \tilde{\br}^\top \widetilde{\btheta} \, \big)$ \quad where \quad $\widetilde{K}=K+\sum_m K_m$, \\
$\textcolor{white}{.} \quad$ $\tilde{\br}= [r_{\bigcdot},r_{1 \bigcdot},\dots,r_{M \bigcdot}]$, and $\widetilde{\btheta} = [\btheta^\top, \btheta_1^\top, \dots, \btheta_M^\top]^\top$ \\
\textbf{for}   $h=1,\dots,K$ \textbf{do} \\
$\textcolor{white}{.} \quad $ set $\psi_h^2 = \psi_\infty^2$; \, if $r_h = 1 $ set $\psi_h^2 = \psi_o^2$ \\
$\textcolor{white}{.} \quad $ \textbf{for}   $m=1,\dots,M$ \textbf{do}\\
$\textcolor{white}{.} \quad \quad$ set $\psi_{mh}^2 = \psi_\infty^2$; \, if $r_{mh} = 1 $ set $\psi_{mh}^2 = \psi_o^2$
\item perform the adaptation step as in Algorithm~\ref{algo_adaptation_step}  \, \textcolor{gray}{\vtop{\vskip-3pt\hsize=140pt \hrulefill} \, \textbf{
Adaptation}} \\[5pt]
\noindent\rule{\linewidth}{1pt} \\
\end{enumerate}

Starting from conservative upper bounds, the number of shared factors $K$ and view-specific ones $\{ K_m \}_m$ are learned as part of the inferential procedure
through the adaptation step, reported in Algorithm~\ref{algo_adaptation_step}.
This is particularly crucial for \textsc{d-cusp}, where the adaptation step is simultaneously used to separate specific factors from fully or partially shared ones.
We set $t_{adapt}=200$, $d_0=-0.5$, and $d_1=-5\cdot10^{-4}$.

\setcounter{algocf}{1} 
\begin{algorithm}[htb]
\caption{Adaptation step in \textsc{jafar} under the \textsc{d-cusp} prior $\textcolor{white}{\Big(}$}
~\\[-1pt]
\For{$t=1,\dots,T_{\textsc{mcmc}}$}{~\\[0pt]
sample all steps in Algorithm~\hyperref[alg:B1]{B1}\\[5pt]
sample $u_t \sim \mathcal{U}(0,1)$ \\[5pt]
\If{$t \geq t_{adapt} \operatorname{and} u_t < \exp(d_0 + d_1 t)$}{
~\\[0pt] $K^* = K - \textstyle{\sum_{h=1}^K} \mathbbm{1}\big({ 1 \geq \sum_{m=1}^M \mathbbm{1}({\zeta_{mh} > h)} }\big)$ \\[5pt]
\If{$K^* < K -1$}{~\\[0pt]
set $K = K^* +1 $\\
drop the inactive columns in $\{ \bLambda_m \}_m$,  $\btheta$, and $\boeta$ \\
add an inactive shared factor, sampling its loadings from the spike
}
\Else {~\\[-2pt]
set $K = K +1 $\\
add an inactive shared factor, sampling its loadings from the spike
}
\For{$m=1,\dots,M$}{~\\[0pt]
$K_m^* = K_m - \textstyle{\sum_{h=1}^{K_m}} \mathbbm{1}{( \delta_{m h} \leq h)}$ \\[5pt]
\If{$K_m^* < K_m -1$}{~\\[0pt]
set $K_m = K_m^* +1 $\\
drop the inactive columns in $\bGamma_m$, $\btheta_m$ and $\bphi_m$ \\
add an inactive factor for the $m^{th}$ view, sampling its loadings from the spike
}
\Else {~\\[-2pt]
set $K_m = K_m +1 $ \\
{add an inactive factor for the $m^{th}$ view, sampling its loadings from the spike}
}
}
}
}
\label{algo_adaptation_step}
\end{algorithm}

In the unsupervised version of \textsc{jafar}, all parts related to the response are clearly dropped.
Most importantly, the update of the latent factors can be optimized by targeting a partially collapsed Gibbs sampler, as explained above. The resulting modified update is reported in Algorithm~\hyperref[alg:B3]{B3}.

\newpage

\noindent\rule{\linewidth}{1pt} \\[-1pt]
\noindent \textbf{Algorithm B3:} Sampling of latent factors for unsupervised \textsc{jafar} $\textcolor{white}{\bigg)}$
\vspace{-3pt}
\\ \noindent\rule{\linewidth}{1pt}
\\[-10pt]
\begin{enumerate}[left=5pt] \label{alg:B3}
\setcounter{enumi}{2} 
\setlength\itemsep{5pt} 
% latent factors ----------------
\item
set $\bV = \left( \bI_K + \sum_{m=1}^M \bLambda_{m}^\top \big( \bGamma_{m} \bGamma_{m}^\top + \operatorname{diag}(\bsigma_m^2) \big)^{-1} \bLambda_{m} \right)^{-1}$ \, \textcolor{gray}{\vtop{\vskip-3pt\hsize=90pt \hrulefill} \, \textbf{Latent Factors}} \\
\textbf{for} $i=1,\dots,n$ \textbf{do} \\
$\textcolor{white}{.} \quad $ sample $\boeta_i$ from $\mathcal{N}_K(\bV \, \bu_i, \bV)$, where
$\hspace{0pt} \bu_i = \sum_{m=1}^M \bLambda_{m}^\top \big( \bGamma_{m} \bGamma_{m}^\top + \operatorname{diag}(\bsigma_m^2) \big)^{-1} (\bx_{m i} - \bmu_m)$ \\
\textbf{for} $m=1,\dots,M$ \textbf{do} \\
$\textcolor{white}{.} \quad $ set $\bV_m = \left( \bI_{K_m} + \bGamma_{m}^\top \operatorname{diag}(\bsigma_m^{-2}) \bGamma_{m} \right)^{-1}$ \\
$\textcolor{white}{.} \quad $ \textbf{for} $i=1,\dots,n$ \textbf{do} \\
$\textcolor{white}{.} \quad \quad $ sample $\bphi_{m i}$ from $\mathcal{N}_{K_m}(\bV_m \bu_{m i}, \bV_m)$, where 
$\bu_{m i} = \bGamma_{m}^\top \operatorname{diag}(\bsigma_m^{-2})(\bx_{m i} - \bmu_m - \bLambda_{m} \boeta_i)$\\[5pt]
\noindent\rule{\linewidth}{1pt}
\end{enumerate}

The Gibbs sampler for \textsc{jfr} under the \textsc{i-cusp} prior is analogous to that of Algorithm~\hyperref[alg:B1]{B1}, modulo a few modifications to rewrite it in terms of a single set of factors, dropping all view-specific components.
Similarly, the adaptation of the overall rank is analogous to the first part of Algorithm~\ref{algo_adaptation_step}, except that $K^*$ is now computed as $K - \textstyle{\sum_{h=1}^K} \mathbbm{1}\big({ \max_m{\zeta_{mh}} \leq h }\big)$.

\subsection{A note on out-of-sample prediction}

In all factor model formulations, out-of-sample predictions such as $\E \big[\, y_i \mid \{\bX_{m i}\}_{m=1}^M, \relbar \big]$ can be easily constructed via Monte Carlo averages $\frac{1}{T_{\textsc{eff}}} \sum_{t=1}^{T_{\textsc{eff}}} \E \big[\, y_i \mid \tilde{\boeta}_i^{(t)} \relbar \big]$ exploiting samples from ${\tilde{\boeta}_i^{(t)} \sim p \big( \tilde{\boeta}_i \mid \{\bX_{m i}\}_{m=1}^M, \relbar \big)}$, where $T_{\textsc{eff}}$ is the number of \textsc{mcmc} samples after burn-in and thinning and $\tilde{\boeta}_i$ are the concatenated factors.
To ensure coherence in this analysis, it was necessary to modify the function \texttt{bsfp.predict} from the main \textsc{bsfp} GitHub repository.
Indeed, the default implementation considers only samples from ${p \big(\boeta_i \mid y_i, \{\bX_{m i}\}_{m=1}^M, \relbar \big)}$ and ${p \big(\bphi_{m i} \mid y_i, \bX_{m i}, \relbar \big)}$, whereas conditioning on the response is ill-posed in our settings.
The updated code is available in the \textsc{jafar} GitHub repository.

\section{Postprocessing and Multiview \texttt{MatchAlign}}

In the current section, we provide the details for the proposed adaptation of \texttt{MatchAlign} \citep{poworoznek2025_MatchAlign} to multiview settings,
solving rotational ambiguity and the associated challenges in inferring latent variables and factor loadings.
Specifically, \texttt{MatchAlign} first applies \texttt{Varimax} rotations \citep{Kaiser1958_Varimax} to each loadings sample and then resolves column label and sign ambiguities by matching each sample to a reference via a greedy optimization procedure.
In multiview settings, \texttt{Varimax} could be naively applied to \textsc{mcmc} samples of the stacked loadings matrices
$\bLambda = [\bLambda_1^\top, \dots, \bLambda_M^\top, \btheta^\top]^\top$ -- and separately on each element view-specific samples $\bGamma_m = [\bGamma_m, \btheta_m^\top]^\top$ in \textsc{jafar}.
Instead, we propose a more nuanced modification to better respect the multiview structure.

% \subsection{Multiview \texttt{Varimax}}

A side-benefit of \texttt{Varimax} is inducing row-wise sparsity in the loadings matrices, which allows for clearer interpretability of the role of different latent sources of variability.
This is because, given any $p \times K$ loading matrix $\bLambda$, the \texttt{Varimax} procedure solves the optimization problem $\bR_o = \argmax_{\bR \in \Re^{K \times K} : \, \bR \bR^\top = \bI_K} V(\bLambda, \bR) $, where
\begin{equation*}
    V(\bLambda, \bR) = \frac{1}{p} \sum_{h=1}^K \sum_{j=1}^p \big( \bLambda \bR \big)_{jh}^4 - \sum_{h=1}^K \bigg( \, \frac{1}{p} \sum_{j=1}^p \big( \bLambda \bR \big)_{jh}^2 \bigg)^2 \; .
\end{equation*}
Accordingly, $\bR_o$ is the optimal rotation matrix maximizing the sum of the variances of the squared loadings.
Intuitively, this is achieved under two conditions.
First, any given $\bx_{j}$ has large loading $\bLambda_{jh^*}$ on a single factor $h^*$, but near-zero loadings $\bLambda_{j-h^*}$ on the remaining $K-1$ factors.
Secondly, any $h^{th}$ factor is loaded on by only a small subset $\mathcal{J}_h \subset \{1,\dots,p\}$ of variables, having high loadings $\bLambda_{\mathcal{J}_h h}$ on such a factor, while the loadings $\bLambda_{-\mathcal{J}_h h}$ associated with the remaining $ \{1,\dots,p\} \setminus \mathcal{J}_h$ variables are close to zero. 

\textcolor{black}{Naively targeting the stacked shared loadings incorporates cross-view variance terms into the \texttt{Varimax} objective, which can be suboptimal.
The resulting rotation may be overly sensitive to the relative number of features per view, giving disproportionate weight to views with higher dimensionality. Likewise, strong signals from one view may mask weaker but meaningful patterns in others, encouraging rotations that enforce a false common sparsity profile across views.}
\textcolor{black}{Therefore, we suggest targeting a modified objective}
\begin{equation}\label{eq_varimax_mod}
    \sum_{m=1}^M V(\bLambda_m, \bR) = \sum_{m=1}^M \left( \frac{1}{p_m} \sum_{h=1}^K \sum_{j=1}^{p_m} \big( \bLambda_m \bR \big)_{jh}^4 - \sum_{h=1}^K \bigg( \, \frac{1}{p_m} \sum_{j=1}^{p_m} \big( \bLambda_m \bR \big)_{jh}^2 \bigg)^2 \right) 
\end{equation}
\textcolor{black}{within the criterion 
$\bR_\star = \argmax_{\bR \in \Re^{K \times K} : \, \bR \bR^\top = \bI_K} \sum_{m=1}^M V(\bLambda_m, \bR)$, thereby
restricting the attention to the within-view squared loadings sum of variances.}
\textcolor{black}{Importantly, excluding cross-view variance terms in equation~\eqref{eq_varimax_mod} does not alter the actual intra-view covariance structure, which is preserved under any rotation.}

\textcolor{black}{Optimization of the modified target entails trivial modification of the original routine and achieves the same cost.
Hence, post-processing of each \textsc{mcmc} sample entails a
$\mathcal{O}\big( (K+\sum_{m=1}^M K_{m})^3 +  (\sum_{m=1}^M p_m) \cdot (K+\sum_{m=1}^M K_{m})^2 \big)$ cost for \textsc{jfr}, compared to the
$\mathcal{O}\big( K^3+ \sum_{m=1}^M K_{m}^3 + (\sum_{m=1}^M p_m) \cdot K^2 +\sum_{m=1}^M p_m \cdot K_{m}^2 \big)$ cost in \textsc{jafar}, again showing the benefit of assuming structure sparsity.}

%%%$$$$$$$$$$$$$$$$$$$$$$$$$$$$$$$$$$$$$$$$$$$$$$$$$
\section{Extreme large-p-small-n settings and tempered \textsc{cusp}}\label{app_tempering}

Our work is motivated by applications such as multiomics data, where multiple views of thousands of features are measured on a small set of subjects, often below $n=100$ or even $n=50$.
To the best of our knowledge, the use of \textsc{cusp} was still unexplored in such high-dimensional and unbalanced settings. This is partly because the original implementation unnecessarily scales cubically in the number of features, which we instead bring down to linear scaling.
The dataset analyzed in Section~\ref{sec_application} falls exactly in this extreme large-p-small-n.
On such data, the methodologies we propose achieve good performance in terms of accuracy, but infer a high number of factors.
In particular, the total inferred ranks under both \textsc{jfr} and \textsc{jafar} are greater than the sample size.

Formally, this is not ill-posed within Bayesian factor models. This is because identifiability of the overall induced covariance and of its diagonal and low-rank components constrains $K$ only against $p$, rather than $n$ \citep{fruhwirth2018sparse,papastamoulis2020identifiability}.
Inferred ranks greater than the sample size potentially signal the need to gather measurements for more subjects, to better capture the nuance of the dependence structure in high dimensions.
Yet, having $K$ greater than $n$ is undesirable -- for example, for interpretability, since the latent factors cannot but be linearly dependent.

In our analysis, we found that the large number of factors comes from the onset of a curse of dimensionality within the \textsc{cusp} construction in extreme large-p-small-n settings.
Focusing on $\bLambda_m$'s, this is evident in the conditional updates of latent categorical variables 
$z_{mh}$
\begin{equation}\label{eq_cusp_update}
    \log\mathbb{P} \big[ z_{m h} = \ell \mid \; \relbar \big] \propto \log\omega_{m \ell \cdot} + \mathbbm{1}_{(\ell>h)} \cdot \Delta \log f (\bLambda_{m \bigcdot h})
\end{equation}
within Step 4. of the Gibbs sampler detailed in Algorithm~\hyperref[alg:B1]{B1}.
Here,  $\Delta \log f (\bLambda_{m \bigcdot h})$ gives the difference in the log-PDF for the $h$-th column of $\bLambda_m$ under the slab and spike component, that is
\begin{equation*}
    \Delta \log f (\bLambda_{m \bigcdot h}) = \log t_{p_m, 2 \, a^{\scriptscriptstyle(L)}_m} \big( \bLambda_{m \bigcdot h} ; \bzero_{p_m}, (b^{\scriptscriptstyle (L)}_m / a^{\scriptscriptstyle (L)}_m) \bI_{p_m} \big) - \log \phi_{p_m} \big(\bLambda_{m \bigcdot h} ; \bzero_{p_m}, \tau_{m\infty}^2 \bI_{p_m} \big) \;.
\end{equation*}
The effective sample size for such an update is given by the number of features $p_m$ rather than the sample size $n$.
Indeed, $\Delta \log f (\bLambda_{m \bigcdot h})$ grows linearly in $p_m$, so that its scale gets uncomparable with that of the prior contribution $\log\omega_{m \ell \cdot}$ for large $p_m$.
This overpowers the increasing shrinkage induced by the \textsc{cusp} construction, which is thus not sufficient to limit the number of factors below $n$.
Conversely, the model adds several factors in an effort to explain the rich dependence structure in the induced covariance.

\subsection{Tempered \textsc{cusp} updates}

We explored adjusting the \textsc{cusp} hyperparameters to combat or moderate this effect, in an effort to induce more aggressive shrinkage. 
Among others, we drew inspiration from theoretical arguments in the sparse regression literature \citep{Castillo2015}, which suggest scaling the slab weights as a fractional power of the problem dimensionality within spike-and-slab formulations.
These strategies proved insufficient for our purposes.
Instead, we offer a solution that prioritizes robustness and practical performance, reflecting a broader trend in the literature of relaxing the canonical Bayesian rules.

For instance, generalized Bayes approaches replace the likelihood with an exponentiated loss function \citep{Bissiri2016generalized_Bayes}, providing flexibility to robustly handle model misspecification and to focus the posterior on aspects of the data that are most relevant to the inferential goal.
Other approaches, such as power posteriors \citep{Holmes2017power_likelihood}, fractional posteriors \citep{Bhattacharya2019fractional}, or the coarsened posterior \citep{Miller2019Coarsening}, hinge on raising the likelihood to a fractional power, either to temper the influence of the data, to control overfitting in high-dimensional settings, or to enhance robustness -- e.g. to model misspecification.

Analogously, the effective solution we propose hinges on tempering the likelihood term \newline
$\Delta \log (f (\bLambda_{m \bigcdot h}))^{\text{T}_m}$ that leads to update~\eqref{eq_cusp_update}, or equivalently rescaling it as
\begin{equation}\label{eq_cusp_tempered_update}
    \log\mathbb{P} \big[ z_{m h} = \ell \mid \; \relbar \big] \propto \log\omega_{m \ell \cdot} + {\text{T}_m} \cdot \mathbbm{1}_{(\ell>h)} \cdot \Delta \log f (\bLambda_{m \bigcdot h}) \;,
\end{equation}
which results in a Pseudo-Gibbs sampler \citep{rendsburg22a}.
The tempering coefficient $\text{T}_m$ is a fractional power that aims at reducing the effective sample size in this update, allowing \textsc{cusp} to control rank estimation more effectively in unbalanced settings.
We propose setting $\text{T}_m=\frac{\min (n,p_m)}{p_m}$, which directly sets the effective sample size for the conditional update to the smallest dimension in the data.
This also avoids cumbersome tuning of $\text{T}_m$.

The choice of tempering function may have important implications for both theoretical properties and practical performance.
We emphasize that this topic warrants a more rigorous and systematic investigation, but defer it to future work. 
In fact, the tempered version of \textsc{cusp} -- and of its extensions \textsc{i-cusp} and \textsc{d-cusp} -- are proposed here merely as an adaptation to reduce the number of factors in extreme large-p-small-n, rather than as a complete replacement of the original.

%%%$$$$$$$$$$$$$$$$$$$$$$$$$$$$$$$$$$$$$$$$$$$$$$$$$$
\section{Modeling extensions for flexible data representations }\label{app_model_extensions_nonlinear}

\setcounter{table}{0}
\setcounter{figure}{0}
\setcounter{equation}{0}
\setcounter{algocf}{0}

Equations~\eqref{eq_jfr_linear} and \eqref{eq_jafar_linear} can be viewed as main building blocks of more complex modeling formulations, allowing greater flexibility in the descriptions of both the multiview data and the response component.
We first address deviations from normality in the multiview data, which is a fragile assumption in Gaussian factor models. 
In many applications, such as multi-omics data, the features are often non-normally distributed, right-skewed, or exhibit multi-modal marginals.
Nonetheless, Gaussian formulations as in equations~\eqref{eq_jfr_linear} and \eqref{eq_jafar_linear} demand that the latent factor decomposition simultaneously describe the dependence structure and the marginal distributions of the features.
This can negatively affect the performance of the methodology, while having a confounding effect on the identification of latent sources of variation.
To address this issue, we develop a copula factor model extension of \textsc{jafar} \citep{hoff2007extending,Murray_2013_Copula_FA,feldman2023nonparametric}, which allows us to disentangle learning of the dependence structure from that of margins.
Notably, the \textsc{d-cusp} prior structure described above readily applies to such extensions as well.

Furthermore, \textsc{jafar} can be easily generalized to account for deviations from normality and linearity in the response, other than binary and count $\by$.
Here we present different ways to achieve this.
For the sake of completeness, we note that higher flexibility could also be achieved by considering alternative approaches, beyond those reported below.
For instance, recent contributions in factor models have shown the benefit of assuming a mixture of normals as prior distribution for the latent factors \citep{chandra2023_lamb}.

\subsection{Non-Gaussian data: single-view case \& copula factor regression}
For ease of exposition, we first introduce Copula Factor Models in the simplified case of a single set of features $\bx_i \in \Re^p$, 
before extending to the multiview case.
Adhering to the formulation in \citep{hoff2007extending}, we model the joint distribution of $\bx_i$ as 
$\mathrm{F}(\bx_{i1},\ldots, \bx_{ip}) = \mathcal{C} (\mathrm{F}_1(\bx_{i1}),\ldots, \mathrm{F}_p(\bx_{ip}))$, 
where $\mathrm{F}_j$ is the univariate marginal distribution of the $j^{th}$ entry, and $\mathcal{C}(\cdot)$ is a distribution function on $[0,1]^p$ that describes the dependence between the variables.
Any joint distribution $\mathrm{F}$ can be completely specified by its marginal distributions and a copula $\mathcal{C}$ \citep{sklar1959fonctions}, with the copula being uniquely determined when the variables are continuous.
Here we employ the Gaussian copula 
$\mathcal{C}(u_1, \ldots, u_p) = \Phi_p(\Phi^{-1}(u_1),\ldots, \Phi^{-1}(u_p) \,| \, \bSigma) $,
where $\Phi_p(\cdot \, | \, \bSigma)$ is the $p$-dimensional Gaussian cdf with correlation matrix $\bSigma$, $\Phi(\cdot)$ is the univariate standard Gaussian cdf and $[u_1, \ldots, u_p] \in [0,1]^p$.
Plugging in the Gaussian copula in the general formulation, the implied joint distribution of $\bx_i$ is
\begin{gather*}
    \mathrm{F}(\bx_{i1},\ldots, \bx_{ip}) = \Phi_p \Big( \Phi^{-1}\big(\mathrm{F}_1(\bx_{i1}) \big), \ldots, \Phi^{-1}\big(\mathrm{F}_p(\bx_{ip}) \big) \mid \bSigma \Big) \; .
\end{gather*}
Hence, the Gaussian distribution is used to model the dependence structure, whereas the data have univariate marginal distributions $\mathrm{F}_j(\cdot)$.
The Gaussian copula model is conveniently rewritten via a latent variable representation, such that $\bx_{ij} = \mathrm{F}_j^{-1} \big( \Phi \left(\bz_{ij} / c_j \big)  \right)$, with $\bz_i \sim \mathcal{N}_p(\bzero_p,\bSigma)$.
Here $\mathrm{F}_j^{-1}(u) =\inf\{ x : \mathrm{F}_j(x) \geq u \}$, $\forall \, u \in (0,1) $, is the pseudo-inverse of the univariate marginal of the $j^{th}$ entry, $\bz_{ij}$ is the latent variable related to predictor $j$ and observation $i$, and $c_j$ is a positive normalizing constant.
Following \cite{Murray_2013_Copula_FA}, the learning of the potentially large correlation structure $\bSigma$ can proceed by endowing $\bz_i$ with a latent factor model
$\bz_i \sim \mathcal{N}_p ( \bLambda \boeta_i, \bD )$,
with $\bD=\operatorname{diag}(\{\sigma_j^2\}_{j=1}^p)$, $p \times k$ factor loadings matrix $\bLambda$ and latent factors $\boeta_i \sim \mathcal{N}_K(\bzero_K,\bI_K)$. 
Likewise, predictions of a continuous health outcome $y_i$ can be accounted for via a regression on the latent factors 
$y_i \sim \mathcal{N} \big( f(\boeta_i),\sigma_y^2 \big)$, where in \textsc{jafar} we consider a simple linear mapping $f(\boeta_i) = \btheta^\top \boeta_i$.
In the latter case, the induced regression is linear also in $\bz_i$:
\begin{align*}
     \E[y_i\mid \bx_i] & = \E[\btheta^\top \boeta_i \mid \bx_i]= \btheta^\top \E[\boeta_i \mid \bx_i] = \btheta^\top \E \big[ \E[ \boeta_i \mid \bz_i] \mid \bx_i \big]
    \\
    & = \btheta^\top \E \big[(\bLambda^\top \bD^{-1} \bLambda + \bI_K )^{-1} \bLambda^\top \bD^{-1} \bz_i \mid \bx_i \big] = \btheta^\top \bA \, \E[ \bz_i \mid \bx_i],
\end{align*}
where $\E[ \bz_i \mid \bx_i]$ is a vector such that the $j^{th}$ element is equal to $c_j \Phi^{-1} \big( \mathrm{F}_j (\bx_{ij}) \big)$.
This follows from the fact that the distribution of $\boeta_i \mid \bz_i$ is normal with covariance 
$\bV= ( \bLambda^\top \bD^{-1} \bLambda + \bI_K )^{-1}$ and mean $\bA \, \bz_i$ where $\bA = \bV \bLambda^\top \bD^{-1}$. 
To enforce standardization of the latent variables, $c_j = \sqrt{\sigma_j^2 + \sum_{h = 1}^k \bLambda_{jh}^2}$, which would non-trivially complicate the sampling process.
However, since the model is invariant to monotone transformations \citep{Murray_2013_Copula_FA},  we can use instead
\begin{gather*}
     \bx_{ij} = \mathrm{F}_j^{-1} \big( \Phi (\bz_{ij} )  \big) \qquad
     \bz_i \sim \mathcal{N}_p(\bLambda \boeta_i, \bD) \qquad
     \eta_i \sim \mathcal{N}(\bzero_K,\bI_K). \; 
\end{gather*}

The only element left to be addressed is the estimation of the marginal distributions $\mathrm{F}_j$.
In many practical scenarios, the features are continuous or treated as such with negligible impact on the overall analysis.
In such a setting, it is common to replace $\mathrm{F}_j(\cdot)$ by the scaled empirical marginal cdf $\hat{\mathrm{F}}_j(t) = \frac{n}{n+1} \sum_{i=1}^n \frac{1}{n} \mathbbm{1} (\bx_{ij} \le t)$, benefiting from the associated theoretical properties \citep{klaassen1997efficient}.
Alternatively, \citet{hoff2007extending} and \citet{Murray_2013_Copula_FA} viewed the marginals as nuisance parameters and targeted learning of the copula correlation for mixed data types via extended rank likelihood.
Recently, \cite{feldman2023nonparametric} proposed an extension for fully Bayesian marginal distribution estimation, with remarkable computational efficiency for discrete data. 

\subsection{Non-Gaussian data: multiview case} 
Extending the same rationale to the multiview case, the copula factor model now targets the joint distribution of $\bx = [ \bx_{1 i}^\top, \dots, \bx_{M i}^\top]^\top$ as
\begin{equation*}
    \mathrm{F}(\bx_{1  i},\dots,\bx_{M  i}) = \mathcal{C} \big( \mathrm{F}_{1  1}(\mathrm{x}_{1  i  1}),\dots, \mathrm{F}_{1  p_1}(\mathrm{x}_{1  i  p_1}), \dots, \mathrm{F}_{M  1}(\mathrm{x}_{M  i  1}),\dots, \mathrm{F}_{M  p_M}(\mathrm{x}_{M  i  p_M} ) \big). \; 
\end{equation*}
Here $p=\sum_{m=1}^M p_m$, while $\mathrm{F}_{mj}$ is the univariate marginal cdf of the $j^{th}$ variable in the $m^{th}$ view.
We focus here on \textsc{jafar}, but analogous extensions can be derived for \textsc{jfr}.
The additive latent factor structure from equation~\eqref{eq_jafar_linear} can be directly imposed on the transformed variables $\bz_i = [ \bz_{1 i}^\top, \dots, \bz_{m i}^\top]^\top$, introducing again the distinction between shared and view-specific factors.
The overall model formulation becomes
\begin{equation*}%\label{eq_jafar_copula}
\begin{aligned}
    \mathrm{\bx}_{m i j} &= \mathrm{F}_{m j}^{-1}\big(\Phi(\bz_{m i j} )\big) \\
    \bz_{m i} &= \bmu_m + \bLambda_{m} \boeta_i + \bGamma_{m} \bphi_{m i} + 
    \bepsilon_{m i}  \\
    y_i &= \mu_y + \bbeta^\top \br_i + \btheta^\top \boeta_i + \textstyle{\sum_{m=1}^M}\btheta_m^\top \bphi_{mi} + e_i \; .
\end{aligned}
\end{equation*}
As before, $\mathrm{F}_{mj}^{-1}$ is the pseudo-inverse of $\mathrm{F}_{mj}$.
Missing data can be imputed by sampling the corresponding entries $\widetilde{\bz}_{mij} \sim \mathcal{N}(\bmu_m + \bLambda_{m j \bigcdot}^\top \boeta_i + \bGamma_{m j \bigcdot}^\top \bphi_{m i}, \bsigma_{m j}^2)$ at each iteration of the sampler. 
When there is no direct interest in reconstructing the missing data, subject-wise marginalization of the missing entries can improve mixing compared to their imputation. 

% \vspace{-15pt}

\subsection{Non-linear response modeling: interactions \& splines}

We now focus on introducing more flexible dependence of $y_i$ on the latent factors.
Non-linearity in this relationship typically breaks down conditionally conjugate updates for the latent factors, requiring a Metropolis-within-Gibbs step.
Accordingly, the Gibbs sampler from the previous section remains unchanged, except for Step 3.

\subsubsection{Interactions among latent factors}\label{par_FIN}
Aside from multiview integration frameworks,
\citet{Ferrari_2021_Interactions} recently generalized Bayesian latent factor regression to accommodate interactions among the latent variables in the response component 
\begin{equation*}
    y_i = \mu_y + \bbeta^\top \br_i + \btheta^\top \boeta_i + \boeta_i^\top \bOmega \, \boeta_i + e_i \; ,
\end{equation*}
where $\bOmega$ is a $K \times K $ symmetric matrix.
Other than providing theory on model misspecification and consistency, the authors showed that the above formulation induces a quadratic regression of $y_i$ on the transformed concatenated features $\bz_i$
\begin{equation}
    \mathbb{E} [\, y_i \mid \bz_i \,]= \mu_y + \bbeta^\top \br_i + (\btheta^\top \bA ) \, \bz_i + \bz_i^\top (\bA^\top \bOmega \, \bA ) \, \bz_i + \operatorname{tr}(\bOmega \bV) \; ,
\label{eq_quadratic_regression} 
\end{equation}
where as before $\bV= ( \bLambda^\top \bD^{-1} \bLambda + \bI_K )^{-1}$ and $\bA = \bV \bLambda^\top \bD^{-1}$.
The inclusion of pairwise interactions within the response component breaks conditional conjugacy for the latent factors.
To address this issue, the authors suggested updating $\boeta_i$ using the Metropolis-adjusted Langevin algorithm (\textsc{mala}) 
\citep{Grenander_1994, Roberts_1996_Langevin}. 

The same approach can be applied directly to \textsc{jfr} and \textsc{jafar}.
However, the additive structure of \textsc{jafar} once more allows us to cut down the computational burden for the exact expression in equation~\eqref{eq_quadratic_regression}, avoiding $\mathcal{O}\big((\sum_{m=1}^M p_m) \cdot (K+\sum_{m=1}^M K_m)^2 \big)$ costs in favor of $\mathcal{O}\big(\sum_{m=1}^M p_m \cdot (K+K_m)^2 \big)$.
This holds in general when computing response predictions, using the same collapsed representation as in Algorithm~\hyperref[alg:B3]{B3}.
Specifically, the corresponding version of equation~\eqref{eq_quadratic_regression} entails the terms
\begin{equation*}
\begin{aligned}
    \mathbb{E} [\, \widetilde{\btheta}^\top \widetilde{\boeta}_i ] & = \int \bigg( {\btheta}^\top {\boeta}_i +
    \Big( \textstyle{\sum_m} \int \btheta_m^\top \bphi_{mi} \, p \big( \bphi_{mi} \mid \, \boeta_i, \bz_{mi}\}_m \big) \Big) \bigg) p \big( \boeta_i \mid \, \{\bz_{mi}\}_m \big) \\
    & = \btheta^\top \bV \bu_i + \textstyle{\sum_m} \btheta_m^\top \bV_m \bGamma_m \bD_m^{-1} (\bz_{mi} - \bmu_m - \bLambda_m \bV \bu_i)\\
    \mathbb{E} [\, \widetilde{\boeta}_i^\top \widetilde{\bOmega}\widetilde{\boeta}_i ] & = 
    \bu_i^\top \bV \bOmega \bV \bu_i +
    \bu_i^\top \bV \bOmega \big(\textstyle{\sum_m} \bV_m  \bGamma_m^\top \bD_m^{-1} (\bz_{mi} - \bmu_m - \bLambda_m \bV \bu_i ) \big) 
    \\
    & + \big( \textstyle{\sum_m} (\bz_{mi} - \bmu_m - \bLambda_m \bV \bu_i )^\top \bD_m^{-1} \bGamma_m \bV_m \big) \bOmega \big(\textstyle{\sum_m} \bV_m  \bGamma_m^\top \bD_m^{-1} (\bz_{mi} - \bmu_m - \bLambda_m \bV \bu_i ) \big) 
    \\
    & + \operatorname{tr}(\bOmega \bV) + \operatorname{tr}\big(\bOmega \textstyle{\sum_m} \bV_m \big) + \operatorname{tr} \big(\bOmega (\textstyle{\sum_m} \bLambda_m^\top \bD_m^{-1} \bGamma_m \bV_m) \bV (\textstyle{\sum_m} \bV_m \bGamma_m^\top \bD_m^{-1} \bLambda_m ) \big) \\
    & - \operatorname{tr} \big(\bOmega (\textstyle{\sum_m} \bV_m \bLambda_m^\top \bD_m^{-1} \bGamma_m ) \bV \big) \; ,
\end{aligned}
\end{equation*}
where $\widetilde{\btheta}=[\btheta^\top,\btheta_1^\top,\dots,\btheta_M^\top]^\top$ and $\tilde{\boeta}_i = [\boeta_i^\top, \bphi_{1i}^\top, \dots, \bphi_{Mi}^\top]^\top$, and now $\widetilde{\bOmega} \in \Re^{\widetilde{K} \times \widetilde{K}}$ with $\widetilde{K}=K+{\sum_m}K_m$.
Here, $\bV$, $\bu_i$, $\bV_m$, and $\bu_{mi}$ are as in Algorithm~\hyperref[alg:B3]{B3}.
Notice that the standard version of \textsc{jafar} would induce a linear regression of $y_i$ on the feature data, which boils down to removing the quadratic term above.

Similarly to the original contribution of \citet{Ferrari_2021_Interactions}, we could define $\widetilde{\bOmega}$ as a diagonal matrix while still retaining pairwise interactions between the regressors.
In such a case, 
we could rewrite $\widetilde{\bOmega} = \operatorname{diag}\big( \big\{ \{\varsigma_h\}_{h=1}^K, \{\varsigma_{1h}\}_{h=1}^{K_1}, \dots, \{\varsigma_{Mh}\}_{h=1}^{K_M} \big\}\big)$ and set priors
\begin{equation*}
\begin{aligned}
    \varsigma{h} &\sim \mathcal{N}(0, \gamma^2_{h}) \qquad \;
    & \gamma^2_{h} & \sim r_h \; \mathcal{I}nv\mathcal{G}a(a^{\scriptscriptstyle (\Omega)},b^{\scriptscriptstyle (\Omega)}) + (1-r_h) \; \delta_{\gamma^2_{\infty}}. \\
    %%%%%
    \varsigma_{mh} &\sim \mathcal{N}(0, \gamma^2_{m h}) \qquad \;
    & \gamma^2_{m h} & \sim r_{m h} \; \mathcal{I}nv\mathcal{G}a(a^{\scriptscriptstyle (\Omega)},b^{\scriptscriptstyle (\Omega)}) + (1-r_{mh}) \; \delta_{\gamma^2_{\infty}}.
\end{aligned}
\end{equation*}
Through appropriate modifications of the factor modeling structure, the same rationale can be extended to accommodate higher-order interactions, or interactions among the shared factors $\boeta_i$ and the clinical covariates $\br_i$.

% \vspace{-15pt}

\subsubsection{Bayesian B-splines}
To allow for higher flexibility of the response surface, one possibility is to model the continuous outcome with a nonparametric function of the latent variables. 
To limit computational challenges, this can be done using Bayesian B-splines of degree $D$:
\begin{align*}
    f(\boeta_i) = \sum_{h = 1}^K \sum_{d = 1}^{D+2} \bTheta_{h d} \, b_d(\boeta_{i h}) \qquad \qquad 
    f_m(\bphi_{mi}) = \sum_{h = 1}^K \sum_{d = 1}^{D+2} \bTheta_{m h d} \, b_d(\bphi_{i h}),,
\end{align*}
where $b_d(\cdot)$, for $d = 1,\ldots,D+2$, denotes the $d^{th}$ function in a B-spline basis of degree $D$ with natural boundary constraints. Let $\varrho = (\varrho_1,\ldots,\varrho_D)$ be the boundary knots, then $b_1(\cdot)$ and $b_{D+2}(\cdot)$ are linear functions in the intervals $[-\infty, \varrho_1]$ and $[\varrho_{D},+\infty ]$, respectively. In particular, we assume cubic splines (i.e. $D = 3$), but the model can be easily estimated for higher-order splines. 
As before, the update of the shared factors needs to be performed via a Metropolis-within-Gibbs step, without modifying the other steps of the sampler.
In such a case, the prior could be simply set to
\begin{equation*}
\begin{aligned}
    \bTheta_{h d} &\sim \mathcal{N}(0, \psi^2_{h}) \qquad \;
    & \psi^2_{h} & \sim r_h \; \mathcal{I}nv\mathcal{G}a(a^{\scriptscriptstyle (\theta)},b^{\scriptscriptstyle (\theta)}) + (1-r_h) \; \delta_{\psi^2_{\infty}} \\
    \bTheta_{m h d} &\sim \mathcal{N}(0, \psi^2_{mh}) \qquad \;
    & \psi^2_{mh} & \sim r_{mh} \; \mathcal{I}nv\mathcal{G}a(a^{\scriptscriptstyle (\theta)},b^{\scriptscriptstyle (\theta)}) + (1-r_{mh}) \; \delta_{\psi^2_{\infty}}.\; . 
\end{aligned}
\end{equation*}

\subsection{Categorical and count outcomes: \textsc{glm}s factor regression}

\textsc{jfr} and \textsc{jafar} can be modified to accommodate non-continuous outcomes $y_i$, while still allowing for deviation from linearity assumptions via the quadratic regression setting presented above.
For notational simplicity, we focus on \textsc{jfr} and its concatenated elements $\bz_i = [ \bz_{1 i}^\top, \dots, \bz_{m i}^\top]^\top$ and $\bLambda = [ \bLambda_1, \dots, \bLambda_{m}]$; analogous extensions for \textsc{jafar} also benefit from the computational reductions discussed above.
For instance, binary responses can be trivially modeled via a probit link $ y_i \sim \mathcal{B}er(\varphi_i)$ with $\varphi_i = \Phi(\theta^\top \boeta_i + \boeta_i^\top \bOmega \, \boeta_i)$.
Except for the latent factors, conditional conjugacy is preserved by appealing to a well-known data augmentation in terms of latent variable $q_i \in \Re$ \citep{Albert_1993}, such that $y_i=1$ if $q_i >0 $ and $y_i =0$ if $q_i \leq 0 $. \\
More generally, in the remainder of this section, we show how to extend the same rationale to generalized linear models (\textsc{glm}) with logarithmic link and responses in the exponential families. 
In doing so, we also compute expressions for induced main and interaction effects, allowing for a straightforward interpretation of the associated coefficients.

\numberwithin{Theorem}{section}

\subsubsection{Factor regression with count data}
In \textsc{glm}s under log-link, the logarithmic function is used to relate the linear predictor $\bbeta^\top \br_i$ to the conditional expectation of $y_i$ given the covariates $\br_i$, such that $\operatorname{log} \big( \mathbb{E}[y_i | \br_i] \big) = \bbeta^\top \br_i$.
Two renowned \textsc{glm}s for count data are the Poisson and the Negative-Binomial models.
Defining $\varphi_i$ as the mean parameter for the $i^{th}$ observation $\varphi_i = \mathbb{E}[y_i | \br_i] = e^{\bbeta^\top \br_i}$, such two alternatives correspond to $(y_i| \br_i) \sim \mathcal{P}oisson(\varphi_i)$ and $(y_i| \br_i) \sim \mathcal{N}eg\mathcal{B}in \big( \kappa / (\varphi_i + \kappa), \kappa \big)$, for some $\kappa \in [0,1]$.
A main limitation of the Poisson distribution is the fact that the mean and variance are equal, which motivates the use of negative-binomial regression to deal with over-dispersed count data.
In both scenarios, we can integrate the \textsc{glm} formulation in  the quadratic latent factor structure presented above 
\begin{equation*}
    \operatorname{log} \big( \mathbb{E}[y_i | \boeta_i] \big) = \btheta^\top \boeta_i + \boeta_i^\top \bOmega \, \boeta_i 
\end{equation*}
Accordingly, it is easy to show the following. 
\begin{Proposition}\label{prop_quad_glm}
Marginalizing out all latent factors in the quadratic \textsc{glm} extension of \textsc{jafar}, both shared and view-specific ones, it holds that
\begin{gather*}
    \mathbb{E}[y_i | \bz_i] = \sqrt{\, |\bV'| \, / \, |\bV| \,} \, \exp \left( \frac{1}{2} \btheta^\top \bV' \btheta +  \btheta_X^\top \bz_i + \bz_i^\top \bOmega_X \bz_i \right)
\end{gather*}
where $\btheta_X^\top = \btheta^\top (\bI_K - 2 \bV \bOmega )^{-1} \bA$, $\bOmega_X = \frac{1}{2} \bA^\top \bV^{-1} \big( (\bI_K - 2 \bV \bOmega )^{-1} - \bI_K \big) \bA$ and $\bV' = (\bV^{-1} - 2 \bOmega)^{-1}$. As before, $\bV= ( \bLambda^\top \bD^{-1} \bLambda + \bI_K )^{-1}$ and $\bA = \bV \bLambda^\top \bD^{-1}$ comes for the full-conditional posterior of the shared factors $\boeta_i \mid \bz_i \sim \mathcal{N}_K ( \bA \, \bz_i, \bV)$, after marginalization of the view-specific factors.
\end{Proposition}
This allows us to estimate quadratic effects with high-dimensional correlated predictors in regression settings with count data. 
As before, the composite structure of \textsc{jafar} can be leveraged in the bottleneck computation of the large matrix $\bLambda^\top \bD^{-1} \bLambda$, enabling a substantial reduction in computational cost via a convenient decomposition.

\subsubsection{Exponential family responses}
We consider here an even more general scenario requiring only that the outcome distribution belongs to the exponential family
\begin{align*}
    p(y_i | \varsigma_i) = \exp \big( \varsigma_i \cdot T(y_i) - U(\varsigma_i) \big) \; ,
\end{align*}
where $\varsigma_i$ is the univariate natural parameter and $T(y_i)$ is a sufficient statistic.
Accordingly, we generalize Gaussian linear factor models and set $\varsigma_i = \btheta^\top \boeta_i + \boeta_i^\top  \bOmega \, \boeta_i$. As before, $\varphi_i = \mathbb{E}[y_i | \boeta_i] = g^{-1}(\btheta^\top \boeta_i + \boeta_i^\top  \bOmega \, \boeta_i)$, where $g(\cdot)$ is a model-specific link function. Our goal is to compute the expectation of $y_i$ given $\bz_i$ after integrating out all latent factors
\begin{align*}
    \mathbb{E}[y_i | \bz_i ] & = \mathbb{E} \big[ \mathbb{E}[y_i | \boeta_i] | \bz_i \big] = \mathbb{E} [ g^{-1}(\btheta^\top \boeta_i + \boeta_i^\top  \bOmega \, \boeta_i)| \bz_i] \\
    & = \int g^{-1}(\btheta^\top \boeta_i + \boeta_i^\top  \bOmega \, \boeta_i) p(\boeta_i | \bz_i) d\boeta_i.
\end{align*}
In general, this represents the expectation of the natural parameter conditional on $\bz_i$ for any distribution within the exponential family.
Endowing the stacked transformed features $\bz_i$ with the addictive factor model above, i.e. $\bz_{m i} \sim \mathcal{N}_{p_m} \big( \bmu_m + \bLambda_m \boeta_i, \bGamma_m \bGamma_m^\top + \operatorname{diag}(\bsigma_m^2) \big)$, we have that $p(\boeta_i | \bz_i)$ is pdf of a normal distribution with mean $\bA \bz_i$ and variance $\bV$ (see Proposition~\ref{prop_quad_glm}). In this case, the above integral can be solved when $g^{-1}(\cdot)$ is the identity function, as in linear regression, or the exponential function, as in regression for count data or survival analysis.
On the contrary, when we are dealing with a binary regression and $g^{-1}(\cdot)$ is equal to the logit, the above integral does not have an analytical solution. However, recalling that in such a case $\varphi_i$ represents the probability of success, we can integrate out the latent variables and compute the expectation of the log-odds conditional on $\bz_i$
\begin{align*}
    \mathbb{E}\left[\text{log} \left( \frac{\varphi_i}{1-\varphi_i}\right) \Big| \, \bz_i \right] =  \mathbb{E}[\btheta^\top \boeta_i + \boeta_i^\top  \bOmega \, \boeta_i | \bz_i] = (\btheta^\top \bA ) \, \bz_i + \bz_i^\top (\bA^\top \bOmega \, \bA ) \, \bz_i + \operatorname{tr}(\bOmega \bV) .
\end{align*}

% \vspace{-15pt}

%%%$$$$$$$$$$$$$$$$$$$$$$$$$$$$$$$$$$$$$$$$$$$$$$$$$$
\section{Realistic data simulation mechanism}\label{app_realistic_simulations}

\setcounter{table}{0}
\setcounter{figure}{0}
\setcounter{equation}{0}
\setcounter{algocf}{0}

\begin{algorithm}[b]
\caption{Generation of realistic loading matrices $\textcolor{white}{\big(}$}
~\\[-3pt]
$\bLambda = \bzero_{p \times K}$ \\[3pt]
\For{$g=1,\dots,G$}{
$\widetilde{\mu}_{g} \sim f_+$ \hspace{38pt} \textcolor{gray}{[hyper-loadings magnitude]} \\[3pt]
$\mu_g = (-1)^g \widetilde{\mu}_{g}$ \hspace{10pt} \textcolor{gray}{[signed hyper-loadings]}\\[3pt]
\For{$h=1,\dots,K$}{
$u_{g h} \sim \mathcal{B}ern(\pi^{(g)})$ \hspace{10pt} \textcolor{gray}{[group-wise sparsity]} \\[3pt] 
$s_{g h} \sim \mathcal{B}ern(\pi^{(s)})$ \hspace{12pt} \textcolor{gray}{[group-wise sign switch]} 
}
}
\For{$j=1,\dots,p$}{
$g(j) \sim \mathcal{C}at_G \big(\{\pi_g\}_{g=1}^G \big)$ \hspace{10pt} \textcolor{gray}{[group assignment]} \\[3pt]
\For{$h=1,\dots,K$}{
$r_{j h} \sim \mathcal{B}ern(\pi^{(e)})$ \hspace{61pt} \textcolor{gray}{[entry-wise sparsity]} \\[3pt]
$\ell^{(1)}_{j h} \sim \mathcal{N}(\mu_{g(j)}/\sqrt{K},v^2_{o}/K)$  \hspace{10pt}  \textcolor{gray}{[main signal]} \\[3pt]
$\ell^{(0)}_{j h} \sim \mathcal{N}(0,r_{damp} v^2_{o}/K)$  \hspace{27pt}  \textcolor{gray}{[spurious signal]} \\[3pt]
$\bLambda_{j h} = u_{g(j) h} \cdot (2s_{g(j) h}-1) \cdot r_{j h} \cdot \ell^{(1)}_{j h} + \ell^{(0)}_{j h} $ \hspace{10pt} \textcolor{gray}{[composite signal]} 
}
}
return $\bLambda$\\[2pt]
\label{algo_loading_sim}
\end{algorithm}

In this section, we describe the proposed mechanism to generate loading matrices that induce realistic block-structured correlations, and in general, the data-generating mechanism for the simulations from Section~\ref{sec_simulations}.
This represents a significant improvement in targeting realistic simulated data, compared to many studies in the literature.
Focusing on a single loading matrix $\bLambda \in \Re^{p \times K}$ for ease of notation,
\citet{Ding_2022_coopL} set $\bLambda = [\bI_{K},\bzero_{K \times (p-K)}]^\top$, which gives $\bLambda \bLambda^\top = \operatorname{block-diag}(\{\bI_{K},\bzero_{K \times (p-K)}\})$.
\citet{samorodnitsky2024bayesian} samples independently $\bLambda_{j h} \sim \mathcal{N}(0,1)$, so that $\E\big[(\bLambda \bLambda^\top)_{j j'}\big] = \delta_{j,j'} \cdot K $.
\citet{poworoznek2025_MatchAlign} enforce a simple sparsity pattern in the loadings, dividing the $p$ features into K groups and sampling $\bLambda_{j h} \sim \delta_{g(j),h} \mathcal{N}(0,v^2_{slab}) + (1-\delta_{g(j),h}) \mathcal{N}(0,v^2_{spike})$, for some $v^2_{slab} \gg v^2_{spike}$ and representing by $g(j)$ the group assignment.
This still gives $\E\big[(\bLambda \bLambda^\top)_{j j'}\big] = \delta_{j,j'} \cdot \big( v^2_{slab} + (K-1)\cdot v^2_{spike}\big)$.
Although the generation of a specific loading matrix entails single samples rather than expectations, the induced correlation matrices are not expected to present any meaningful structure.

To overcome this issue, we further leverage the grouping of the features, allowing each group to load on multiple latent factors and centering the entries of each group around some common hyper-loading $\mu_g$, for $g=1,\dots, G$.
To induce blocks of positive and negatively correlated features, we propose setting $\mu_g = (-1)^g \widetilde{\mu}_g$, with $\widetilde{\mu}_g$ sampled from a density $f_+$ with support on the positive real line. 
Our default suggestion is to set $f_+$ to be a beta distribution $\mathcal{B}e ( 5, 3 )$.
Conditioned on such hyper-loadings and the group assignments, we sample the loading entries independently from $\bLambda_{j h} \sim \mathcal{N}(\mu_{g(j)}/\sqrt{K},v^2_{o}/K)$, resulting in  $\E\big[(\bLambda \bLambda^\top)_{j j'}\big] = (-1)^{g(j)} (-1)^{g(j')} \widetilde{\mu}_{g(j)} \widetilde{\mu}_{g(j')} + \delta_{j,j'}  v^2_{o}$. 
This naturally translates into blocks of features with correlations of alternating signs and different magnitudes.

The core structure above can be complemented with further nuances to recreate more realistic patterns.
This includes group-wise sign permutation, introducing entry-wise and group-wise sparsity, and the addition of a layer of noise loadings $\mathcal{N}(0,r_{damp} v^2_{o}/K)$ to avoid exact zeros.
Similarly, view-wise sparsity can be imposed on the resulting shared loadings to achieve composite activity patterns in the respective components of the model.
The resulting generation procedure for the loading matrices is summarized in Algorithm~\ref{algo_loading_sim}.
In our simulation studies from Section~\ref{sec_simulations}, we set $v^2_{o}=0.1$ and $r_{damp}=1e^{-2}$.

Given a sample of the loading matrices $\bLambda_m$ and $\bGamma_m$ as above, we generate the data from a factor model with the same structure as \textsc{jafar}.
We sample the target signal-to-noise ratios $\{ \operatorname{snr}_{m j} \}_{j=1}^{p_m}$ from an inverse gamma distribution $\mathcal{I}nv\mathcal{G}a ( 10, 30 )$, and set accordingly each idiosyncratic variance $\bsigma_{m j}^2 $ to $(\bLambda_m \bLambda_m^\top + \bGamma_m \bGamma_m^\top)_{jj}/ \operatorname{snr}_{m j}$, $\forall j=1,\dots,p_m$.
Similarly to the generation of loading matrices, the absolute value of the active response coefficients $\theta_h$ was sampled from a beta distribution $\mathcal{B}e ( 5, 3 )$, and their signs were randomly assigned with equal probability. The response variance $\sigma_y^2$ was adjusted such that the signal-to-noise ratio $\btheta^\top \btheta/\sigma_y^2$ equals 1.

% \vspace{-15pt}

% ||||||||||||||||||||||||||||||||||||||||||||||||||
\section{Simulation Studies: further results}\label{app_sim_extra_ris}

\setcounter{table}{0}
\setcounter{figure}{0}
\setcounter{equation}{0}
\setcounter{algocf}{0}

In the present section, we provide further evidence of the performance of the proposed methodology on simulated data.

\subsection{Separation of shared and view-specific components in unsupervised settings}\label{app_extra_sim_unsupervised}

We begin by complementing the simulation in Section~\ref{sec_sim_supervised} with analogous results on real data.
In particular, we select a random subset of $\{100,200,300\}$ features for each of the three omics layers in the application from Section~\ref{sec_application}, focusing on one single replicate.
We run all Gibbs
samplers under the same hyperparameters of Section~\ref{sec_sim_unsupervised}, but for a total of $T_\textsc{mcmc} = 20000$ iterations, with a burn-in of $T_\textsc{burn-in} = 15000$ steps, still thinning every $T_\textsc{thin} = 10$ samples for memory efficiency.
In Table~\ref{tab_sec2_real_data} we report the inferred number of factors, which shows a very similar pattern to the one observed in simulated data.
Under the \textsc{naive} prior, \textsc{jafar} incorrectly assigns many study-specific factors to the shared component. Under the \textsc{full-d} prior, \textsc{jafar} overestimates the shared rank even more severely, although avoiding misallocation of specific factors. By contrast, \textsc{jfr} under \textsc{i-cusp} and \textsc{jafar} under \textsc{d-cusp} produce self-consistent estimates of factor activity, both in overall rank and in the number of effectively shared factors.

\begin{table}[ht!]
\centering
\begin{tabular}{c|c|rrr|r|rrr|r
}
\toprule
Model & Prior & \multicolumn{1}{l}{$\Gamma_{1}$} & \multicolumn{1}{l}{$\Gamma_{2}$} & \multicolumn{1}{l|}{$\Gamma_{3}$} & \multicolumn{1}{l|}{$\Lambda_{tot}$} & \multicolumn{1}{l}{$\Lambda_{1}$} & \multicolumn{1}{l}{$\Lambda_{2}$} & \multicolumn{1}{l|}{$\Lambda_{3}$} &  \multicolumn{1}{l}{$\Lambda_{\text{one-only}}$} \\ 
 \midrule
\textsc{jafar} & \textsc{d-cusp} & 9.0 & 11.0 & 10.0 & 4.0 & 1.0 & 4.0 & 2.7 &0.3 \\[2pt]
& \textsc{naive} & 7.0 & 7.9 & 6.0 & 18.0 & 3.0 & 7.7 & 6.0 & 14.7 \\[2pt]
& \textsc{full-d} & 4.0 & 4.0 & 4.0 & 35.0 & 17.6 & 23.4 & 29.0 & 0.0\\[2pt]
 \cmidrule{1-10}
\textsc{jfr} & \textsc{i-cusp} & -\hspace{5pt} & -\hspace{5pt} & -\hspace{5pt} & 38.0 & 10.5 & 14.0 & 13.0 & 35.5 \\
\bottomrule
\end{tabular}
\vspace{10pt}
\caption{Number of active factors inferred under different model-prior combinations in random features subsets of dimensions $\{100,200,300\}$ from the Time-to-labor-onset data analyzed in Section~\ref{sec_application}.}
\label{tab_sec2_real_data}
\end{table}

\subsection{Predictive accuracy and dependence reconstruction in supervised settings}\label{app_extra_sim_supervised}

\begin{figure}[ht!]
   \centering
\includegraphics[width=0.45\linewidth]{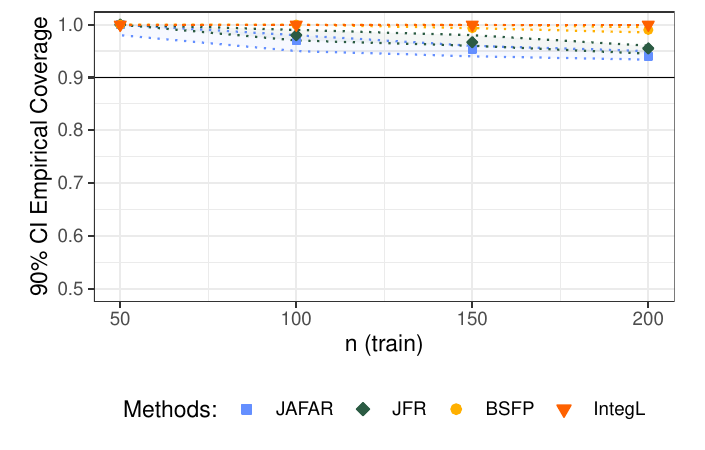}
    \put(-100,142){\makebox(0,15){Train Set}}
    \includegraphics[width=0.45\linewidth]{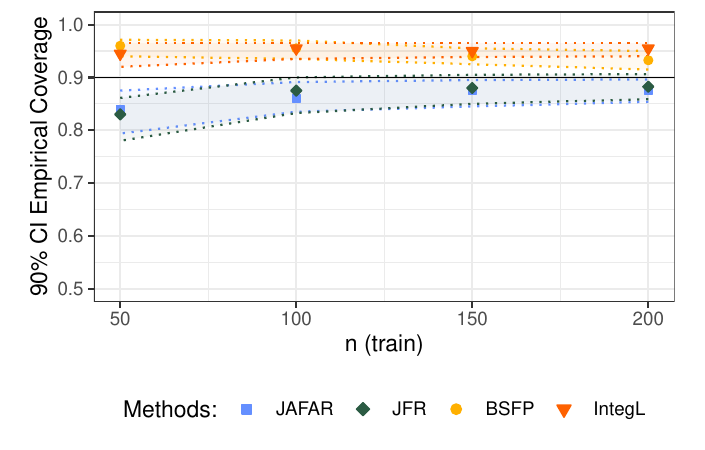}
    \put(-100,142){\makebox(0,15){Test Set}}
   % \vspace{-15pt}
   \caption{Empirical coverage of the $90\%$ predictive intervals on the training sets (left)  and testing sets (right) of simulated data from Section~\ref{sec_sim_supervised}.
    The x-axis values correspond to increasing training set sizes.
    Interior points and band edges correspond to quartiles across 100 independent replicates.
    The horizontal gray lines correspond to the correct coverage.}
   \label{fig_sim_response_CI}
\end{figure}

Secondly, we complement the results from Section~\ref{sec_sim_supervised} with the associated uncertainty quantification.
Figure~\ref{fig_sim_response_CI} shows that all the methods considered incur an overcoverage when predicting in-sample, although \textsc{jfr} and \textsc{jafar} show hints of convergence to the correct coverage for increasing $n$.
In out-of-sample predictions, \textsc{bsfp} and \texttt{IntegLearn} still incur overcoverage, while both proposed methodologies slightly undercover, but include the correct coverage in the quartiles over replication for even moderate $n$.

\begin{table}[hb!]
\centering
\begin{tabular}{ccrrrr}
\toprule
\multirow{1}{*}{\centering n} & \multirow{1}{*}{\centering Method} 
& \multicolumn{1}{c}{$\text{N}_\text{MCMC}$} & \multicolumn{1}{c}{$\text{N}_\text{Burn-in}$} & \multicolumn{1}{c}{$K_\text{tot}$} & \multicolumn{1}{c}{Time [min]} \\[2pt]
\toprule
\multirow{4}{*}{50} 
 & \textsc{jafar} & 20000 & 15000 & 34.0 & \num{\fpeval{415.1/60}}
 \\
 & \textsc{jfr}   & 20000 & 15000 & 31.0 & \num{\fpeval{640.3/60}} \\
 & \textsc{bsfp}  & 6000  & 3000  & 37.0 & \num{\fpeval{2338.7/60}} \\
 & \textsc{bip}   & 6000  & 3000  & 20.0 & \num{\fpeval{780.1/60}} \\
\cmidrule{1-6}
\multirow{4}{*}{100} 
 & \textsc{jafar} & 20000 & 15000 & 38.0 & \num{\fpeval{468.0/60}} \\
 & \textsc{jfr}   & 20000 & 15000 & 34.0 & \num{\fpeval{788.9/60}} \\
 & \textsc{bsfp}  & 6000  & 3000  & 50.0 & \num{\fpeval{3337.7/60}} \\
 & \textsc{bip}   & 6000  & 3000  & 20.0 & \num{\fpeval{2292.7/60}} \\
\cmidrule{1-6}
\multirow{4}{*}{150} 
 & \textsc{jafar} & 20000 & 15000 & 40.0 & \num{\fpeval{555.4/60}} \\
 & \textsc{jfr}   & 20000 & 15000 & 37.0 & \num{\fpeval{890.5/60}} \\
 & \textsc{bsfp}  & 6000  & 3000  & 55.0 & \num{\fpeval{3382.1/60}} \\
 & \textsc{bip}   & 6000  & 3000  & 20.0 & \num{\fpeval{4200.6/60}} \\
\cmidrule{1-6}
\multirow{4}{*}{200} 
 & \textsc{jafar} & 20000 & 15000 & 40.5 & \num{\fpeval{700.4/60}} \\
 & \textsc{jfr}   & 20000 & 15000 & 38.0 & \num{\fpeval{966.1/60}} \\
 & \textsc{bsfp}  & 6000  & 3000  & 57.0 & \num{\fpeval{4372.1/60}} \\
 & \textsc{bip}   & 6000  & 3000  & 20.0 & \num{\fpeval{5167.6/60}} \\
\bottomrule
\end{tabular}
\vspace{10pt}
\caption{Runtime versus number of samples and sum of latent dimensions for different sample sizes in the simulation studies in Section~\ref{sec_sim_supervised}.}
\label{tab_sec3_times}
\end{table}

We also assess computational performance and \textsc{mcmc} mixing of the Bayesian factor model considered. Table~\ref{tab_sec3_times} reports the overall runtimes for different sample sizes, complemented with information on the number of
\textsc{mcmc} samples and the total number of latent factors. 
Adjusting for this information, the computational improvements of \textsc{jfr} and \textsc{jafar} over \textsc{bsfp} and \textsc{bip} are even more significant.
The simulations were run on a 13-inch MacBook Pro (2018) with a 2.7 GHz quad-core Intel Core i7 processor and 16Gb RAM (macOS 14.6.1). 
For reference, \texttt{CoopLearn} and \texttt{IntegLearn} run in less than 3 seconds and roughly 1.5 minutes on the same machine.
In Table~\ref{tab_sec3_ess} we further analyze the \textsc{ess} for the same methodologies. 
Such an analysis was not possible on \textsc{bip}, whose code only reports point estimates.
As summary statistics, we focus on the \textsc{ess} of the idiosyncratic noise variance for the response and of the induced marginal variances on the predictors, averaging over each view's features.

\begin{table}[!h]
\centering
\begin{tabular}[t]{ccccrrrrr}
\toprule
\multirow{2}{*}{\centering n} & 
\multirow{2}{*}{\centering Method} & 
\multirow{2}{*}{\centering Samples} & 
\multirow{2}{*}{\centering Time [s]} & 
\multicolumn{4}{c}{{Median ESS [\%]}} \\
\cmidrule(lr){5-8}
& & & & 
\multicolumn{1}{c}{$\text{var}(X_1)$} & 
\multicolumn{1}{c}{$\text{var}(X_2)$} & 
\multicolumn{1}{c}{$\text{var}(X_3)$} & 
\multicolumn{1}{c}{$\sigma^2_y$} \\[2pt]
\toprule
\multirow{3}{*}{50} & {\textsc{jafar}} & {500} & \num{\fpeval{{103.8}/60}} & {100.0} & {100.0} & {100.0} & {47.1}\\
 & \textsc{jfr} & 500 & \num{\fpeval{160.1/60}} & 98.1 & 87.3 & 100.0 & 81.2\\
 & {\textsc{bsfp}} & {300} & \num{\fpeval{{1169.4}/60}} & {100.0} & {100.0} & {100.0} & {23.2}\\
 \cmidrule{1-8}
\multirow{3}{*}{100} & \textsc{jafar} & 500 & \num{\fpeval{117.0/60}} & 100.0 & 100.0 & 100.0 & 68.7\\
 & {\textsc{jfr}} & {500} & \num{\fpeval{{197.2}/60}} & {86.4} & {84.8} & {89.3} & {100.0}\\
 & \textsc{bsfp} & 300 & \num{\fpeval{1668.9/60}} & 100.0 & 100.0 & 100.0 & 15.8\\
 \cmidrule{1-8}
\multirow{3}{*}{150} & {\textsc{jafar}} & {500} & \num{\fpeval{{138.8}/60}} & {100.0} & {100.0} & {100.0} & {82.1}\\
 & \textsc{jfr} & 500 & \num{\fpeval{222.6/60}} & 84.3 & 82.3 & 85.9 & 100.0\\
 & {\textsc{bsfp}} & {300} & \num{\fpeval{{1691.0}/60}} & {100.0} & {100.0} & {100.0} & {12.1}\\
 \cmidrule{1-8}
\multirow{3}{*}{200} & \textsc{jafar} & 500 & \num{\fpeval{175.1/60}} & 100.0 & 100.0 & 100.0 & 92.9\\
 & {\textsc{jfr}} & {500} & \num{\fpeval{{241.5}/60}} & {82.5} & {80.6} & {84.5} & {100.0}\\
 & \textsc{bsfp} & 300 & \num{\fpeval{2186.0/60}} & 100.0 & 100.0 & 100.0 & 8.9\\
\bottomrule
\end{tabular}
\vspace{10pt}
\caption{\textsc{ess} versus number of effective samples and runtime for different sample sizes in the simulation studies in Section~\ref{sec_sim_supervised}.}
\label{tab_sec3_ess}
\end{table}

\begin{figure}[hp!]
   \centering
   \begin{minipage}{0.25\textwidth}
       \centering
       \includegraphics[width=\linewidth]{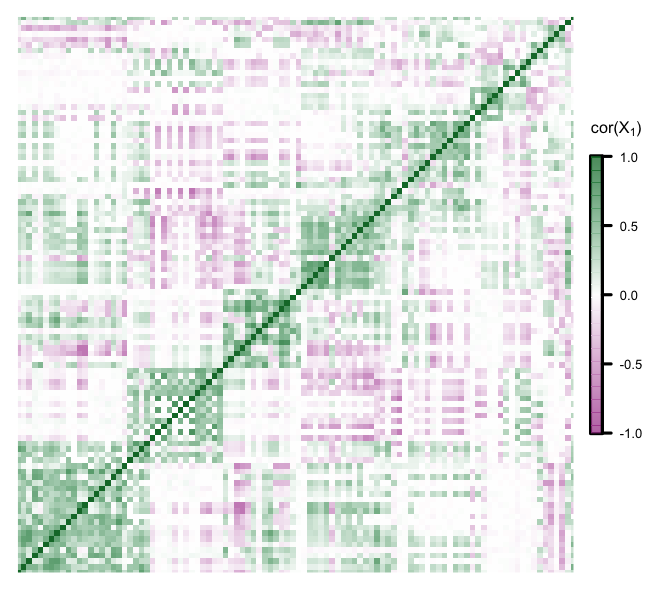}
   \put(-65,115){\makebox(0,0){$m=1$}}
   \put(-130,50){\rotatebox{90}{\makebox(0,0){Truth}}}
   \end{minipage}
   \begin{minipage}{0.25\textwidth}
       \includegraphics[width=\linewidth]{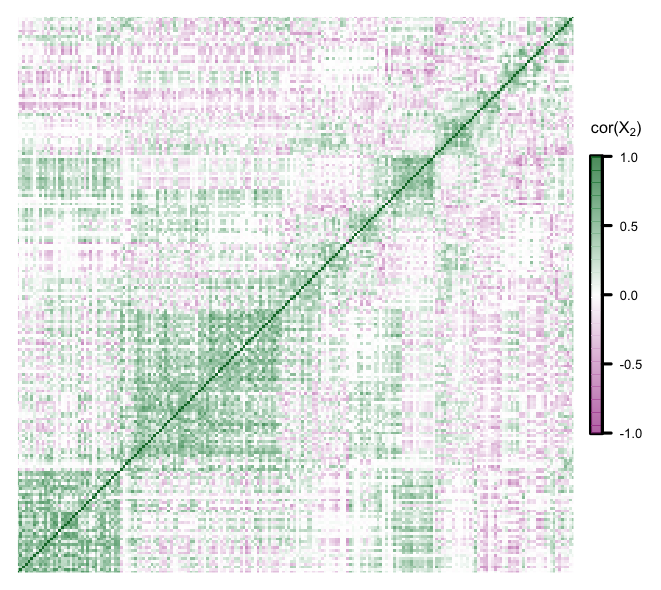}
   \put(-65,115){\makebox(0,0){$m=2$}}
   \end{minipage}
   \begin{minipage}{0.25\textwidth}
       \centering
       \includegraphics[width=\linewidth]{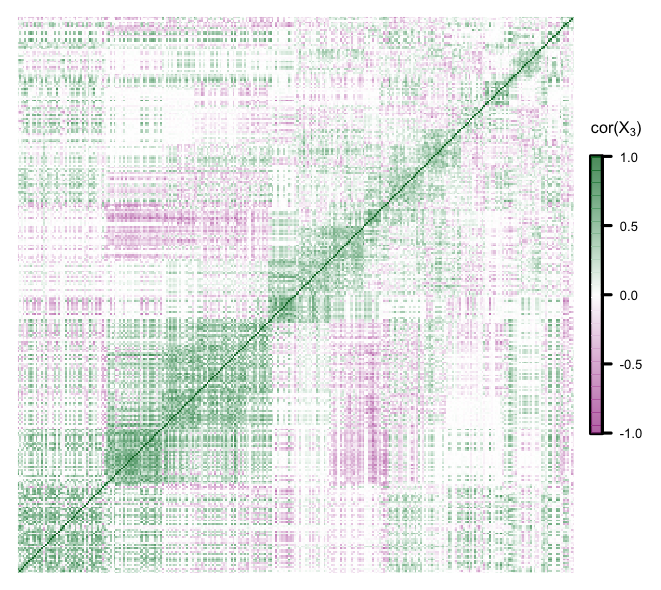}
   \put(-65,115){\makebox(0,0){$m=3$}}
   \end{minipage}
   \\
   % % % % % % % % % % % % % % % 
   \begin{minipage}{0.25\textwidth}
       \centering
       \includegraphics[width=\linewidth]{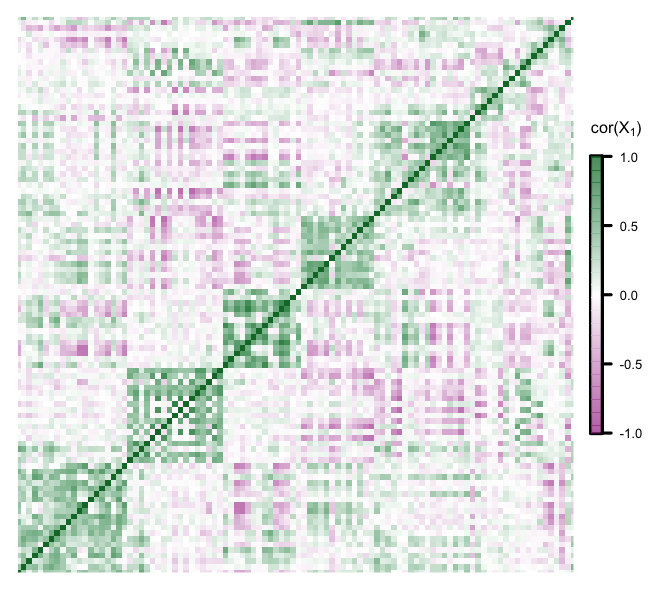}
   \put(-130,50){\rotatebox{90}{\makebox(0,0){\textsc{jfr\textit{}}}}}
   \end{minipage}
   \begin{minipage}{0.25\textwidth}
       \centering
       \includegraphics[width=\linewidth]{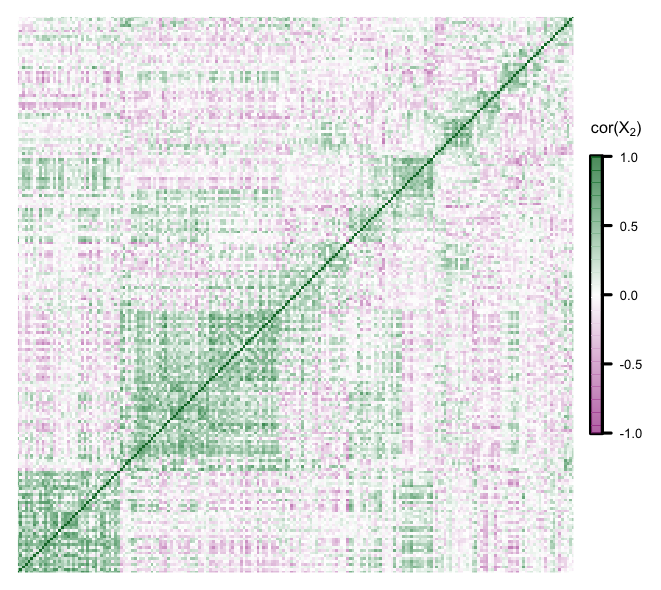}
   \end{minipage}
   \begin{minipage}{0.25\textwidth}
       \centering
       \includegraphics[width=\linewidth]{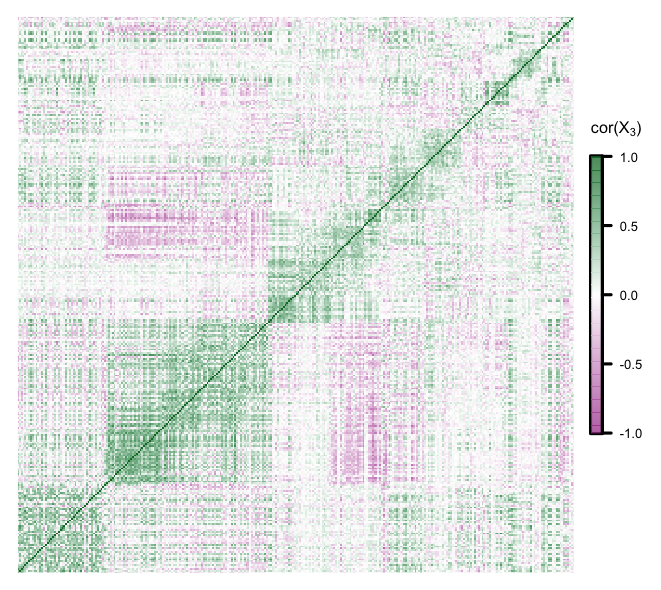}
   \end{minipage}
   \\
   % % % % % % % % % % % % % % % 
   \begin{minipage}{0.25\textwidth}
       \centering
       \includegraphics[width=\linewidth]{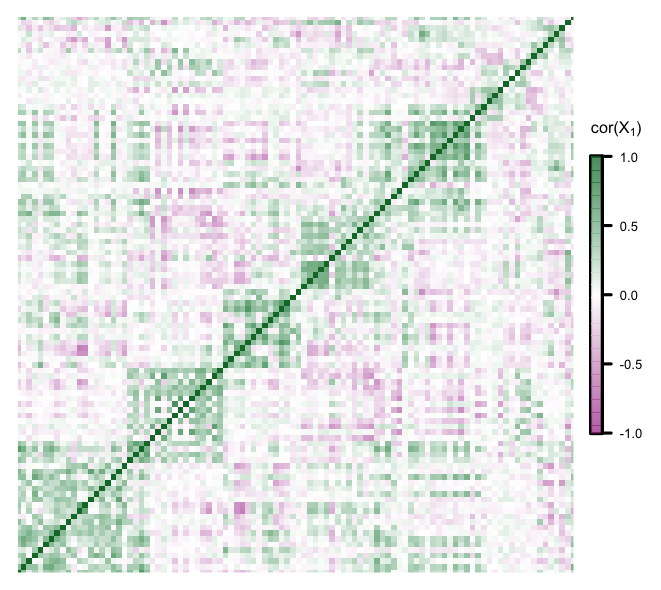}
   \put(-130,50){\rotatebox{90}{\makebox(0,0){\textsc{jafar}}}}
   \end{minipage}
   \begin{minipage}{0.25\textwidth}
       \centering
       \includegraphics[width=\linewidth]{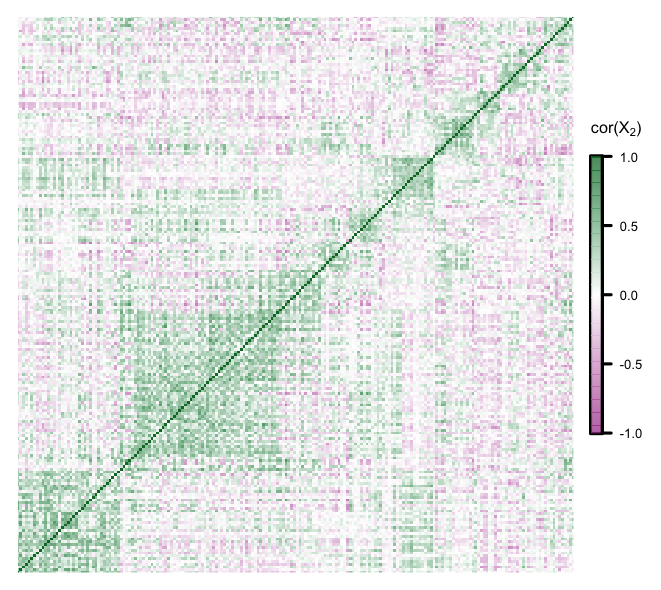}
   \end{minipage}
   \begin{minipage}{0.25\textwidth}
       \centering
       \includegraphics[width=\linewidth]{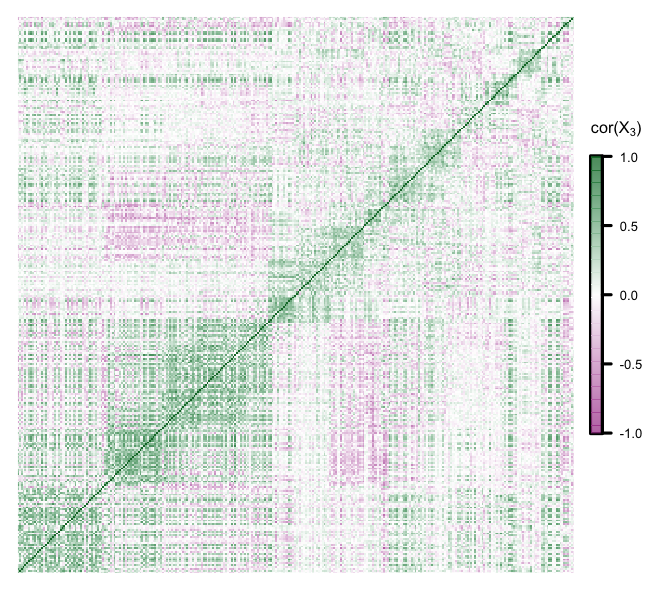}
   \end{minipage}
   \\
   % % % % % % % % % % % % % % % 
   \begin{minipage}{0.25\textwidth}
       \centering
       \includegraphics[width=\linewidth]{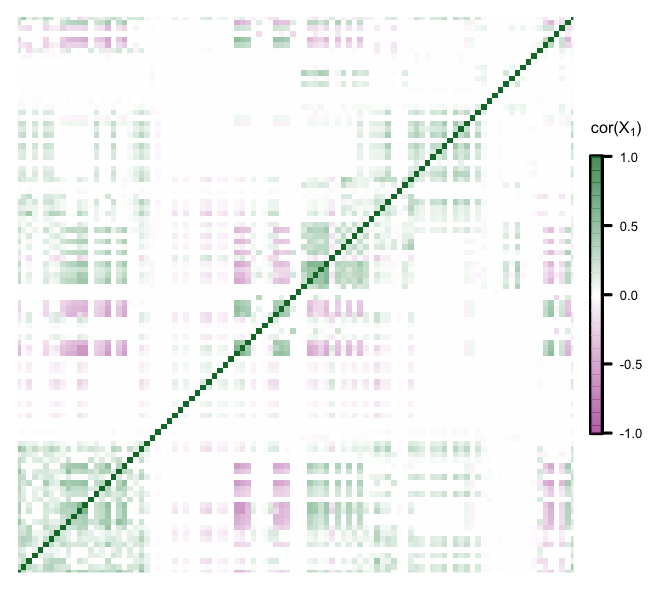}
   \put(-130,50){\rotatebox{90}{\makebox(0,0){\textsc{bip}}}}
   \end{minipage}
   \begin{minipage}{0.25\textwidth}
       \centering
       \includegraphics[width=\linewidth]{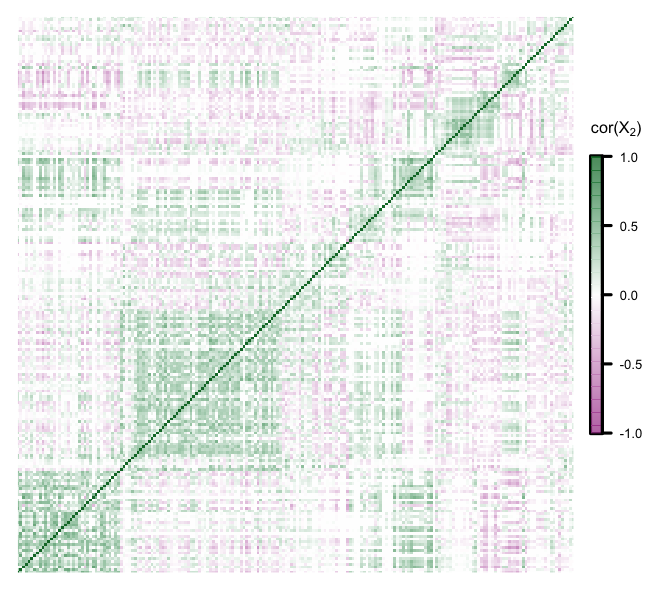}
   \end{minipage}
   \begin{minipage}{0.25\textwidth}
       \centering
       \includegraphics[width=\linewidth]{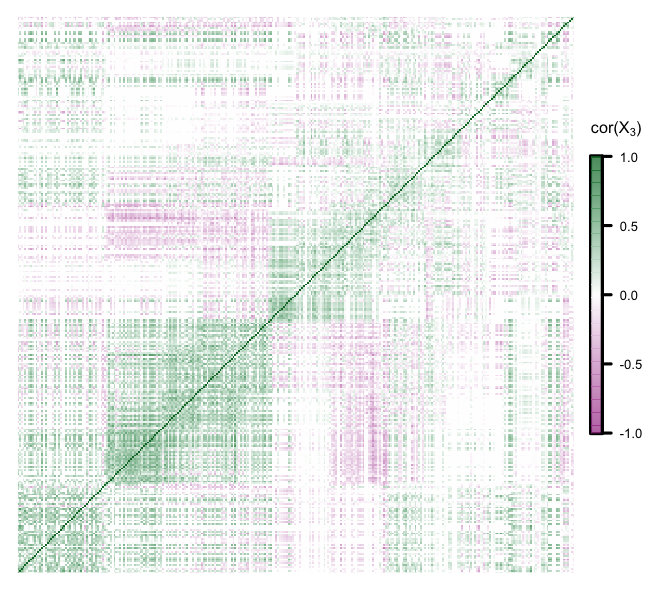}
   \end{minipage}
   \\
   % % % % % % % % % % % % % % % 
   \begin{minipage}{0.25\textwidth}
       \centering
       \includegraphics[width=\linewidth]{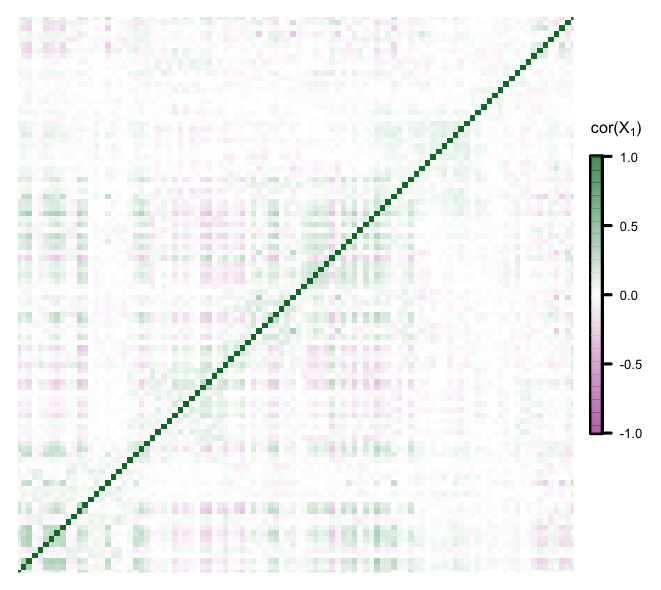}
   \put(-130,50){\rotatebox{90}{\makebox(0,0){\textsc{bsfp}}}}
   \end{minipage}
   \begin{minipage}{0.25\textwidth}
       \centering
       \includegraphics[width=\linewidth]{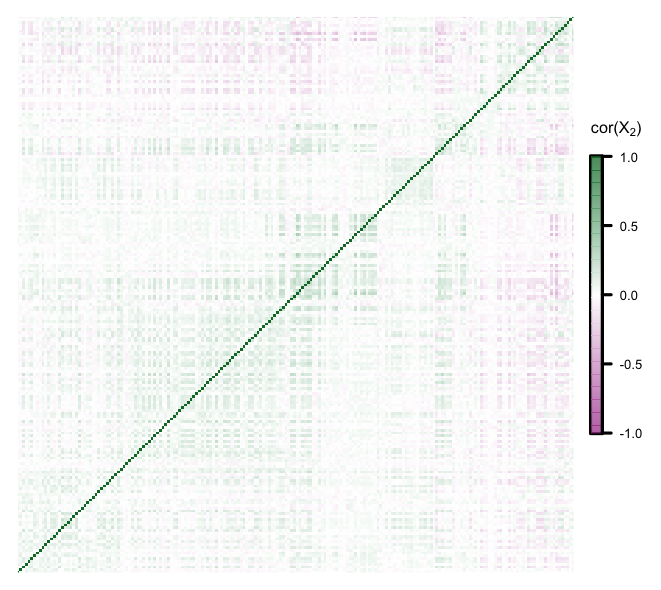}
   \end{minipage}
   \begin{minipage}{0.25\textwidth}
       \centering
       \includegraphics[width=\linewidth]{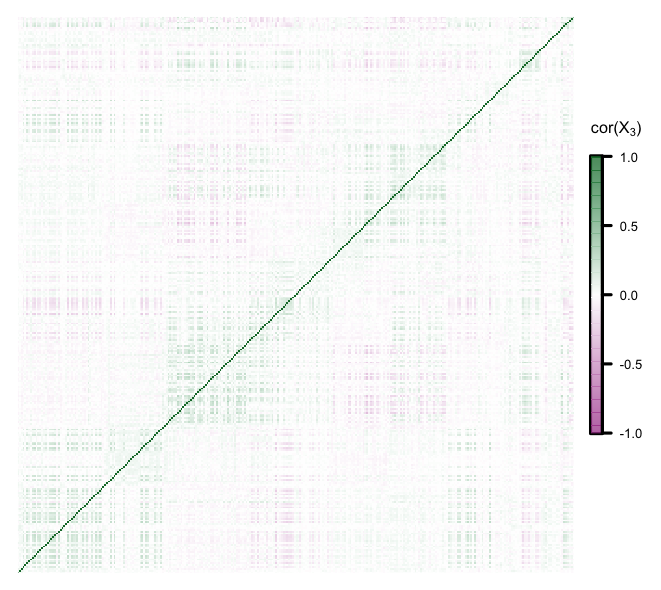}
   \end{minipage}
   \caption{Within-view correlation matrices in one replicate of the considered simulated data for $n=200$ from Section~\ref{sec_sim_supervised}.
   }
   \label{fig_sim_inferred_cor_emp}
\end{figure}

To provide more insight into feature structure learning, we further break down the results for one of the $100$ replicates for $n=200$ from Section~\ref{sec_sim_supervised}.
In Figure~\ref{fig_sim_inferred_cor_emp}, we report the correlation matrices for each view.
The true underlying covariances reflect the realistic block structure obtained via our original simulation setup for the loading matrices, detailed in Appendix~\ref{app_realistic_simulations}.
The covariance for \textsc{bip} corresponds to the point estimate provided. 

Since \textsc{bsfp} does not directly provide the inferred covariances, we obtained the latter from the empirical correlations computed on independent samples from $p(\bX_m \mid \{\bX_{m'}\}_{m' \neq m})$.
Specifically, here we consider the average correlation matrix $\frac{1}{T_{\textsc{mcmc}}} \sum_{t=1}^{T_{\textsc{mcmc}}}\operatorname{cor}(\bX_m^{(t)})$ over \textsc{mcmc} samples from the full conditional $\bX_m^{(t)} \sim p(\bX_m \mid \{\bX_{m'}\}_{m' \neq m})$ of the corresponding view given all others (response excluded).
For a fair comparison, we computed the covariances inferred by \textsc{jafar} in the same way, although our accompanying code provides the closed-form expression for them.
We notice little difference in the two proposed approaches, relative to the subpar reconstruction of \textsc{bsfp}.
This indirectly gives insight also into the performance of missing data imputation.

\vspace{15pt}

~

\vspace{15pt}

% ||||||||||||||||||||||||||||||||||||||||||||||||||||||||||
\section{Labor onset prediction: further results}\label{app_stelzer_extra_ris}

\setcounter{table}{0}
\setcounter{figure}{0}
\setcounter{equation}{0}
\setcounter{algocf}{0}

In the present section, we provide further details and results on the study analyzed in Section~4, originally published in \citet{Stelzer_2021_labor_onset}.
Prior to analysis, we standardized the data and log-transformed the metabolomics and proteomics features. Despite these preprocessing steps, all omics layers exhibited considerable deviation from Gaussianity, with over $30\%$ of features in each view yielding univariate Shapiro test statistics below $0.95$. To address this challenge, we target copula factor model variants \citep{Murray_2013_Copula_FA} for \textsc{jfr}, \textsc{jafar}, and \textsc{bsfp}, as detailed in Web Appendix~C. Given the continuous nature of the omics data and the absence of missing entries, the incorporation of the copula layer boils down to a deterministic preprocessing procedure, involving feature-wise transformations that leverage estimates of the associated empirical cumulative distribution functions.

\begin{figure}[ht!]
    \centering
\includegraphics[width=0.45\linewidth]{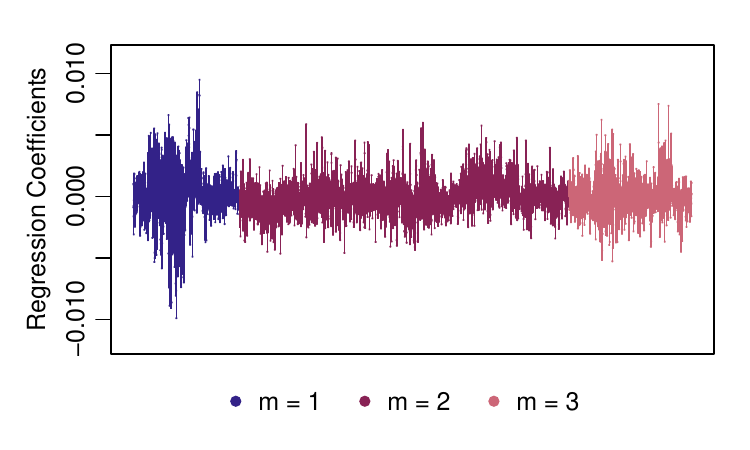}
\put(-100,125){\makebox(0,0){\textsc{jfr}}}
\includegraphics[width=0.45\linewidth]{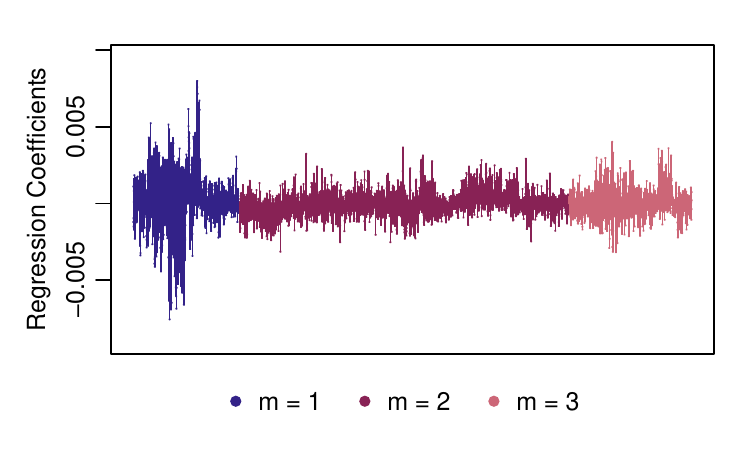}
\put(-100,125){\makebox(0,0){\textsc{jafar}}}
\\[5pt]
\includegraphics[width=0.45\linewidth]{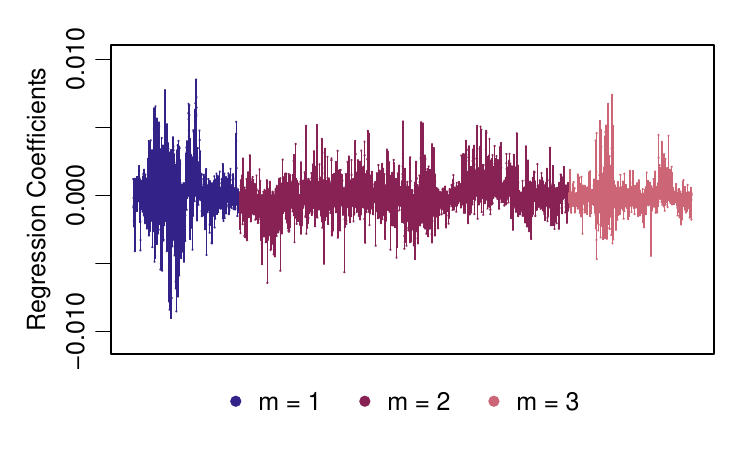}
\put(-100,125){\makebox(0,0){Tempered \textsc{jfr}}}
\includegraphics[width=0.45\linewidth]{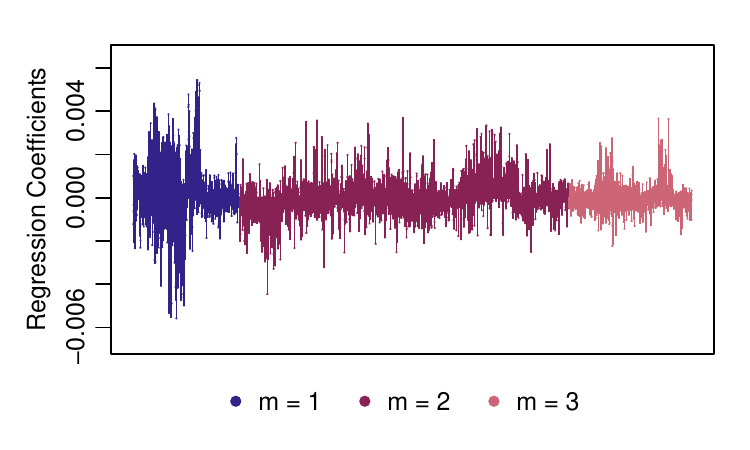}
\put(-100,125){\makebox(0,0){Tempered \textsc{jafar}}}
% \vspace{-15pt}
\caption{Induced linear regression coefficients for the response $y_i$ on the omics data $\{\bx_{m i} \}_m$ for the data application in Section~\ref{sec_application}. The reported values correspond to the average over the \textsc{mcmc} chain for the same exemplar random split considered in Figure~\ref{fig_response_prediction} and Figure~\ref{fig_inferred_cor}.}
\label{fig_lin_reg_coeff}
\end{figure}

\textcolor{black}{Regarding \textsc{mcmc} convergence, we obtained consistent results across runs with different starting points, including coherence in the selected ranks.}
In Figure~\ref{fig_lin_reg_coeff}, we report the induced coefficients for the regression directly tackling $p \big(\, y_i \mid \{\bX_{m i}\}_{m=1}^M, \relbar \big)$, i.e. marginalizing out all latent factors.
This supports the evidence that all omics layers bear predictive power.
These suggest that the corresponding joint latent sources of variation capture underlying biological processes that affect the system as a whole.
Tempering appears to be associated with sharper coefficients, which help to isolate clinically relevant features.

% \vspace{-5pt}

\begin{figure}[htbp!]
   \centering
   \begin{minipage}{0.25\textwidth}
       \centering
       \includegraphics[width=\linewidth]{Rev_Figures/Sec4_cov/s5_cor_Emp_m1_.png}
   \put(-65,115){\makebox(0,0){Immunome}}
   \put(-130,50){\rotatebox{90}{\makebox(0,0){Empirical}}}
   \end{minipage}
   \begin{minipage}{0.25\textwidth}
       \includegraphics[width=\linewidth]{Rev_Figures/Sec4_cov/s5_cor_Emp_m2_.png}
   \put(-65,115){\makebox(0,0){Metabolome}}
   \end{minipage}
   \begin{minipage}{0.25\textwidth}
       \centering
       \includegraphics[width=\linewidth]{Rev_Figures/Sec4_cov/s5_cor_Emp_m3_.png}
   \put(-65,115){\makebox(0,0){Proteome}}
   \end{minipage}
   \\
   % % % % % % % % % % % % % % % 
   \begin{minipage}{0.25\textwidth}
       \centering
       \includegraphics[width=\linewidth]{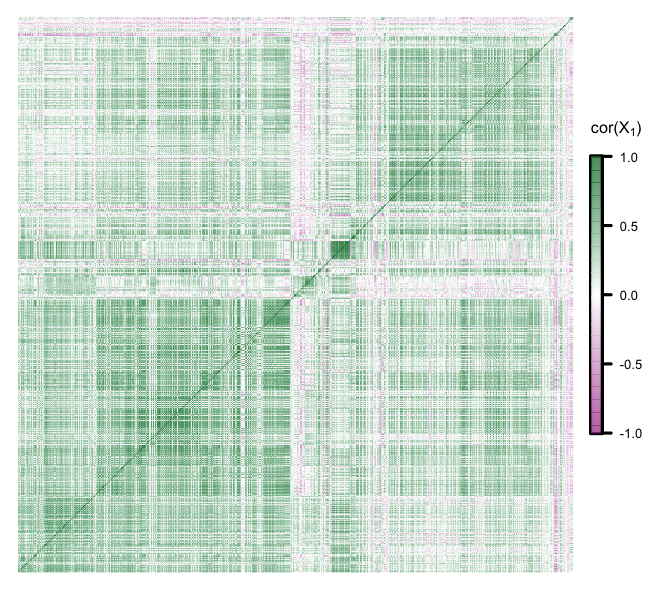}
   \put(-130,50){\rotatebox{90}{\makebox(0,0){\textsc{jfr}}}}
   \end{minipage}
   \begin{minipage}{0.25\textwidth}
       \centering
       \includegraphics[width=\linewidth]{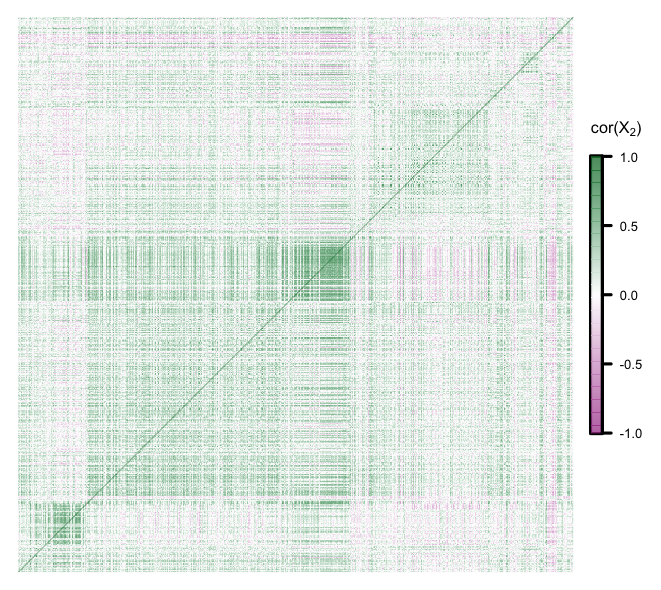}
   \end{minipage}
   \begin{minipage}{0.25\textwidth}
       \centering
       \includegraphics[width=\linewidth]{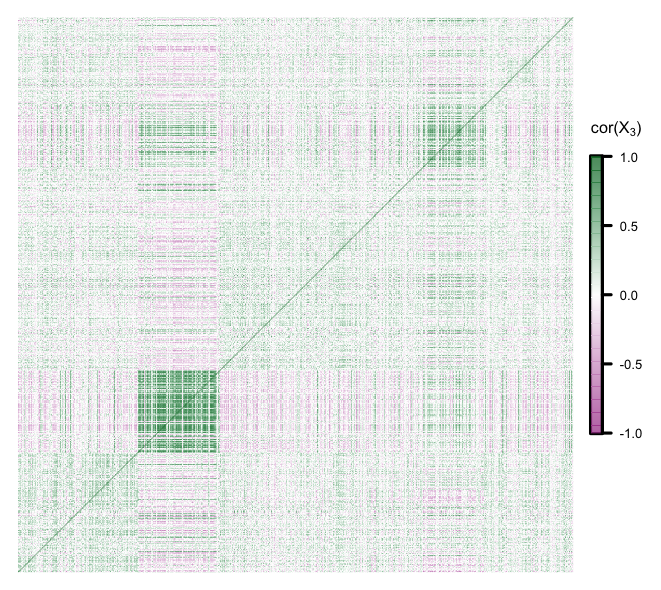}
   \end{minipage}
   \\% % % % % % % % % % % % % % % 
   \begin{minipage}{0.25\textwidth}
       \centering
       \includegraphics[width=\linewidth]{Rev_Figures/Sec4_cov/s5_cor_jfr_T_m1_.png}
   \put(-130,50){\rotatebox{90}{\makebox(0,0){\textsc{jfr}$_T$}}}
   \end{minipage}
   \begin{minipage}{0.25\textwidth}
       \centering
       \includegraphics[width=\linewidth]{Rev_Figures/Sec4_cov/s5_cor_jfr_T_m2_.png}
   \end{minipage}
   \begin{minipage}{0.25\textwidth}
       \centering
       \includegraphics[width=\linewidth]{Rev_Figures/Sec4_cov/s5_cor_jfr_T_m3_.png}
   \end{minipage}
   \\
   % % % % % % % % % % % % % % % 
   \begin{minipage}{0.25\textwidth}
       \centering
       \includegraphics[width=\linewidth]{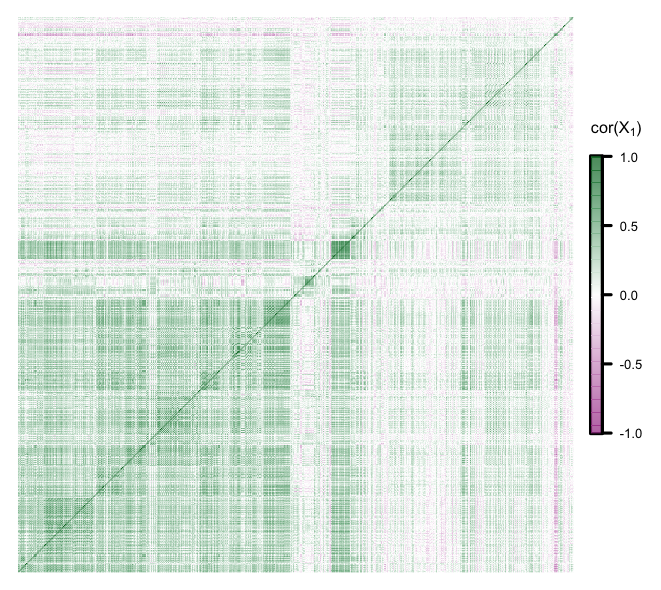}
   \put(-130,50){\rotatebox{90}{\makebox(0,0){\textsc{jafar}}}}
   \end{minipage}
   \begin{minipage}{0.25\textwidth}
       \centering
       \includegraphics[width=\linewidth]{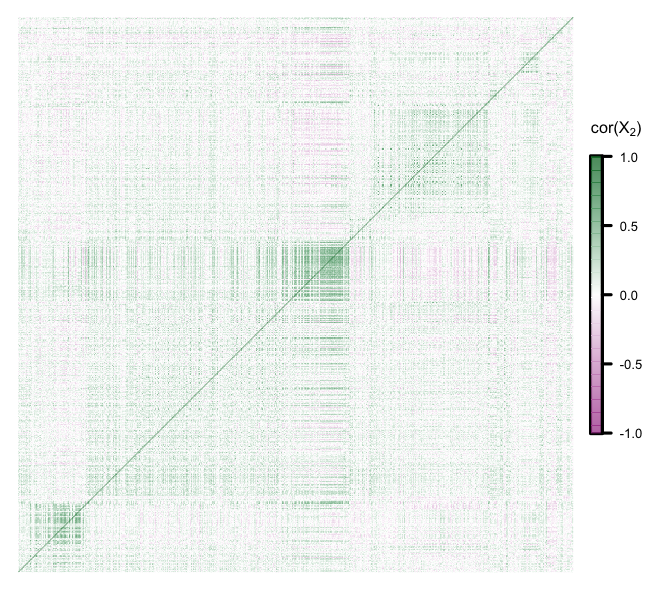}
   \end{minipage}
   \begin{minipage}{0.25\textwidth}
       \centering
       \includegraphics[width=\linewidth]{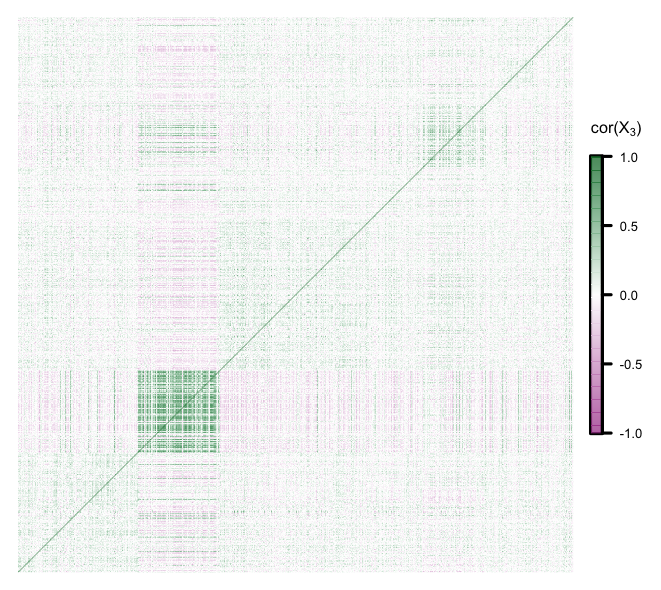}
   \end{minipage}
   \\
   % % % % % % % % % % % % % % % 
   \begin{minipage}{0.25\textwidth}
       \centering
       \includegraphics[width=\linewidth]{Rev_Figures/Sec4_cov/s5_cor_jafar_T_m1_.png}
   \put(-130,50){\rotatebox{90}{\makebox(0,0){\textsc{jafar}$_T$}}}
   \end{minipage}
   \begin{minipage}{0.25\textwidth}
       \centering
       \includegraphics[width=\linewidth]{Rev_Figures/Sec4_cov/s5_cor_jafar_T_m2_.png}
   \end{minipage}
   \begin{minipage}{0.25\textwidth}
       \centering
       \includegraphics[width=\linewidth]{Rev_Figures/Sec4_cov/s5_cor_jafar_T_m3_.png}
   \end{minipage}
   \\[5pt]
   %%%%%%%%%%%%%%%%%%%%%%%%%%%%%%%
   \caption{Correlation reconstruction for the three omics layers in the data application from Section~\ref{sec_application}, visualized in the same random split considered in Figure~\ref{fig_inferred_cor} from the main paper.}
   \label{sec4_inferred_cor_extra}
\end{figure}

% \newpage

\begin{figure}[b!]
   \centering \hspace{17pt}
   \begin{minipage}{0.48\textwidth}
    \centering ~\hspace{-18pt}~
    \includegraphics[height=1.5cm,width=2.8cm,
    trim=0 1.5cm 1.5cm 1.5cm, clip]{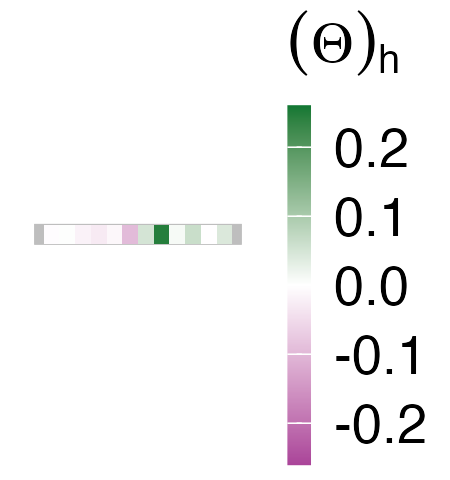}
    \put(-40,37){\makebox(0,0){$\theta$}}
   \end{minipage}
   \begin{minipage}{0.15\textwidth}
    \centering ~\hspace{-11pt}~
    \includegraphics[height=1.5cm,width=2.5cm,
    trim=0 1.5cm 1.5cm 1.5cm, clip]{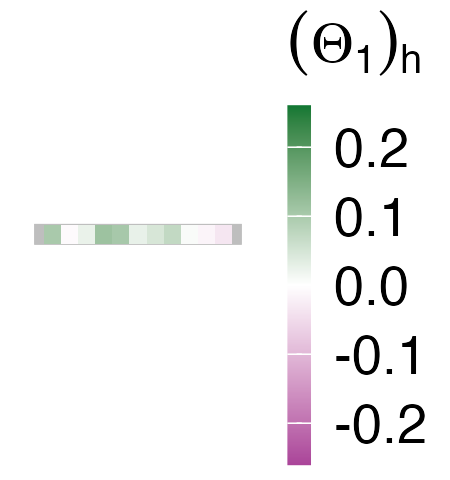}
    \put(-37,37){\makebox(0,0){$\theta_1$}}
   \end{minipage}
   \begin{minipage}{0.09\textwidth}
    \centering ~\hspace{-12pt}~
    \includegraphics[height=1.5cm,width=1.4cm,
    trim=0 1.5cm 1.5cm 1.5cm, clip]{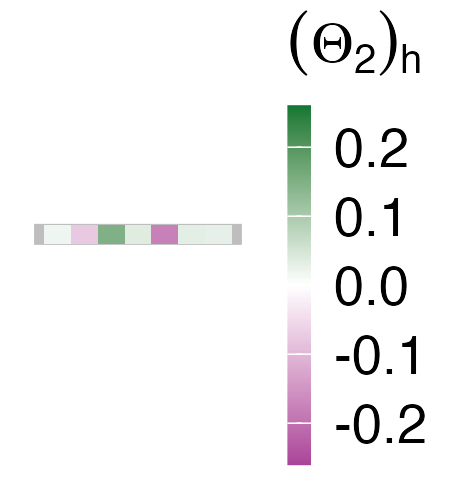}
    \put(-17,37){\makebox(0,0){$\theta_2$}}
   \end{minipage}
   \begin{minipage}{0.10\textwidth}
    \centering ~\hspace{-3pt}~
    \includegraphics[height=1.5cm,width=1.65cm,
    trim=0 1.5cm 1.5cm 1.5cm, clip]{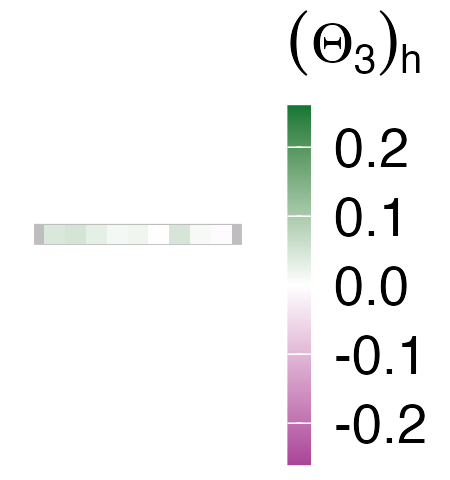}
    \put(-17,37){\makebox(0,0){$\theta_3$}}
   \end{minipage}
   \\
   %%%%%%%%%%%%%%%%%%%%%%%%%%%%%%%%%
   \begin{minipage}{0.16\textwidth}
    \centering
    \includegraphics[width=2.9cm,
    height=10cm, trim=0 0 1.8cm 0, clip]{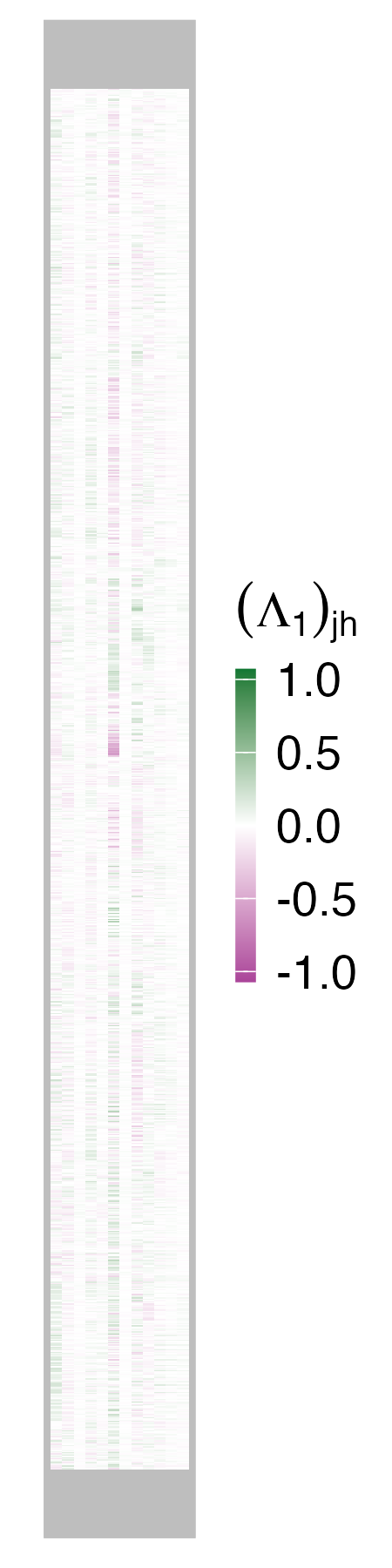}
    \put(-35,290){\makebox(0,0){$\Lambda_1$}}
   \end{minipage}
   \begin{minipage}{0.16\textwidth}
    \centering
    \includegraphics[width=2.9cm,
    height=10cm, trim=0 0 1.8cm 0, clip]{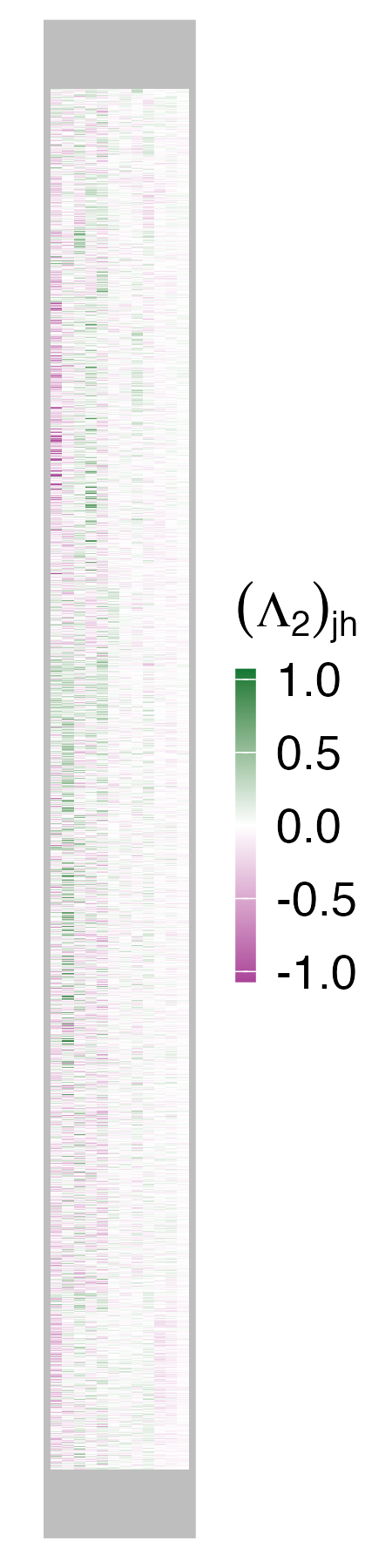}
    \put(-35,290){\makebox(0,0){$\Lambda_2$}}
   \end{minipage}
   \begin{minipage}{0.16\textwidth}
    \centering
    \includegraphics[width=2.9cm,
    height=10cm, trim=0 0 1.8cm 0, clip]{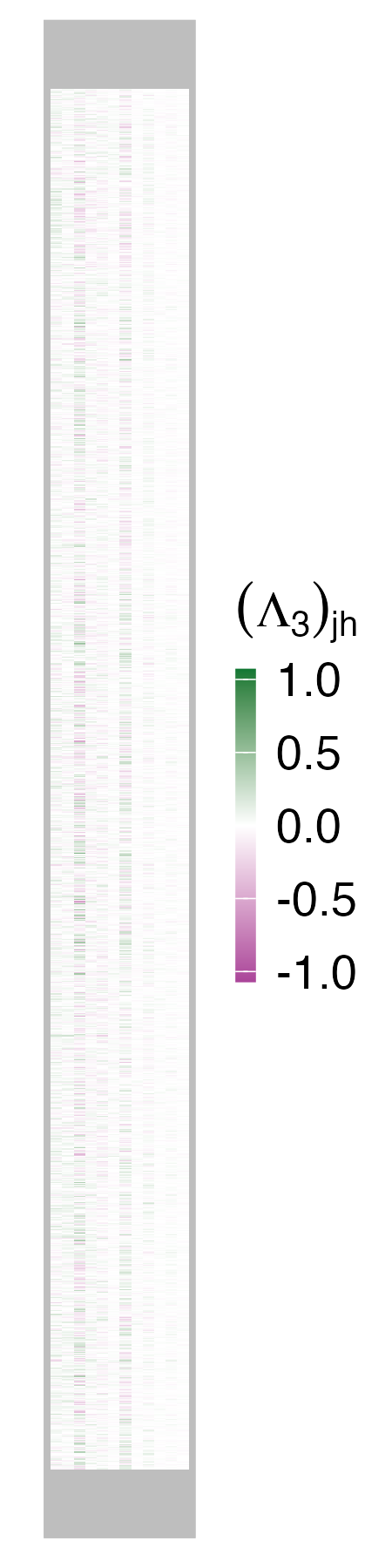}
    \put(-35,290){\makebox(0,0){$\Lambda_3$}}
   \end{minipage}
   %%%%%%%%%%%%%%%%%%%%%%%%%%%%%%%%%
   \begin{minipage}{0.15\textwidth}
    \centering
    \includegraphics[width=2.7cm,
    height=10cm, trim=0 0 1.8cm 0, clip]{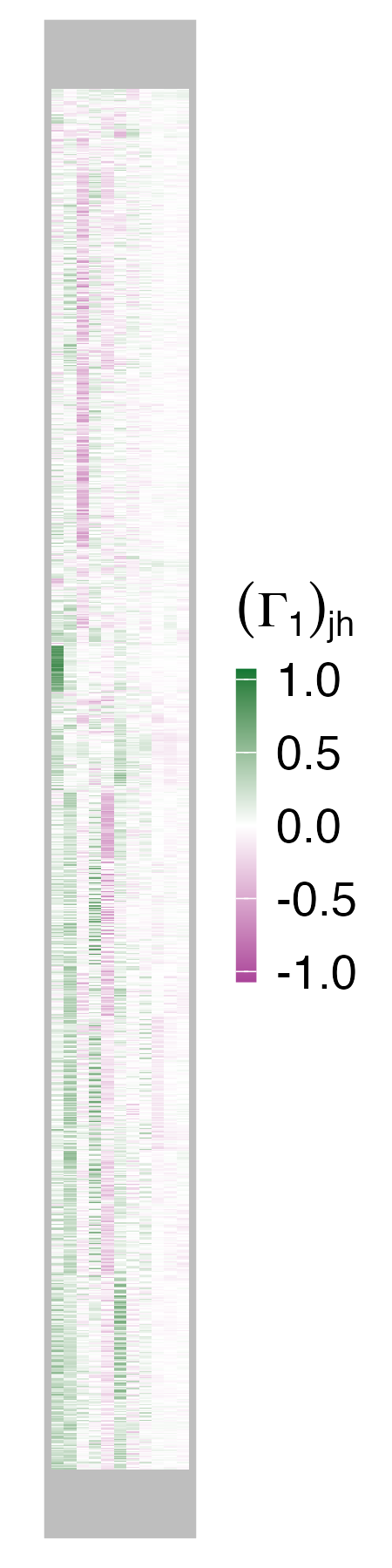}
    \put(-32,290){\makebox(0,0){$\Gamma_1$}}
   \end{minipage}
    \hspace{2.5pt}
   \begin{minipage}{0.09\textwidth}
    \centering
    \includegraphics[width=1.5cm,
    height=10cm, trim=0 0 1.8cm 0, clip]{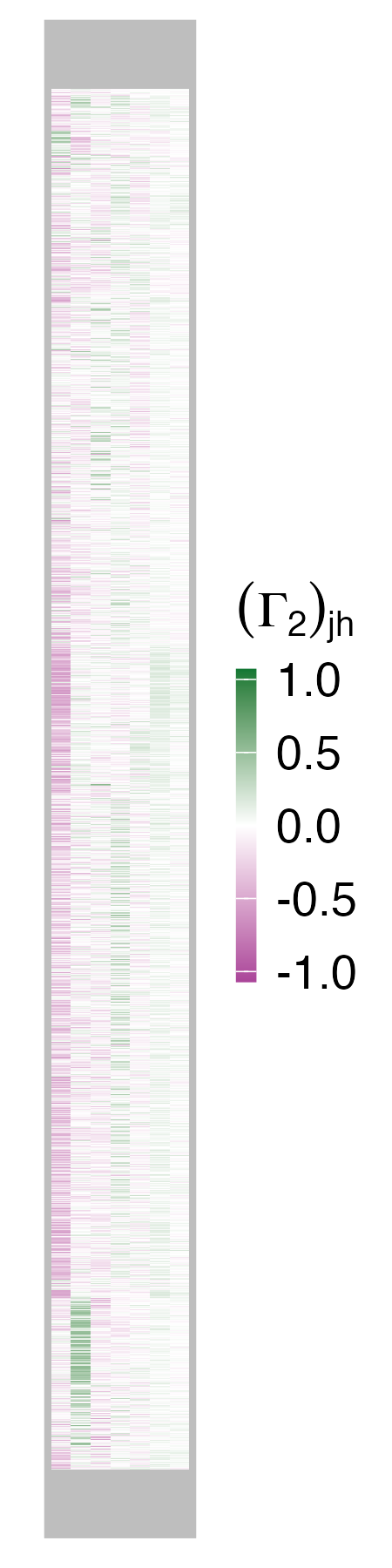}
    \put(-18,290){\makebox(0,0){$\Gamma_2$}}
   \end{minipage}
    \hspace{0pt}
   \begin{minipage}{0.10\textwidth}
    \centering
    \includegraphics[width=1.8cm,
    height=10cm, trim=0 0 1.8cm 0, clip]{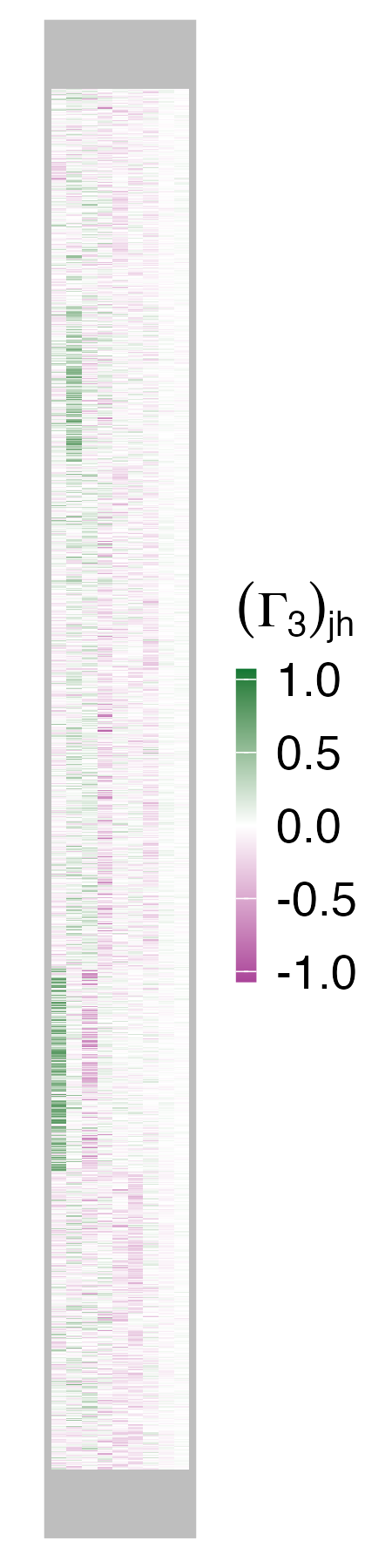}
    \put(-20,290){\makebox(0,0){$\Gamma_3$}}
   \end{minipage}
   \\[10pt]
   %%%%%%%%%%%%%%%%%%%%%%%%%%%%%
   \begin{minipage}{0.48\textwidth}
    \centering ~~~~~~
    \includegraphics[width=0.33\linewidth,
    trim=0.2cm 0.2cm 0.2cm 2.5cm, clip]{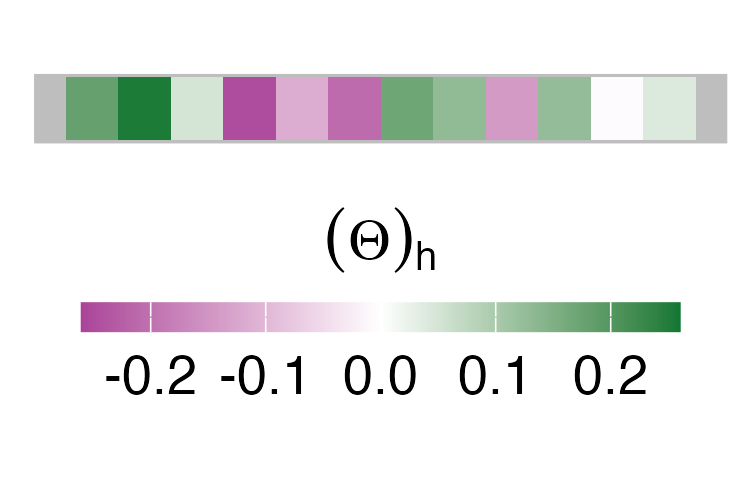}
    \put(-85,15){\makebox(0,0){$\theta_m$}}
   \end{minipage}
   \begin{minipage}{0.48\textwidth}
    \centering
    \includegraphics[width=0.33\linewidth,
    trim=0.2cm 0.2cm 0.2cm 2.5cm, clip]{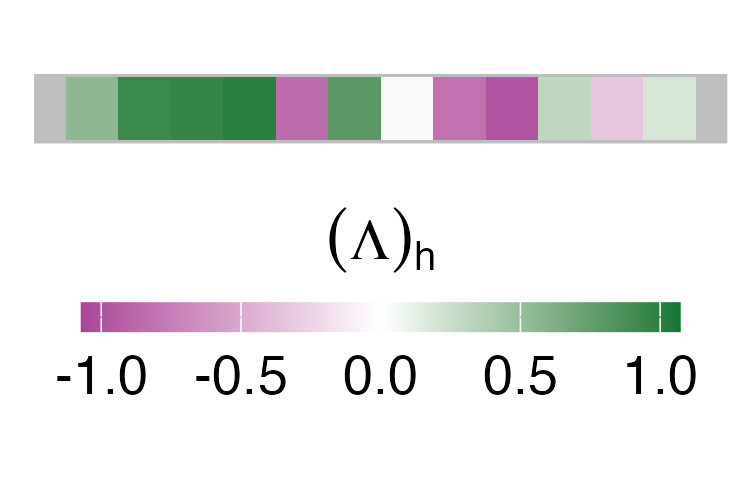}
    \put(-100,15){\makebox(0,0){$\Lambda_m, \, \Gamma_m$}}
   \end{minipage}
   % %%%%%%%%%%%%%%%%%%%%%%
   \begin{tikzpicture}[overlay, remember picture]
    \draw[black, thick] (-6.65cm, 0.9cm) -- (-6.65cm, 12.4cm);
    \draw[black, thick] (-4.15cm, 0.9cm) -- (-4.15cm, 12.4cm);
    \draw[black, thick] (-2.45cm, 0.9cm) -- (-2.45cm, 12.4cm);
    \end{tikzpicture}
\caption{Posterior means of the shared and view-specific loadings matrices inferred by  \textsc{jafar}$_T$ after post-processing via multiview \texttt{Varimax}. The results are reported for one exemplar random split of the dataset analyzed in Section~\ref{sec_application}.
Here, \textsc{jafar}$_T$ learns $K=12$ shared factors and $K_1=11$, $K_2=7$, and $K_3=9$ view-specific factors immunome, metabolome, and proteome data, respectively.}
\label{fig_postrprocess}
\end{figure}

\vspace{-5pt}

Figure~\ref{sec4_inferred_cor_extra} is an extended version of Figure~\ref{fig_inferred_cor} from the main paper, including the reconstructed covariances for the untempered versions of \textsc{jfr} and \textsc{jafar} of the same exemplar random split.
Both \textsc{jfr} and \textsc{jafar} capture the main features of the dependence structure, though with some limitations. In both cases, the tempered versions improve reconstruction, as is evident even from visual inspection. 
Still,
\textsc{jfr}$_T$ tends to be overconfident in estimating certain entries, for example, in the two off-diagonal blocks of positive correlations visible in the rightmost part of the immunome covariance matrix.
Conversely, \textsc{jafar}$_T$ exhibits a slight overall underestimation.
These opposing effects balance out, resulting in comparable quantitative reconstruction errors, as reported in Figure~\ref{fig_inferred_cor}. %of the main manuscript.
While achieving more accurate covariance reconstruction in such a high-dimensional setting would likely require many more observations, recovering the main block structures still enables meaningful interpretation of the latent directions of variation.

\subsection{Interpretation of latent factors}

\begin{figure}[t!]
    \centering
    \includegraphics[width=\linewidth]{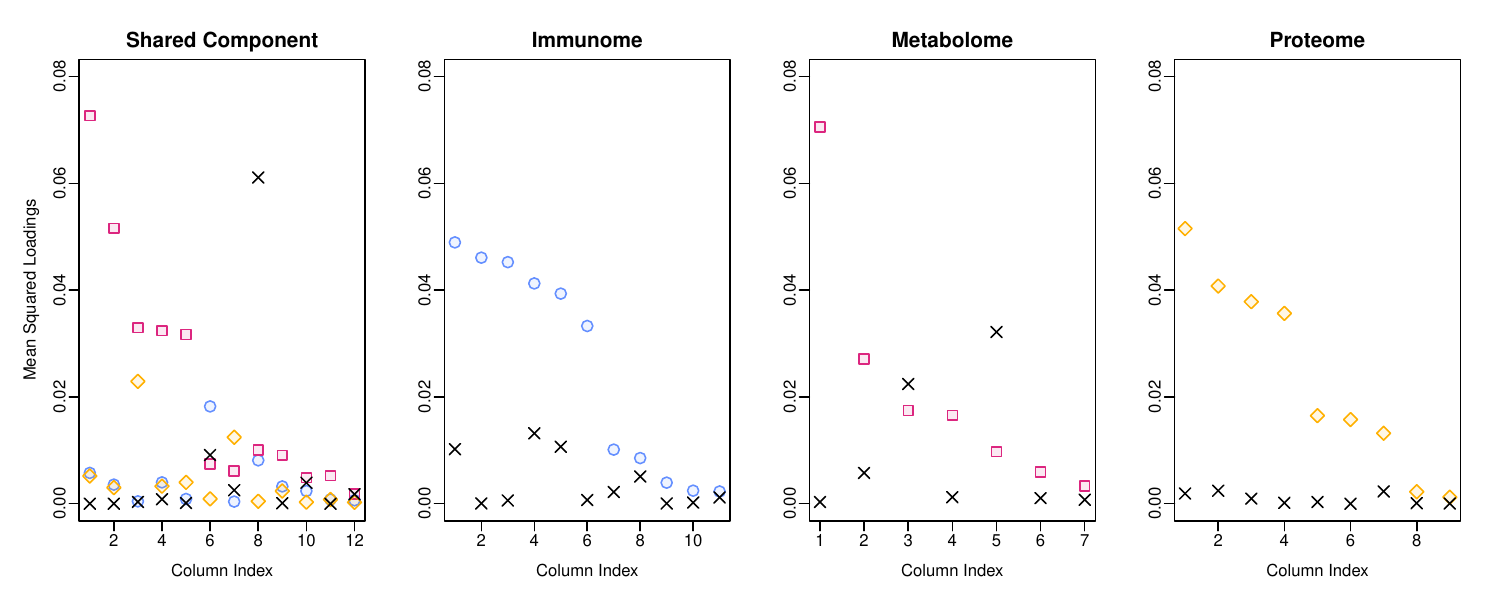}
    \caption{Squared loadings values across column indices, considering the mean across features for omics data. The three leftmost panels unpack the specific components. In all panels, violet circles correspond to immunome data, red squares to metabolome, yellow rhomboids to proteome, and black crosses to time-to-labor.
    }
    \label{fig_rotated_colMeans2}
\end{figure}

\begin{figure}[hb!]
    \centering
    \begin{minipage}[b]{0.33\textwidth}
        \centering
        \includegraphics[width=\textwidth]{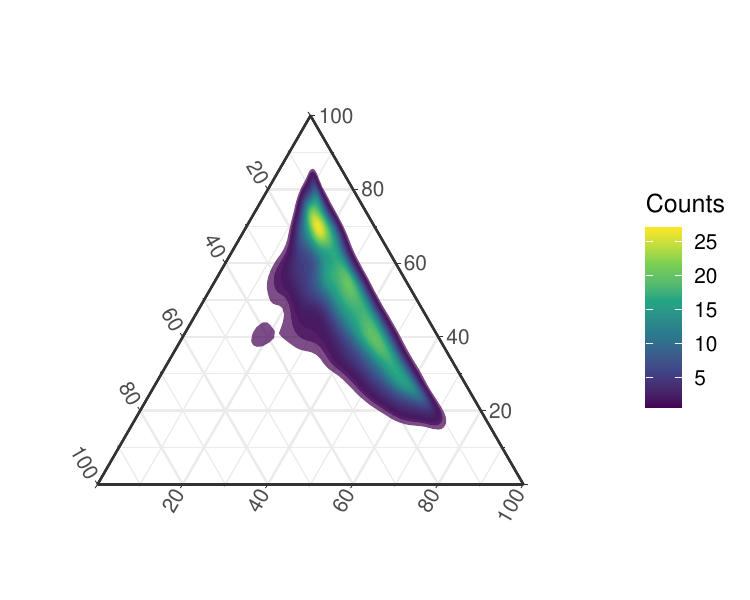}
        \put(-70,100){Immunome}
        \put(-145,70){$\Lambda_1$}
        \put(-60,70){$\Gamma_1$}
        \put(-100,5){$\sigma_1^2$}
    \end{minipage}%
    \begin{minipage}[b]{0.33\textwidth}
        \centering
        \includegraphics[width=\textwidth]{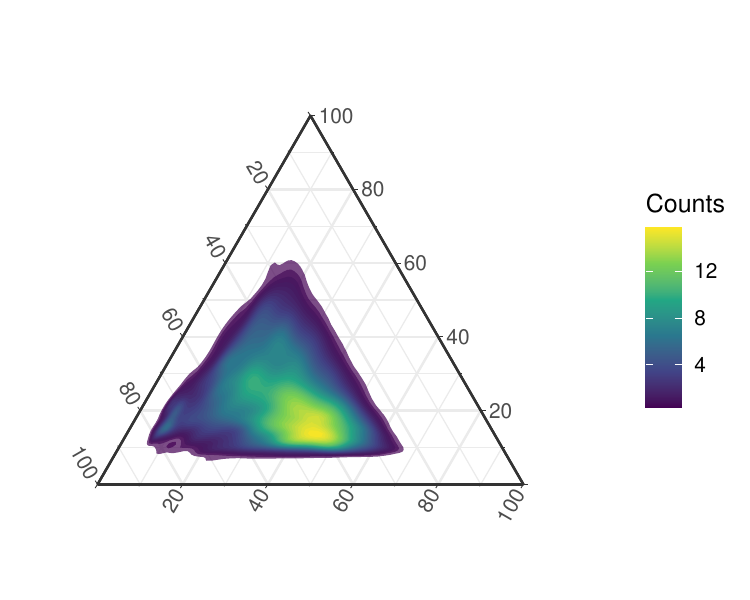}
        \put(-70,100){Metabolome}
        \put(-145,70){$\Lambda_2$}
        \put(-60,70){$\Gamma_2$}
        \put(-100,5){$\sigma_2^2$}
    \end{minipage}%
    \begin{minipage}[b]{0.33\textwidth}
        \centering
        \includegraphics[width=\textwidth]{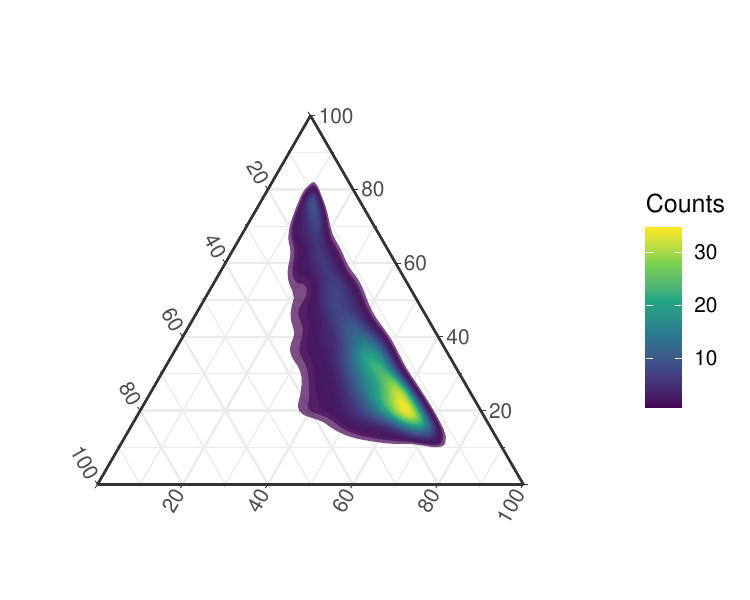}
        \put(-70,100){Proteome}
        \put(-145,70){$\Lambda_3$}
        \put(-60,70){$\Gamma_3$}
        \put(-100,5){$\sigma_3^2$}
    \end{minipage}
    \caption{Percentages of variance explained within \textsc{jafar} across the features of each omics layer. For each feature, we considered the posterior mean.}
    \label{fig_ternary_plots_var_explained}
\end{figure}

In Figure~\ref{fig_postrprocess}, we report the posterior mean of the shared loading matrices after postprocessing with the extended version of \texttt{MatchAlign} via Multiview \texttt{Varimax}, and in Figure~\ref{fig_rotated_colMeans2} we show the mean squared loadings across factors.
Visual inspection suggests stronger view-specific signals than shared ones for the immunome and proteome, whereas the metabolome shows more balanced contributions.
This impression is confirmed by Figure~\ref{fig_ternary_plots_var_explained}, which summarizes the proportions of variance explained in each omics using ternary plots. 
Notably, the $8^{th}$ shared factor and the $5^{th}$ metabolome-specific factor exhibit the “least important eigenvalue” phenomenon \citep{Carvalho_2008_gene_expression}, where a factor is highly relevant for the response but contributes little to the predictors.
Importantly, the low response loadings on immunome-specific factors should not be taken as evidence of a lack of predictive signal, as clarified by the regression coefficients in Figure~\ref{fig_lin_reg_coeff}. 
\textcolor{black}{Below, we highlight the interpretation of some clinically relevant factors.}

\begin{itemize}[leftmargin=2em]
    \item[$\bLambda_{\bigcdot\,\bigcdot\,6}$ –] Immunome-dominant with supporting lipids, positively associated with labor timing. Positive loadings reflect basal regulatory signaling across immune subsets, paired with structural/storage lipids; negatives capture stimulus-driven inflammatory responses and polar metabolites. This factor mainly represents baseline immune balance modulated by lipid metabolism. Regulatory immune signaling coupled with structural/storage lipids—predict delayed labor, whereas stronger inducible inflammatory responses with polar metabolites predict earlier onset \citep{Sharma2022_Immune_metabolic}.
    \item[$\bLambda_{\bigcdot\,\bigcdot\,8}$ –] Balanced immunome and metabolome contributions, with a strong positive association to labor onset. Positive loadings reflect cytokine-tuned, tolerogenic signaling with polar/mid-sized metabolites; negatives emphasize inducible inflammatory responses and large lipid accumulation. This factor captures immune tolerance coupled with metabolic flexibility. Tsolerogenic signaling with polar metabolite enrichment predicts later onset, whereas inducible inflammation coupled with lipid accumulation predicts earlier labor \citep{Stelzer_2021_labor_onset}.
    \item[$\bGamma_{1\,\bigcdot\,1}$ -] The $1^{st}$ immunome-specific factor captures baseline immune signaling across both innate and adaptive compartments, including naïve and memory CD4$^+$/CD8$^+$ T cells, B cells, NK cells, and monocyte/dendritic cell subsets. Positive loadings reflect high basal activity in pathways such as NF-kB (IkB), S6, and MAPKAPK2, indicating a homeostatic immune tone. 
    Positive association with time to labor implies that higher baseline immune signaling is linked to delayed labor onset, consistent with the concept that balanced maternal immune activity helps maintain gestational homeostasis \citep{Aghaeepour2017_immunome}.
    \item[$\bGamma_{1\,\bigcdot\,4}$ -]  The $4^{th}$ immunome-specific factor reflects STAT5-driven signaling across monocytes, MDSCs, dendritic cells, NK cells, especially under GM-CSF and IFN$\alpha$ stimulation. Positive loadings indicate strong baseline and cytokine-induced STAT5 activity, consistent with immune readiness and balance. Its positive association with time to labor suggests that enhanced STAT5 signaling may help maintain gestational homeostasis and delay labor onset \citep{kazuo2021_immunome}.
   \item[$\bGamma_{1\,\bigcdot\,5}$ -] The $5^{th}$ immunome-specific factor captures maternal myeloid cell signaling, with strong negative loadings from classical, intermediate, and non-classical monocytes, MDSCs, and dendritic cells under STAT6, CREB, S6, NF-kB, ERK, and p38 pathways following GM-CSF, LPS, or IL2/4/6 stimulation. Because the response loads positively while these features load negatively, higher stimulus-responsive signaling in myeloid populations is associated with earlier labor onset, consistent with the role of a quiescent immune state in maintaining gestational homeostasis \citep{Aghaeepour2017_immunome}.
   \item[$\bGamma_{2\,\bigcdot\,3}$ -] The $3^{rd}$ metabolome-specific factor is positively associated with time to labor. Higher levels of long-chain lipids, diacylglycerols, triacylglycerols, and lysophospholipids (positively loaded features) predict delayed labor onset, whereas higher levels of small polar metabolites such as amino acids and energy intermediates (negatively loaded features) predict earlier labor onset. This factor thus highlights a contrast between lipid mobilization, linked to prolonged gestation, and small-molecule enrichment, linked to shorter time to labor \citep{Yi2025_metabolites}.
   \item[$\bGamma_{2\,\bigcdot\,5}$ -] The $5^{th}$ metabolome-specific factor is positively loaded by long-chain lipids, glycerophospholipids, triacylglycerols, and sterol-like features, while negatively loaded by small- to mid-sized polar metabolites such as amino acid derivatives and energy intermediates. With a strong positive association with time to labor, this factor captures a lipid–polar metabolite axis in which lipid dominance predicts prolonged gestation. Unlike $\bGamma_{2\,\bigcdot\,3}$, which emphasizes bioactive signaling lipids such as DAGs and LPCs, this factor is driven more by bulk structural and storage lipids \citep{Hong2020_metabolomic}.
\end{itemize}

\subsection{Time-to-labor prediction under block-missingness for metabolome data}\label{app_application_validation_set}

In this section, we illustrate the potential of the proposed methodologies for handling mixed data scenarios.
Beyond sparse entry-wise missingness, which can be addressed through standard imputation, the proposed composite factor models are particularly well-suited to deal with block-wise missingness.
In particular, when an entire modality is absent for new subjects, our framework effectively handles such cases by leveraging the collapsed model, obtained by marginalizing out the missing view from the full fit.
In fact, this scenario arises in the same study analyzed in Section~\ref{sec_application}, which includes an additional small cohort of $n_{\mathrm{valid}}=8$ subjects lacking metabolome measurements.
The data are available in the GitHub repository \texttt{IntegratedLearner} as a side validation dataset.
Previous analyses using supervised integrative methods on this study (e.g., \citet{mallick2024integrated}, \citet{Ding_2022_coopL}) excluded these subjects due to the structural inability of the associated methodologies to handle the missing modality.
By contrast, our methodologies and accompanying code readily enable time-to-labor predictions using the available views, immunome and proteome.

\begin{figure}[h!]
\centering
\begin{minipage}{0.49\textwidth}
\centering
\includegraphics[width=\linewidth]{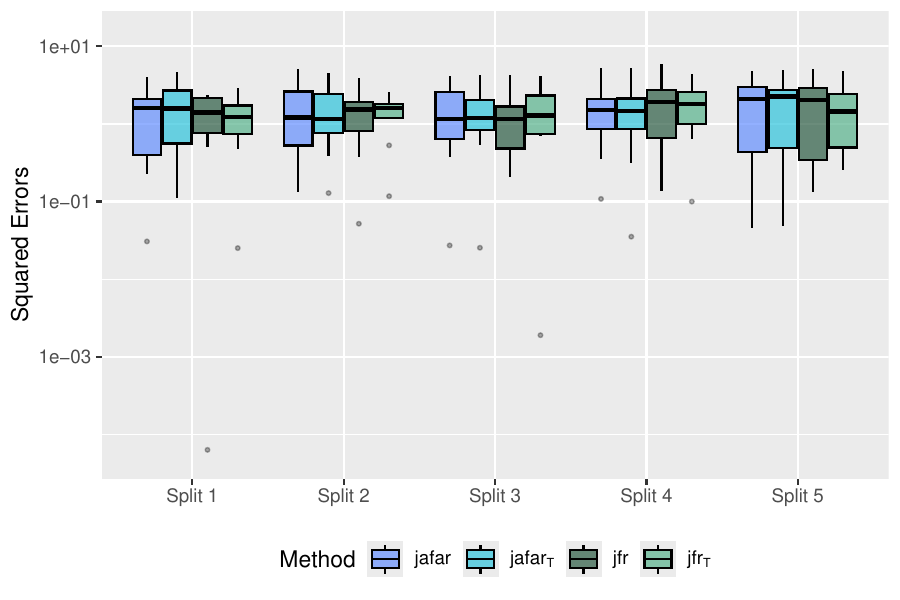}
\end{minipage}
\begin{minipage}{0.49\textwidth}
\centering
\vspace{-30pt}
\begin{tabular}{c|rrrr}
\toprule
 &  jafar & jafar$_T$ & jfr & jfr$_T$ \\
\cmidrule{1-5}
MSE & 1.81 & 1.80 & 1.74 & 1.67 \\
$R^2$ & 0.15 & 0.16 & 0.18 & 0.22 \\
90\% C.I. & 100 & 100 & 100 & 97.5 \\
\bottomrule
\end{tabular}
\end{minipage}
\caption{Out-of-sample predictive performance on the additional validation data from the application in Section~4. Boxplots of the prediction squared errors across five random splits of the training data (left), and means over the same splits for different aggregated metrics (right). Metabolome measurements were absent for all subjects included in the validation dataset.}
\label{fig_sec4_validation}
\end{figure}

An exploratory data analysis revealed the need for batch effect correction to align the scales of the Immunome measurements, which we addressed using the \texttt{ComBat} function from the \texttt{sva} package in \texttt{R}.
Summaries of the resulting predictions are reported in Figure~\ref{fig_sec4_validation}.
Compared to Figure~\ref{fig_response_prediction}, the mean squared error is lower than that obtained on the main test set, but the $R^2$ is also reduced.
This mismatch highlights the increased difficulty of predicting for this small cohort and is further reflected in the over-coverage of the predictive intervals.

\end{document}